%% file: root.tex
\documentclass[journal,twoside]{IEEEtran}

\IEEEoverridecommandlockouts                              %

\usepackage{graphicx} %
\usepackage{xcolor}
\usepackage{amsthm}
\usepackage{amsmath} %
\usepackage{amssymb}  %
\usepackage[utf8]{inputenc}
\usepackage{cite}
\usepackage{nicefrac}
\usepackage{siunitx}
\usepackage{placeins}
\usepackage{hyperref}
\usepackage[caption=false,font=footnotesize]{subfig}
\usepackage{fixltx2e}
\usepackage{wrapfig}

\input{shortcuts}

\graphicspath{{pdftex}{figures}}

\author{Arne Sachtler$^{1,2}$ and Alin Albu-Schäffer$^{2,1}$%
\thanks{Manuscript received: September 7, 2021; revised December 6, 2021; accepted December 23, 2021.}%
\thanks{This paper was recommended for publication by Associate Editor T. Asfour and Editor C. Gosselin upon evaluation of the reviewers' comments.
This work is supported by the ERC advanced grant \emph{M-Runners} (ID: 835284) by the European Research Council (ERC).}%
\thanks{$^{1}$A. Sachtler and A. Albu-Schäffer are with the Technical University of Munich (TUM), Department of Informatics; Boltzmannstr. 3, 85748 Garching;}%
\thanks{$^{2}$They are also with the German Aerospace Center (DLR), Institute of Robotics and Mecha\-tronics; Münchener Str. 20, 82234 Weßling, Germany}%
\thanks{\textcolor{blue}{\copyright 2022 IEEE. Official URL: \url{https://ieeexplore.ieee.org/document/9681384}}}
}

\title{
    Strict Modes Everywhere -- 
    Bringing Order into Dynamics of Mechanical Systems by a Potential Compatible with the Geodesic Flow
}

\begin{document}
\bstctlcite{IEEEexample:BSTcontrol}
\nocite{AlbuSchaeffer2021}
\maketitle
\begin{abstract}
\input{abstract}
\end{abstract}
\begin{IEEEkeywords}
Compliant Joints and Mechanisms;
Modeling, Control, and Learning for Soft Robots;
Mechanism Design;
Energy and Environment-Aware Automation;
Dynamics
\end{IEEEkeywords}
\section{Introduction}\label{sec:introduction}
\IEEEPARstart{S}{everal}
\input{sec/introduction}
\input{sec/modes}
\input{sec/approach}
\input{sec/casestudy}

\input{sec/controllers}

\input{sec/concl}

\appendices
\input{sec/appendix}

\bibliographystyle{IEEEtran}
\bibliography{bibliography}

\end{document}

%% file: shortcuts.tex
\newcommand{\Qc}{{\mathcal{Q} }}

\newcommand{\bs}[1]{\boldsymbol{#1}}
\newcommand{\ttheta}{\bs{\theta}}

\newcommand{\qq}{\bs{q}}

\newcommand{\vv}{\bs{v}}

\newcommand{\dd}{\bs{d}}

\newcommand{\CC}{\bs{C}}
\newcommand{\DD}{\bs{D}}

\newcommand{\II}{\bs{I}}

\newcommand{\MM}{\bs{M}}

\newcommand{\ggamma}{\bs{\gamma}}

\newcommand{\ttau}{\bs{\tau}}

\newcommand{\oomega}{\bs{\omega}}

\newcommand{\LLambda}{\bs{\Lambda}}

\newcommand{\nnull}{\bs{0}}

\newtheorem{theorem}{Theorem}
\newtheorem{theorem*}{Theorem}

\newtheorem*{definition*}{Definition}

\newtheorem{lemma*}{Lemma}

\newtheorem{corollary*}{Corollary}

\newtheorem*{problem*}{Problem Statement}
\newtheorem{proof*}{Proof}

\makeatletter
\renewcommand*\env@matrix[1][\arraystretch]{%
	\edef\arraystretch{#1}%
	\hskip -\arraycolsep
	\let\@ifnextchar\new@ifnextchar
	\array{*\c@MaxMatrixCols c}}
\makeatother

\definecolor{TUMBlue}{HTML}{0065BD}

\newcommand{\q}{\boldsymbol{q}}
\newcommand{\qdot}{\dot{\boldsymbol{q}}}
\newcommand{\qddot}{\ddot{\boldsymbol{q}}}

\newcommand{\qeq}{\boldsymbol{q}_{\mathrm{eq}}}

\newcommand{\torque}{\boldsymbol{\tau}}

\newcommand{\M}{\boldsymbol{M}}

\newcommand{\Minv}{\boldsymbol{M}^{-1}}

\renewcommand{\b}[1]{\boldsymbol{#1}}
\newcommand{\gd}{\dot{\boldsymbol{\gamma}}}

\DeclareUnicodeCharacter{2212}{-}

\DeclareMathOperator{\atantwo}{atan2}
\newcommand*{\tran}{^\mathsf{T}}

%% file: abstract.tex
Strict nonlinear normal modes provide very regular families of oscillations within conservative mechanical systems.
However, a strict normal mode will generally be an isolated curve within the configuration space of the system.
In this paper, we design a potential that will densely fill the configuration space with strict normal modes such that each configuration belongs to one mode and each mode passes through a common point, the equilibrium.
As the potential can be realized by (nonlinear) elastic elements it can be used to execute a variety of periodic trajectories very efficiently.
Most of the required torques will come from the elastic elements in the system and not from the actuators.
We also design a controller stabilizing the system to a desired target mode and a controller performing swing-up and compensating dissipated energy.
Finally, we showcase the approach for a two DoF manipulator.
The experiments show that the approach performed well for the example system.

%% file: sec/introduction.tex
 applications of robotics can be seen as periodic or close-to-periodic motions.
Consider a pick-and-place application where a robot picks objects from a conveyor belt and assorts them into a box.
This can be seen as a sequence of close-to-periodic motions: a base oscillation between the conveyor belt and the center of the box is slightly modulated in order to reach the specific positions in the box.
Beside the industrial case, also legged and non-legged locomotion involves robotic structures oscillating periodically.

Classical industrial robots and also many recent humanoid (e.g. \cite{Englsberger2014}) or zoomorphic robots use stiff actuation without any (wanted) elastic elements.
The actuation and controllers generate all the required forces to steer the system onto the desired trajectories.
In that case, a lot of energy exchange occurs between the actuation (usually the electrical part) and the mechanical part.
This comes with subobtimal energy efficiency, as energy exchange between these two parts is subject to losses.
Especially, mechanical energy is usually not recovered, but dumped to heat in the motor drivers.
Therefore, the question arises if elastic elements in the system may help to increase energy efficiency.
Elastic elements store and release mechanical energy and create a variety of natural oscillations in the mechanical system.
Optimally, these natural oscillations coincide with the desired trajectories.
Thus, the actuation will only need to stabilize them and it is expected that the effort needed for corrective actions on natural oscillations is much less than for classical trajectory tracking on stiff robots.
If the natural oscillations do not coincide with the desired trajectories, they might have the correct tendency and will only need to be modulated by the actuators.

The combination of robotic structures and elastic elements leads to very complex systems showing a wide variety of nonlinear (including chaotic \cite{Shinbrot1992}) behaviors.
Recently, the notion of nonlinear normal modes was introduced to the robotics community \cite{AlbuSchaeffer2020,AlbuSchaeffer2021}:
Despite the dynamics being very complex, very regular and simple, i.e., low-dimensional intrinsic motions can be found.
These can be collected on continuous families of periodic solutions.
A special case of such periodic solutions are \emph{strict nonlinear normal modes} \cite{AlbuSchaeffer2020}, which can be seen as invariant curves containing the equilibrium.
These curves lie within the configuration space of the elastic robotic system.
In particular, when starting at a configuration on the strict normal mode, it will oscillate back and forth periodically on a breaking trajectory and the configuration path will trace out a subpath of the strict mode.

\begin{figure}
	\centering
	\def\svgwidth{.75\columnwidth}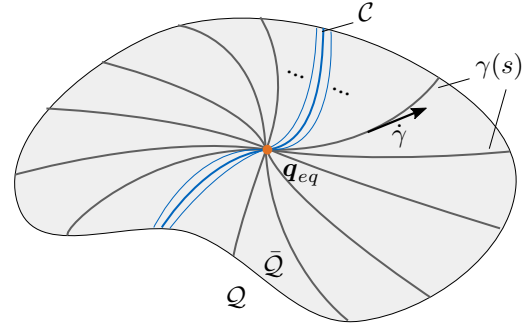
	\caption{
		Aim of this paper: find a potential function such that the configuration is densely packed with strict modes.
		}
	\label{fig:basic_scheme}
\end{figure}

Strict normal modes provide very structured oscillations.
However, they usually occur, if at all, as isolated curves within the configuration space.
The trajectories will be very complex if not initialized directly on a strict normal mode.

Several state of the art publications already deal with the exploitation of oscillations in elastic robotic systems.
Especially, in the field of legged locomotion research the question arises if the oscillations can be exploited \cite{Remy2011a,Gan2014,Kashiri2018,Vanderborght2006}.
These publications usually deal with periodic orbits in general, not with the special case of normal modes.
Moreover, a couple of publications deal with the question of how to control a robot onto a nonlinear normal mode \cite{Santina2021a,Santina2021}.
All these publications have one thing in common: the mechanics and the elastic elements are considered given and the main question is how to exploit existing nonlinear normal modes or periodic orbits.

In the present paper, we address another question: given the structure of a mechanical system, especially its multi-body dynamics; can we find elastic elements such that strict modes are everywhere, i.e., such that each point in configuration space belongs to a strict mode?
Such a system would always oscillate periodically when initializing it with zero velocity (or another \emph{compatible} velocity) from anywhere.
Therefore, the elastic elements would help, in a sense, ordering the complex mechanical system with otherwise complicated solution behavior.

%% file: pdftex/scheme.pdf_tex
\begingroup%
  \makeatletter%
  \providecommand\color[2][]{%
    \errmessage{(Inkscape) Color is used for the text in Inkscape, but the package 'color.sty' is not loaded}%
    \renewcommand\color[2][]{}%
  }%
  \providecommand\transparent[1]{%
    \errmessage{(Inkscape) Transparency is used (non-zero) for the text in Inkscape, but the package 'transparent.sty' is not loaded}%
    \renewcommand\transparent[1]{}%
  }%
  \providecommand\rotatebox[2]{#2}%
  \newcommand*\fsize{\dimexpr\f@size pt\relax}%
  \newcommand*\lineheight[1]{\fontsize{\fsize}{#1\fsize}\selectfont}%
  \ifx\svgwidth\undefined%
    \setlength{\unitlength}{425.37548636bp}%
    \ifx\svgscale\undefined%
      \relax%
    \else%
      \setlength{\unitlength}{\unitlength * \real{\svgscale}}%
    \fi%
  \else%
    \setlength{\unitlength}{\svgwidth}%
  \fi%
  \global\let\svgwidth\undefined%
  \global\let\svgscale\undefined%
  \makeatother%
  \begin{picture}(1,0.63347694)%
    \lineheight{1}%
    \setlength\tabcolsep{0pt}%
    \put(0,0){\includegraphics[width=\unitlength,page=1]{scheme.pdf}}%
    \put(0.41857501,0.03365919){\color[rgb]{0,0,0}\makebox(0,0)[lt]{\lineheight{1.25}\smash{\begin{tabular}[t]{l}$\mathcal{Q}$\end{tabular}}}}%
    \put(0.53424286,0.29195837){\color[rgb]{0,0,0}\makebox(0,0)[lt]{\lineheight{1.25}\smash{\begin{tabular}[t]{l}$\q_{eq}$\end{tabular}}}}%
    \put(0.49321111,0.09640271){\color[rgb]{0,0,0}\makebox(0,0)[lt]{\lineheight{1.25}\smash{\begin{tabular}[t]{l}$\bar{\mathcal{Q}}$\end{tabular}}}}%
    \put(0,0){\includegraphics[width=\unitlength,page=2]{scheme.pdf}}%
    \put(0.68556438,0.6004395){\color[rgb]{0,0,0}\makebox(0,0)[lt]{\lineheight{1.25}\smash{\begin{tabular}[t]{l}$\mathcal{C}$\end{tabular}}}}%
    \put(0,0){\includegraphics[width=\unitlength,page=3]{scheme.pdf}}%
    \put(0.63087115,0.4634206){\color[rgb]{0,0,0}\rotatebox{-18.266404}{\makebox(0,0)[lt]{\lineheight{1.25}\smash{\begin{tabular}[t]{l}...\end{tabular}}}}}%
    \put(0.54082063,0.4972928){\color[rgb]{0,0,0}\rotatebox{-18.266405}{\makebox(0,0)[lt]{\lineheight{1.25}\smash{\begin{tabular}[t]{l}...\end{tabular}}}}}%
    \put(0,0){\includegraphics[width=\unitlength,page=4]{scheme.pdf}}%
    \put(0.91817234,0.49652473){\color[rgb]{0,0,0}\makebox(0,0)[lt]{\lineheight{1.25}\smash{\begin{tabular}[t]{l}$\gamma(s)$\end{tabular}}}}%
    \put(0,0){\includegraphics[width=\unitlength,page=5]{scheme.pdf}}%
    \put(0.75185509,0.35876288){\color[rgb]{0,0,0}\makebox(0,0)[lt]{\lineheight{1.25}\smash{\begin{tabular}[t]{l}$\dot{\gamma}$\end{tabular}}}}%
    \put(0,0){\includegraphics[width=\unitlength,page=6]{scheme.pdf}}%
  \end{picture}%
\endgroup%

%% file: sec/modes.tex
\section{Strict Nonlinear Normal Modes}\label{sec:modes}

Consider a conservative mechanical system and let $\mathcal{Q}$ denote the corresponding configuration space manifold.
In coordinates it is characterized by the ordinary differential equation (ODE) \begin{equation}\label{eq:system-in-coordinates}
	\MM(\qq)\qddot + \CC(\qq, \qdot)\qdot + \frac{\partial U}{\partial \qq} = \nnull,
\end{equation}
where $\MM(\qq)$ denotes the positive definite and symmetric mass tensor, $\CC(\qq, \qdot)\qdot$ denotes centrifugal and Coriolis forces and $U(\q)$ denotes a scalar potential function.\footnote{When clear from the context we will refer to $\frac{\partial U}{\partial \qq}$ as a column vector.}
As in \cite{AlbuSchaeffer2021} we use some abuse of notation and refer to $\q$ as both the point on the manifold $\q \in \mathcal{Q}$ as well as its coordinates $\q \in \mathbb{R}^{\dim \mathcal{Q}}$.
In (\ref{eq:system-in-coordinates}) $\q$ denotes the configuration variable coordinates and $\qdot \in \mathcal{T}_{\q}\mathcal{Q}$ the joint velocities.

The ODE (\ref{eq:system-in-coordinates}) can also be written coordinate-independently.
Considering that, we endow the manifold $\mathcal{Q}$ with a Riemannian metric $g$ generated by the coordinate-independent version of the inertia tensor $\M(\qq)$.
Then, let $\b\nabla_\cdot \cdot$ denote the Levi-Civita connection \cite{Lee1997} on the Riemannian manifold $(\mathcal{Q}, g)$ and let $\mathcal{U}(\qq)$ be the coordinate-independent version of the potential.
We can then write the system dynamics as \begin{equation}\label{eq:coordinate-free}
	\b{\nabla}_{\qdot}\qdot = - \b{\nabla}\mathcal{U},	
\end{equation}
where $\b{\nabla}$ (no subscript) denotes the gradient\footnote{On Riemannian manifolds the gradient is a vector $\b{\nabla}\mathcal{U}$ obtained by applying the  inverse metric to the covector of partial derivatives: $\Minv(\q)\frac{\partial U}{\partial \qq}$.}.
We restate here the definition of strict normal modes and a theorem from \cite{AlbuSchaeffer2021}:
\begin{definition*}[Strict Normal Mode]
Let $\mathcal{C}\subset\mathcal{Q}$ be a one-dimensional submanifold of $\mathcal{Q}$ with line segment topology.
$\mathcal{C}$ can be parametrized by the parametric curve $\gamma: [0, 1] \rightarrow \mathcal{Q}$.
If the tangent bundle $\mathcal{TC} \subset \mathcal{TQ}$ is an invariant set of the differential equation, it will be called a \emph{strict normal mode}.
\end{definition*}
\begin{theorem}\label{thm:strict_modes}
$\mathcal{C}$ is a strict normal mode of (\ref{eq:coordinate-free}) iff \begin{enumerate}
	\renewcommand{\theenumi}{\alph{enumi}}
	\item $\mathcal{C}$ is a geodesic of the Riemannian manifold $(\mathcal{Q}, g)$ and
	\item $\b{\nabla}\mathcal{U}$ is tangential to $\mathcal{C}$ everywhere on $\mathcal{C}$.
\end{enumerate}
\end{theorem}

%% file: sec/approach.tex
\section{Approach}\label{sec:approach}
Strict modes are generally isolated curves in the configuration space of a conservative mechanical system.
In this work, we aim at finding a (nonlinear) elastic potential such that strict modes lie densely in the configuration space $\mathcal{Q}$.
In other words, we require that each $\q \in \mathcal{Q}$ is touched by at least one strict mode.
As will turn out this generally cannot be achieved on arbitrary large regions of $\q \in \mathcal{Q}$, so we may need to select a region $\bar{\mathcal{Q}} \subset \mathcal{Q}$.

The goal is summarized as follows (compare also Fig.~\ref{fig:basic_scheme}):
\begin{problem*}
Find a potential function $\mathcal{U}_e$ such that a predefined region of the configuration space $\bar{\mathcal{Q}}$ containing the equilibrium $\qq_{eq}\in\bar{\mathcal{Q}}$ is densely packed with strict normal modes, i.e., every configuration $\q \in \bar{\mathcal{Q}}$ is part of at least one strict mode $\mathcal{C}\subset\bar{\mathcal{Q}}$\,.
\end{problem*}

From Thm.~\ref{thm:strict_modes}a) it is obvious that each strict mode must be a geodesic of $(\mathcal{Q}, g)$.
Therefore, each of the lines in Fig.~\ref{fig:basic_scheme} can be computed by starting at the equilibrium, choosing an initial velocity and integrating the geodesic flow.
The phrase \emph{integrating the geodesic flow} is another term for simulating the system dynamics without potential. 

When keeping the above mentioned statements in mind, we can interpret the results in \cite{AlbuSchaeffer2021} like this: integrating the geodesic flow starting from the equilibrium will create \emph{candidate strict normal modes}.
However, in the classical setting, where $\mathcal{U}$ is fixed, only a few or even no geodesics will also satisfy Thm.~\ref{thm:strict_modes}b), i.e., the choice of the potential field $\mathcal{U}$ selects the actual strict modes out of the candidates.
In this paper, we desire a potential compatible with all of the candidates.

Take the system dynamics without the potential and let us write it in coordinates \begin{equation}\label{eq:no_pot_coor}
	\MM(\qq)\qddot + \CC(\qq, \qdot)\qdot = \nnull.
\end{equation}
The approach in this paper allows to freely select the equilibrium configuration of the target system.
Let us call the desired target equilibrium $\qeq$.
Using standard integration schemes like the Runge-Kutta method the potential-free dynamics (\ref{eq:no_pot_coor}) can be integrated.
We set the system to the target equilibrium configuration $\qeq$ and shoot the system into arbitrary directions by setting the initial velocity $\qdot_0$ to arbitrary vectors.
Then, we integrate (\ref{eq:no_pot_coor}) in order to obtain trajectories $\q(t)$.
Let us parameterize $\q(t)$ with respect to arc length and call it $\ggamma: [0, l] \rightarrow \mathcal{Q}$ where $l$ is the arc length of the curve that $\q(t)$ traces out, i.e., we define a new time $s$ such that the system travels with unit-speed.
For each initial velocity $\qdot_0$ a parametrized curve $\ggamma(s)$ is obtained (Fig.~\ref{fig:basic_scheme}).
It is important to note that the curves $\ggamma(s)$ do not depend on the magnitude, but only the direction of the initial velocity $\qdot_0$.
This is a result of the definition of geodesics and the fact that the trajectories of the potential-free multi-body system are geodesics with respect to the inertia metric tensor \cite{Eisenhart1928,Arnold1989}.
Each of the geodesics $\ggamma(s)$ should correspond to a strict nonlinear normal mode of the system including the potential $U$.

Let us split the potential $U$ into a fixed and known potential $U_g$ (usually gravity) and an unknown potential $U_e$ to be determined
\begin{equation}
	U(\qq) = U_g(\qq) + U_e(\qq).
\end{equation} 
We now need to ensure that for every $\ggamma(s)$ computed according to the above scheme, the acceleration due to the generalized force $\torque = \frac{\partial U}{\partial \qq}$ is tangential to $\ggamma(s)$ for every point on $\ggamma$.
Formally, the potential $U_e(\qq)$ must satisfy
\begin{equation}\label{eq:condition}
	\Minv(\ggamma) \frac{\partial (U_g + U_e)}{\partial \qq} = \alpha(\ggamma) \dot{\ggamma},
\end{equation}
where $\alpha(\ggamma(s))$ denotes an unknown scalar function introduced to write tangentiality as equality.
Let $[\gd]^\perp$ denote the orthogonal complement of the vector $\gd \in \mathbb{R}^n$ written in matrix form, i.e., a matrix of rank $n-1$ whose columns are orthogonal to $\gd$ such that $\gd\tran [\gd]^\perp = \nnull$.
Then, the requirement (\ref{eq:condition}) can be reformulated to \begin{equation}\label{eq:condition_orth}
 	\frac{\partial (U_g + U_e)}{\partial \qq}\tran \Minv(\ggamma) [\dot\ggamma]^{\perp} = \nnull.
\end{equation}
This eliminates the unknown scalar function $\alpha$ and allows to formulate the problem as an optimization problem where (\ref{eq:condition_orth}) can be used for a cost function.

The conditions (\ref{eq:condition}) and also (\ref{eq:condition_orth}) are inherently local and do not provide any information on the overall shape and the global behavior of the final potential function.
In order to achieve a final function $U(\qq) = U_g(\qq) + U_e(\qq)$ that does behave like a potential we specify further constraints on the potential: \begin{subequations}
	\begin{align}
		\min (U_g + U_e)(\qq) &= U(\qeq) = U_0\label{cond:min}\\
		\forall \qq \ne \qeq: \mathcal{L}_{\dot\ggamma}(U_g + U_e)(\qq) &> 0.\label{cond:lie}
	\end{align}
\end{subequations}
The first constraint (\ref{cond:min}) requires that the potential has its minimum at the equilibrium $\qeq$.
The operator $\mathcal{L}_X f$ in (\ref{cond:lie}) denotes the Lie-derivative of $f$ along the vector field $X$.
By enforcing positivity of the Lie-derivative along the geodesics we ensure that the potential strictly monotonically increases when following the geodesics starting at the equilibrium.

Various methods to find a solution to (\ref{eq:condition_orth}) and the constraints (\ref{cond:min}) - (\ref{cond:lie}) are conceivable.
These also include finite element methods \cite{Braess2007} and optimization algorithms.
For the latter the constraint (\ref{eq:condition_orth}) can be relaxed and turned into an optimization objective.
Let us write the target potential $U_e(\qq)$ as parametric function $U_{e}(\oomega, \qq)$, where $\oomega$ is a collection of all the parameters of the function.
Then training and test data can be generated by computing geodesics according to the proposed scheme.
Assume that a total of $N$ geodesics have been computed and let the $i$-th geodesic $\ggamma_i$ be expressed by $T_i$ samples.
The $t$-th sample of the $i$-th geodesic consist of a position $\ggamma_{it}$ and a tangent $\dot\ggamma_{it}$.
We then write the cost function \begin{equation}\label{eq:cost}
	L(\oomega) = \sum\limits_{i=1}^N \sum\limits_{t=1}^{T_i} \left\Vert\frac{\partial (U_g + U_e)}{\partial \q}\tran\Minv(\ggamma_{it}) [\dot\ggamma_{it}]^\perp\right\Vert^2.
\end{equation}
In combination with (\ref{cond:min}b) the cost function (\ref{eq:cost}) yields a constrained optimization problem.

Also for the choice of the parametric model $U_e(\oomega, \qq)$ a large palette of possibilities is available \cite{Murphy2022,Bishop2006}.

One choice would be using linear regression methods with basis functions.
Each of the basis functions can be chosen such that they are realizable by an elastic element.
Assume a selection of nonlinear elastic elements are available - \cite{Vanderborght2013} and \cite{Wolf2016} review a broad set of methods to design such elements.

For instance, given linear and cubic springs a lot of functions can already be implemented.
Besides the obvious direct connection of a springs to a joint, the springs can, among other possibilities, be attached via crankshaft mechanisms implementing a sine-like force profile.
Additionally, when using one-side mounted compression springs non-linear \textsc{ReLU}-like behaviour can be implemented, which can also be combined with other mechanisms.

Coupling terms could be introduced by constructing mechanical components that compute sums or differences of adjacent joints, e.g., a shaft and two pairs of bevel gears combined with an elastic element provide coupled opposing torques in two adjacent joints.
These mechanical implementations create a selection of possible potential functions that can be used as basis functions for the regression.
This way the resulting potential can be implemented on the physical system by tuning the stiffnesses of the springs according to the optimized parameters.
In order to achieve physically realizable elements, a non-negative least squares solver \cite{Lawson1995} must be used.
Another advantage of non-negative least squares is sparsity, i.e., one can remove all the elastic elements not creating any force from the system.

However, due to lack of space, the physical implementation of the optimized potential is not in focus in this paper and it will be addressed in future work.
We, therefore, here choose to employ a neural network model with a single output for the parametric function $U_e(\oomega, \qq)$.

When implementing the potential $U_e$ on the mechanical system it will provide infinitely many strict modes.
For conservative systems the modes will neither be attractive nor repulsive.
External forces and modelling errors result in the system not stably oscillating on a desired mode.
Also, when starting at rest the modal oscillation of the system must be excited.
Friction and other dissipative terms remove energy from the system which must be reinjected.

For these reasons a controller is required for practical applications.
The controller will stabilize the system onto the desired mode and keep the total energy at a desired level.
Let $\ttau_c(\q, \qdot)$ be that controller, then the dynamics of the system can, in coordinates, be written as \begin{equation}\label{eq:dynamics_all}
	\underbrace{\MM(\qq) \qddot + \CC(\qq, \qdot)\qdot + \frac{\partial U_g}{\partial \qq}}_{\text{conservative mechanical system}} + \underbrace{\dd(\q, \qdot)}_{\substack{\text{damping}\\\text{and friction}}} + \underbrace{\frac{\partial U_e}{\partial \qq}}_{\substack{\text{elas. pot.}\\\text{(passive)}}} = \underbrace{\ttau_c(\qq, \qdot)}_{\substack{\text{controller}\\\text{(active)}}}.\nonumber
\end{equation}
A classical robotic system would generate the sum of the elastic potential and the active controller using the actuators.
The main advantage of the proposed approach is that torques due to the potential $U_e$ are large compared to the controller torques and that $U_e$ creates \emph{useful} torques, i.e., torques steering the system on a desired natural oscillation.
Because the potential $U_e$ can be implemented by (nonlinear) elastic elements on the physical system, only the torques due to $\ttau_c(\qq, \qdot)$ need to be generated by the actuators.
Therefore, a large part of required torques will be provided by the mechanics requiring smaller actuators and less actuator power.
This allows to perform all of the modal oscillations and desired trajectories that are close to the modes very efficiently.

%% file: sec/casestudy.tex
\section{Case Study: Double Pendulum}\label{sec:casestudy}

\begin{figure}
	\centering
	\def\svgwidth{.3\linewidth}
	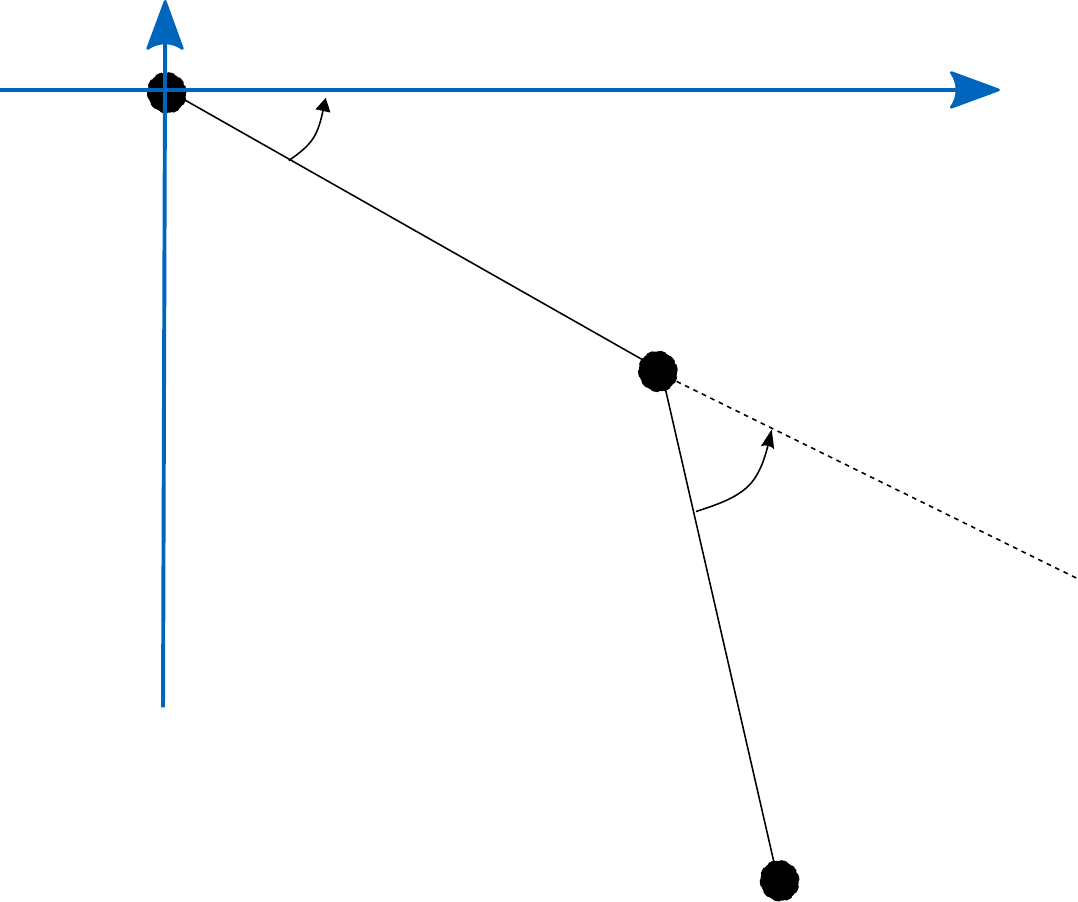
	\caption{Double Pendulum}
	\label{fig:example_system}
\end{figure}

We showcase the approach for a serial system with two degrees of freedom (Fig.~\ref{fig:example_system}).
Further, we take the system outside of a gravitational field ($U_g = 0$).
You can either think of a double pendulum in space or a two degree of freedom manipulator on earth parallel to the earth's surface.
The system parameters are summarized in Tab.~\ref{tab:pendulum_parameters} in App.~\ref{app:pendulum_parameters}.
The potential-free dynamics is governed by (\ref{eq:no_pot_coor}).

An equilibrium $\qeq$ for the target system must be selected.
As derived before, the choice of equilibrium and the inertia tensor fully determines the shape of the strict normal modes of the target system.
Therefore, the equilibrium configuration should be carefully selected, as it will determine the strict normal modes that can later be used.

\begin{figure}[h]
	\vspace{-.5cm}
	\centering
	\subfloat[Long Geodesics\label{fig:unclipped_geodesics_boomerang}]{
		\def\svgwidth{.49\columnwidth}\tiny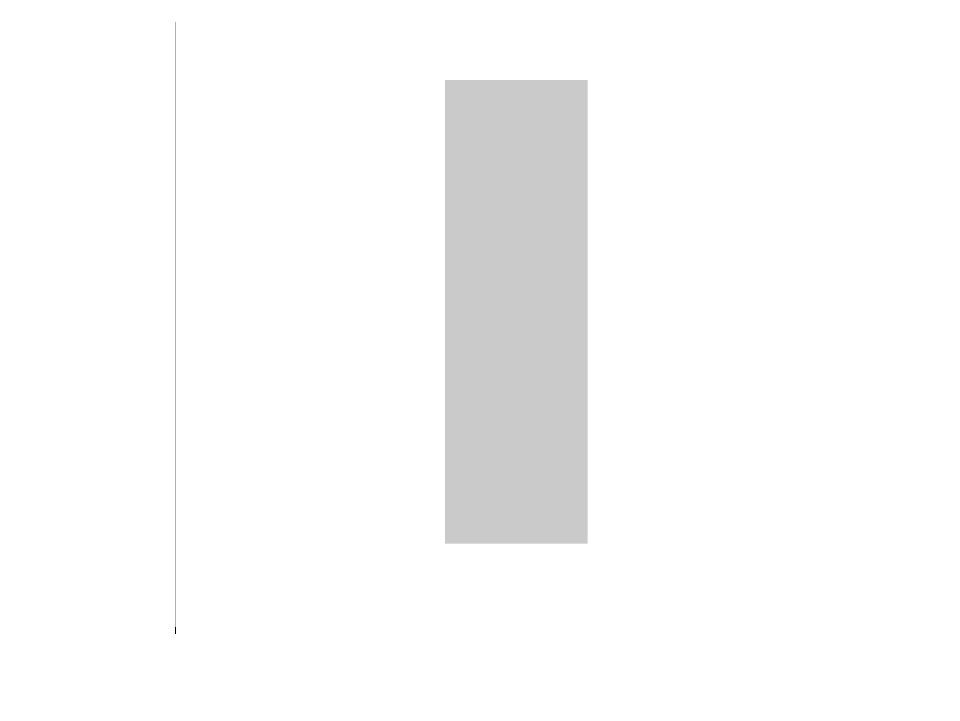
	}
	\subfloat[Clipped to $\bar{\mathcal{Q}}$\label{fig:clipped_geodesics_boomerang}]{
		\def\svgwidth{.49\columnwidth}\tiny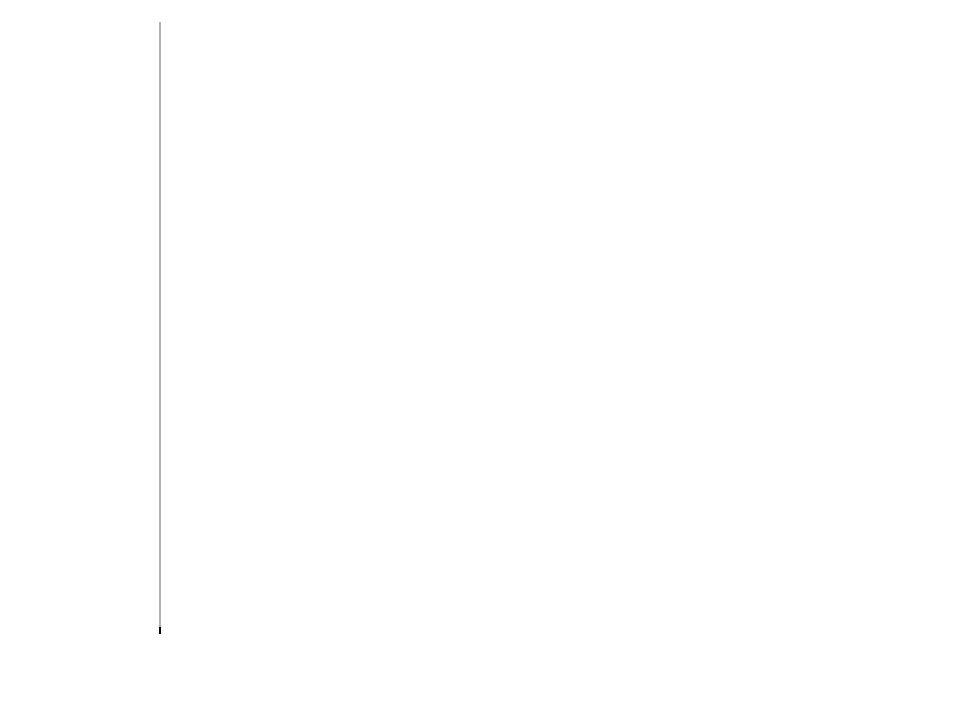
	}
	\caption{
		Geodesics of the example system. 
		Restriction to a subregion results in a space where the geodesics only intersect at the equilibrium.
	}
	\label{fig:pure_geodesics}
\end{figure}

We start with the choice $\qeq = [-\nicefrac{\pi}{6}, -\nicefrac{\pi}{4}]\tran$ (Fig.~\ref{fig:example_system}).
Using an Runge-Kutta integrator \cite{Hairer2008}, we compute geodesics of the system by the proposed method.
We set the initial configuration to $\q_0 = \qeq$ and set an upper time limit in order to stop the integration.
The initial velocity $\qdot_0$ is set to unit vectors by sampling the unit-sphere equidistantly.
Due to the nonlinearity, distributing the initial velocities uniformly does not imply that the configuration space is evenly covered with geodesics. 
However, for every point $\qq$ in the subspace $\bar{\mathcal{Q}}$ a geodesic passing through it and the equilibrium can be found.
Fig.~\ref{fig:unclipped_geodesics_boomerang} shows some of the computed geodesics.
Note that the axes are in radians, so the trajectories correspond to long motions, where the joints are turned multiple times.
In this large region $\mathcal{Q}$ the geodesics intersect, but when selecting a smaller region the intersections can be avoided.
The gray rectangle in Fig.~\ref{fig:unclipped_geodesics_boomerang} shows one such region $\bar{\mathcal{Q}}$ and Fig.~\ref{fig:clipped_geodesics_boomerang} shows it magnified.

We perform this procedure for three different settings for the equilibrium $\qeq$ and show resulting geodesics in Tab.~\ref{tab:comparison_different_equilibria}.
The first column in Tab.~\ref{tab:comparison_different_equilibria} shows the equilibrium configuration, the second column the geodesics in a properly selected region $\bar{\mathcal{Q}}$ of the configuration space and the rightmost column shows the geodesics in Cartesian space.
Using the color of the lines, the geodesics in configuration space can be matched to their Cartesian reflection.
At this point we will continue the example with the elbow-up setting (row (b) in Tab.~\ref{tab:comparison_different_equilibria}) for the remainder of this section.

\input{sec/equilibrium_examples}

We choose a two-layer neural network with $\tanh$-activation functions and $200$ neurons in the first and $100$ neurons in the second layer as model function for $U_e(\oomega, \qq)$.
The network model is implemented in TensorFlow \cite{Abadi2015a} and the parameters are initialized randomly.
In general we formulated the objective of finding the potential as constrained optimization problem in (\ref{cond:min})-(\ref{eq:cost}).
The cost function is only dependent on the output-to-input gradient of the neural network, not on the output itself.
It is tuning the network very locally, but does not have a strong preference in the global shape of the potential.
Therefore, we do not need to enforce the constraints (\ref{cond:min}b) explicitly, but rather pre-train the network to a convex surface and then switch to the cost (\ref{eq:cost}) to warp and stretch the initially trained surface to match the desired tangentiality condition (\ref{eq:cost}).

In a first step the network is fitted to a parabola \begin{equation}\label{eq:parabola}
	V(\qq) = (\qq -\qeq)\tran (\qq - \qeq)
\end{equation}
using an $L_2$-loss based on (\ref{eq:parabola}).
The parameters $\oomega$ are optimized using the \textsc{Adam} algorithm \cite{Kingma2017}.
A parabola is easy to fit, therefore we obtain a very low pre-training loss.

Afterwards, we switch to the loss function (\ref{eq:cost}).
During the training we frequently sample new training data in order to prevent overfitting.
Although we have not explicitly enforced the constraints (\ref{cond:min}b), the local nature of the loss function keeps the global bowl-like shape of the pre-trained parabola during the training.

\begin{figure}[h]
	\centering
	\def\svgwidth{.8\linewidth}\tiny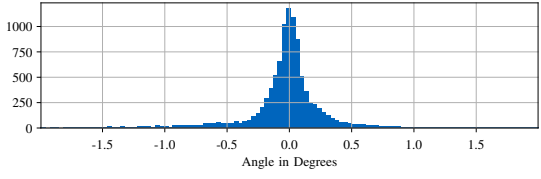
	\caption{Residuals}\label{subfig:hist}
\end{figure}

\begin{figure}[h]
	\centering
	\subfloat[Surface Plot\label{subfig:surface_potential}]{
		\def\svgwidth{.34\columnwidth}\tiny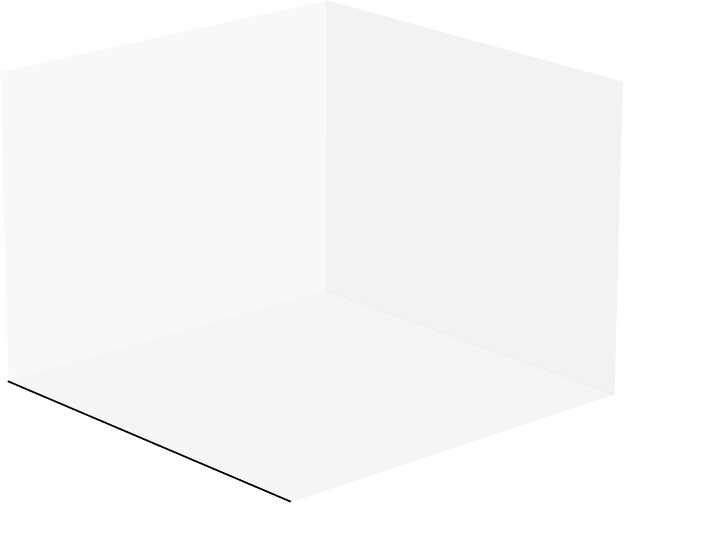
	}
	\subfloat[$q_1 q_2$-Plane\label{subfig:pp_phase}]{
		\def\svgwidth{.68\columnwidth}\tiny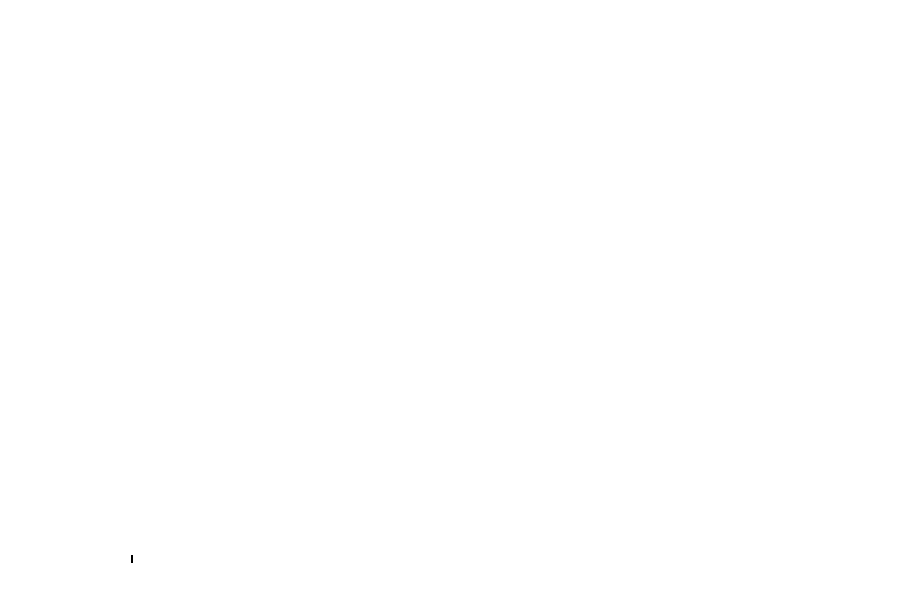
	}\\
	\subfloat[Trajectory 3\label{subfig:pp_q_3}]{
		\def\svgwidth{.32\columnwidth}\tiny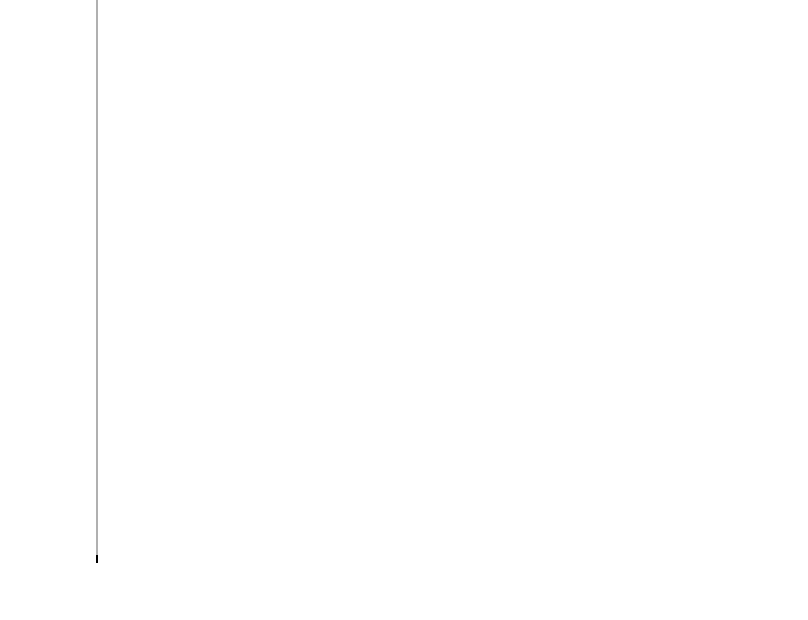
	}
	\subfloat[Trajectory 5\label{subfig:pp_q_5}]{
		\def\svgwidth{.32\columnwidth}\tiny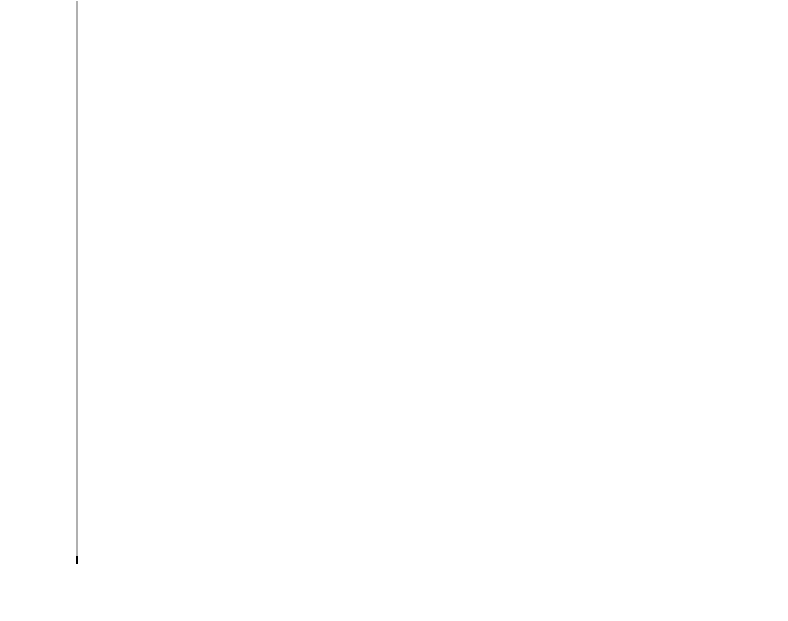
	}
	\subfloat[Trajectory 6\label{subfig:pp_q_6}]{
		\def\svgwidth{.32\columnwidth}\tiny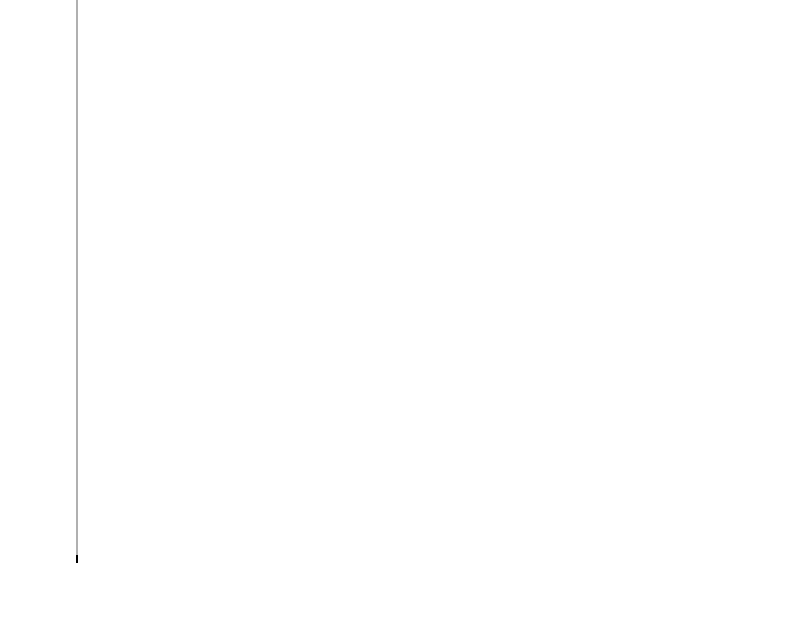
	}
	\caption{Simulation of the double pendulum with the optimized potential for different initial conditions and visualization of the potential function.}
	\label{fig:pp_results}
\end{figure}

In Fig.~\ref{subfig:pp_phase}, isolines of the resulting potential are shown in orange and Fig.~\ref{subfig:surface_potential} shows a surface plot.
Additionally, the gray lines show some geodesics.
We analyze how parallel $\vv_1 = \Minv(\qq)\frac{\partial U}{\partial \qq}$ and $\vv_2 = \dot{\ggamma}$ are.
Therefore, we compute the angle between the vectors $\vv_1$ and $\vv_2$ for many points on randomly sampled geodesics and show a histogram in Fig.~\ref{subfig:hist}.
The angles are computed in degrees, where an angle of $0^\circ$ is optimal in a sense that $\vv_1$ and $\vv_2$ are parallel.

For validation of the potential $U_e$ we apply the potential on the system dynamics, i.e., we simulate \begin{equation}
	\MM(\qq)\qddot + \CC(\qq, \qdot)\qdot + \frac{\partial U_e}{\partial \qq}= \nnull
\end{equation}
for some initial configurations and show the results in Fig.~\ref{fig:pp_results}.
The initial velocity is always set to zero.
Panel (b) shows the simulated trajectories in the $q_1 q_2$-plane.
The black stars show the initial configurations.
Moreover, we show time plots for three of the six trajectories (c-e).
We would like to highlight that the results in Fig.~\ref{fig:pp_results} are without any active controller.
The periodic motions are due to the properly tuned potential $U_e$.

The results in Fig.~\ref{fig:pp_results} confirm that the proposed approach leads to the desired result.
We have shown six example modes, but infinitely many are present in the system.
There is no preferred mode.
Hence, when the system is slightly disturbed onto another mode, there is no mechanism pulling the same back onto the desired mode.
When the system is subject to damping, the damping will dissipate energy from the system and the oscillation will decay.
Moreover, the damping will most likely not be compatible with the mode, i.e., the damping forces may generally pull the system off the mode.

Hence, we design a controller stabilizing the system onto the desired mode and a controller injecting energy in the next section.

%% file: pdftex/pendulum.pdf_tex
\begingroup%
  \makeatletter%
  \providecommand\color[2][]{%
    \errmessage{(Inkscape) Color is used for the text in Inkscape, but the package 'color.sty' is not loaded}%
    \renewcommand\color[2][]{}%
  }%
  \providecommand\transparent[1]{%
    \errmessage{(Inkscape) Transparency is used (non-zero) for the text in Inkscape, but the package 'transparent.sty' is not loaded}%
    \renewcommand\transparent[1]{}%
  }%
  \providecommand\rotatebox[2]{#2}%
  \newcommand*\fsize{\dimexpr\f@size pt\relax}%
  \newcommand*\lineheight[1]{\fontsize{\fsize}{#1\fsize}\selectfont}%
  \ifx\svgwidth\undefined%
    \setlength{\unitlength}{517.28533407bp}%
    \ifx\svgscale\undefined%
      \relax%
    \else%
      \setlength{\unitlength}{\unitlength * \real{\svgscale}}%
    \fi%
  \else%
    \setlength{\unitlength}{\svgwidth}%
  \fi%
  \global\let\svgwidth\undefined%
  \global\let\svgscale\undefined%
  \makeatother%
  \begin{picture}(1,0.83651175)%
    \lineheight{1}%
    \setlength\tabcolsep{0pt}%
    \put(0,0){\includegraphics[width=\unitlength,page=1]{pendulum.pdf}}%
    \put(0.88235357,0.78698762){\color[rgb]{0,0,0}\makebox(0,0)[lt]{\lineheight{1.25}\smash{\begin{tabular}[t]{l}$x$\end{tabular}}}}%
    \put(0.02485298,0.77810339){\color[rgb]{0,0,0}\makebox(0,0)[lt]{\lineheight{1.25}\smash{\begin{tabular}[t]{l}$y$\end{tabular}}}}%
    \put(0.21126769,0.78054083){\color[rgb]{0,0,0}\makebox(0,0)[lt]{\lineheight{1.25}\smash{\begin{tabular}[t]{l}$q_1$\end{tabular}}}}%
    \put(0.67644202,0.48226356){\color[rgb]{0,0,0}\makebox(0,0)[lt]{\lineheight{1.25}\smash{\begin{tabular}[t]{l}$q_2$\end{tabular}}}}%
  \end{picture}%
\endgroup%

%% file: pdftex/unclipped_boomerang.pdf_tex
\begingroup%
  \makeatletter%
  \providecommand\color[2][]{%
    \errmessage{(Inkscape) Color is used for the text in Inkscape, but the package 'color.sty' is not loaded}%
    \renewcommand\color[2][]{}%
  }%
  \providecommand\transparent[1]{%
    \errmessage{(Inkscape) Transparency is used (non-zero) for the text in Inkscape, but the package 'transparent.sty' is not loaded}%
    \renewcommand\transparent[1]{}%
  }%
  \providecommand\rotatebox[2]{#2}%
  \newcommand*\fsize{\dimexpr\f@size pt\relax}%
  \newcommand*\lineheight[1]{\fontsize{\fsize}{#1\fsize}\selectfont}%
  \ifx\svgwidth\undefined%
    \setlength{\unitlength}{460.79998779bp}%
    \ifx\svgscale\undefined%
      \relax%
    \else%
      \setlength{\unitlength}{\unitlength * \real{\svgscale}}%
    \fi%
  \else%
    \setlength{\unitlength}{\svgwidth}%
  \fi%
  \global\let\svgwidth\undefined%
  \global\let\svgscale\undefined%
  \makeatother%
  \begin{picture}(1,0.75000003)%
    \lineheight{1}%
    \setlength\tabcolsep{0pt}%
    \put(0,0){\includegraphics[width=\unitlength,page=1]{unclipped_boomerang.pdf}}%
    \put(0.18263804,0.05959816){\makebox(0,0)[t]{\lineheight{1.25}\smash{\begin{tabular}[t]{c}−10.0\end{tabular}}}}%
    \put(0,0){\includegraphics[width=\unitlength,page=2]{unclipped_boomerang.pdf}}%
    \put(0.28199849,0.05959816){\makebox(0,0)[t]{\lineheight{1.25}\smash{\begin{tabular}[t]{c}−7.5\end{tabular}}}}%
    \put(0,0){\includegraphics[width=\unitlength,page=3]{unclipped_boomerang.pdf}}%
    \put(0.38135897,0.05959816){\makebox(0,0)[t]{\lineheight{1.25}\smash{\begin{tabular}[t]{c}−5.0\end{tabular}}}}%
    \put(0,0){\includegraphics[width=\unitlength,page=4]{unclipped_boomerang.pdf}}%
    \put(0.48071942,0.05959816){\makebox(0,0)[t]{\lineheight{1.25}\smash{\begin{tabular}[t]{c}−2.5\end{tabular}}}}%
    \put(0,0){\includegraphics[width=\unitlength,page=5]{unclipped_boomerang.pdf}}%
    \put(0.5800799,0.05959816){\makebox(0,0)[t]{\lineheight{1.25}\smash{\begin{tabular}[t]{c}0.0\end{tabular}}}}%
    \put(0,0){\includegraphics[width=\unitlength,page=6]{unclipped_boomerang.pdf}}%
    \put(0.67944032,0.05959816){\makebox(0,0)[t]{\lineheight{1.25}\smash{\begin{tabular}[t]{c}2.5\end{tabular}}}}%
    \put(0,0){\includegraphics[width=\unitlength,page=7]{unclipped_boomerang.pdf}}%
    \put(0.7788008,0.05959816){\makebox(0,0)[t]{\lineheight{1.25}\smash{\begin{tabular}[t]{c}5.0\end{tabular}}}}%
    \put(0,0){\includegraphics[width=\unitlength,page=8]{unclipped_boomerang.pdf}}%
    \put(0.87816128,0.05959816){\makebox(0,0)[t]{\lineheight{1.25}\smash{\begin{tabular}[t]{c}7.5\end{tabular}}}}%
    \put(0.53899876,0.02991471){\makebox(0,0)[t]{\lineheight{1.25}\smash{\begin{tabular}[t]{c}$q_1$ in rad\end{tabular}}}}%
    \put(0,0){\includegraphics[width=\unitlength,page=9]{unclipped_boomerang.pdf}}%
    \put(0.08624403,0.13748971){\makebox(0,0)[rt]{\lineheight{1.25}\smash{\begin{tabular}[t]{r}−8\end{tabular}}}}%
    \put(0,0){\includegraphics[width=\unitlength,page=10]{unclipped_boomerang.pdf}}%
    \put(0.08624403,0.20857441){\makebox(0,0)[rt]{\lineheight{1.25}\smash{\begin{tabular}[t]{r}−6\end{tabular}}}}%
    \put(0,0){\includegraphics[width=\unitlength,page=11]{unclipped_boomerang.pdf}}%
    \put(0.08624403,0.27965911){\makebox(0,0)[rt]{\lineheight{1.25}\smash{\begin{tabular}[t]{r}−4\end{tabular}}}}%
    \put(0,0){\includegraphics[width=\unitlength,page=12]{unclipped_boomerang.pdf}}%
    \put(0.08624403,0.35074381){\makebox(0,0)[rt]{\lineheight{1.25}\smash{\begin{tabular}[t]{r}−2\end{tabular}}}}%
    \put(0,0){\includegraphics[width=\unitlength,page=13]{unclipped_boomerang.pdf}}%
    \put(0.08624403,0.42182847){\makebox(0,0)[rt]{\lineheight{1.25}\smash{\begin{tabular}[t]{r}0\end{tabular}}}}%
    \put(0,0){\includegraphics[width=\unitlength,page=14]{unclipped_boomerang.pdf}}%
    \put(0.08624403,0.49291317){\makebox(0,0)[rt]{\lineheight{1.25}\smash{\begin{tabular}[t]{r}2\end{tabular}}}}%
    \put(0,0){\includegraphics[width=\unitlength,page=15]{unclipped_boomerang.pdf}}%
    \put(0.08624403,0.56399787){\makebox(0,0)[rt]{\lineheight{1.25}\smash{\begin{tabular}[t]{r}4\end{tabular}}}}%
    \put(0,0){\includegraphics[width=\unitlength,page=16]{unclipped_boomerang.pdf}}%
    \put(0.08624403,0.63508256){\makebox(0,0)[rt]{\lineheight{1.25}\smash{\begin{tabular}[t]{r}6\end{tabular}}}}%
    \put(0,0){\includegraphics[width=\unitlength,page=17]{unclipped_boomerang.pdf}}%
    \put(0.08624403,0.70616725){\makebox(0,0)[rt]{\lineheight{1.25}\smash{\begin{tabular}[t]{r}8\end{tabular}}}}%
    \put(0.04045411,0.41163334){\rotatebox{90}{\makebox(0,0)[t]{\lineheight{1.25}\smash{\begin{tabular}[t]{c}$q_2$ in rad\end{tabular}}}}}%
    \put(0,0){\includegraphics[width=\unitlength,page=18]{unclipped_boomerang.pdf}}%
  \end{picture}%
\endgroup%

%% file: pdftex/clipped_boomerang.pdf_tex
\begingroup%
  \makeatletter%
  \providecommand\color[2][]{%
    \errmessage{(Inkscape) Color is used for the text in Inkscape, but the package 'color.sty' is not loaded}%
    \renewcommand\color[2][]{}%
  }%
  \providecommand\transparent[1]{%
    \errmessage{(Inkscape) Transparency is used (non-zero) for the text in Inkscape, but the package 'transparent.sty' is not loaded}%
    \renewcommand\transparent[1]{}%
  }%
  \providecommand\rotatebox[2]{#2}%
  \newcommand*\fsize{\dimexpr\f@size pt\relax}%
  \newcommand*\lineheight[1]{\fontsize{\fsize}{#1\fsize}\selectfont}%
  \ifx\svgwidth\undefined%
    \setlength{\unitlength}{460.79998779bp}%
    \ifx\svgscale\undefined%
      \relax%
    \else%
      \setlength{\unitlength}{\unitlength * \real{\svgscale}}%
    \fi%
  \else%
    \setlength{\unitlength}{\svgwidth}%
  \fi%
  \global\let\svgwidth\undefined%
  \global\let\svgscale\undefined%
  \makeatother%
  \begin{picture}(1,0.75000003)%
    \lineheight{1}%
    \setlength\tabcolsep{0pt}%
    \put(0,0){\includegraphics[width=\unitlength,page=1]{clipped_boomerang.pdf}}%
    \put(0.16690238,0.05959816){\makebox(0,0)[t]{\lineheight{1.25}\smash{\begin{tabular}[t]{c}−2.5\end{tabular}}}}%
    \put(0,0){\includegraphics[width=\unitlength,page=2]{clipped_boomerang.pdf}}%
    \put(0.28000567,0.05959816){\makebox(0,0)[t]{\lineheight{1.25}\smash{\begin{tabular}[t]{c}−2.0\end{tabular}}}}%
    \put(0,0){\includegraphics[width=\unitlength,page=3]{clipped_boomerang.pdf}}%
    \put(0.393109,0.05959816){\makebox(0,0)[t]{\lineheight{1.25}\smash{\begin{tabular}[t]{c}−1.5\end{tabular}}}}%
    \put(0,0){\includegraphics[width=\unitlength,page=4]{clipped_boomerang.pdf}}%
    \put(0.50621229,0.05959816){\makebox(0,0)[t]{\lineheight{1.25}\smash{\begin{tabular}[t]{c}−1.0\end{tabular}}}}%
    \put(0,0){\includegraphics[width=\unitlength,page=5]{clipped_boomerang.pdf}}%
    \put(0.61931559,0.05959816){\makebox(0,0)[t]{\lineheight{1.25}\smash{\begin{tabular}[t]{c}−0.5\end{tabular}}}}%
    \put(0,0){\includegraphics[width=\unitlength,page=6]{clipped_boomerang.pdf}}%
    \put(0.73241891,0.05959816){\makebox(0,0)[t]{\lineheight{1.25}\smash{\begin{tabular}[t]{c}0.0\end{tabular}}}}%
    \put(0,0){\includegraphics[width=\unitlength,page=7]{clipped_boomerang.pdf}}%
    \put(0.84552217,0.05959816){\makebox(0,0)[t]{\lineheight{1.25}\smash{\begin{tabular}[t]{c}0.5\end{tabular}}}}%
    \put(0,0){\includegraphics[width=\unitlength,page=8]{clipped_boomerang.pdf}}%
    \put(0.9586255,0.05959816){\makebox(0,0)[t]{\lineheight{1.25}\smash{\begin{tabular}[t]{c}1.0\end{tabular}}}}%
    \put(0.53320886,0.02991471){\makebox(0,0)[t]{\lineheight{1.25}\smash{\begin{tabular}[t]{c}$q_1$ in rad\end{tabular}}}}%
    \put(0,0){\includegraphics[width=\unitlength,page=9]{clipped_boomerang.pdf}}%
    \put(0.08624403,0.13359415){\makebox(0,0)[rt]{\lineheight{1.25}\smash{\begin{tabular}[t]{r}−6\end{tabular}}}}%
    \put(0,0){\includegraphics[width=\unitlength,page=10]{clipped_boomerang.pdf}}%
    \put(0.08624403,0.22363547){\makebox(0,0)[rt]{\lineheight{1.25}\smash{\begin{tabular}[t]{r}−4\end{tabular}}}}%
    \put(0,0){\includegraphics[width=\unitlength,page=11]{clipped_boomerang.pdf}}%
    \put(0.08624403,0.31367677){\makebox(0,0)[rt]{\lineheight{1.25}\smash{\begin{tabular}[t]{r}−2\end{tabular}}}}%
    \put(0,0){\includegraphics[width=\unitlength,page=12]{clipped_boomerang.pdf}}%
    \put(0.08624403,0.40371806){\makebox(0,0)[rt]{\lineheight{1.25}\smash{\begin{tabular}[t]{r}0\end{tabular}}}}%
    \put(0,0){\includegraphics[width=\unitlength,page=13]{clipped_boomerang.pdf}}%
    \put(0.08624403,0.49375938){\makebox(0,0)[rt]{\lineheight{1.25}\smash{\begin{tabular}[t]{r}2\end{tabular}}}}%
    \put(0,0){\includegraphics[width=\unitlength,page=14]{clipped_boomerang.pdf}}%
    \put(0.08624403,0.58380069){\makebox(0,0)[rt]{\lineheight{1.25}\smash{\begin{tabular}[t]{r}4\end{tabular}}}}%
    \put(0,0){\includegraphics[width=\unitlength,page=15]{clipped_boomerang.pdf}}%
    \put(0.08624403,0.673842){\makebox(0,0)[rt]{\lineheight{1.25}\smash{\begin{tabular}[t]{r}6\end{tabular}}}}%
    \put(0.0404541,0.41163333){\rotatebox{90}{\makebox(0,0)[t]{\lineheight{1.25}\smash{\begin{tabular}[t]{c}$q_2$ in rad\end{tabular}}}}}%
    \put(0,0){\includegraphics[width=\unitlength,page=16]{clipped_boomerang.pdf}}%
  \end{picture}%
\endgroup%

%% file: sec/equilibrium_examples.tex
\def\equiwidth{.2\columnwidth}
\def\geowidth{.28\columnwidth}
\def\cartwidth{.28\columnwidth}
\def\eqcaption{Equilibrium}
\def\geocaption{Geodesics}
\def\cartgeocaption{Cartesian Space}
\begin{table}
	\caption{Comparison of different target equilibria.}
	\label{tab:comparison_different_equilibria}
	\centering
	\begin{tabular}{ccc}
		\textbf{Equilibrium}&
		\textbf{Some Geodesics}&
		\textbf{Cartesian Space}\\\hline

		(a)
		\subfloat{\def\svgwidth{\equiwidth}\tiny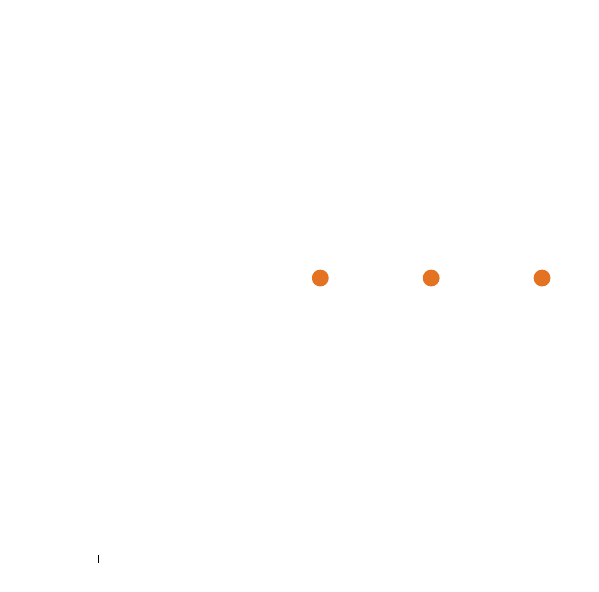}&
		\subfloat{\def\svgwidth{\geowidth}\tiny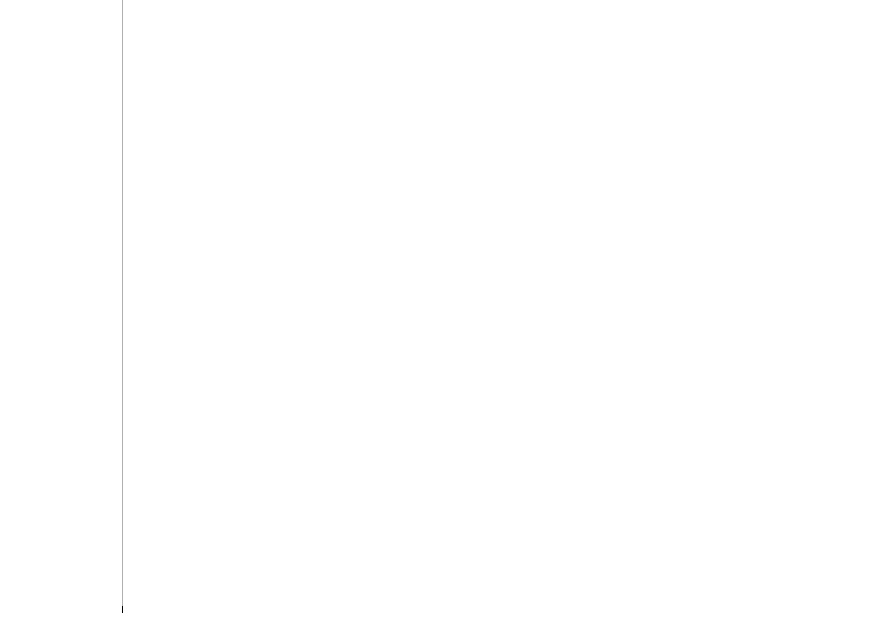}&
		\subfloat{\def\svgwidth{\cartwidth}\tiny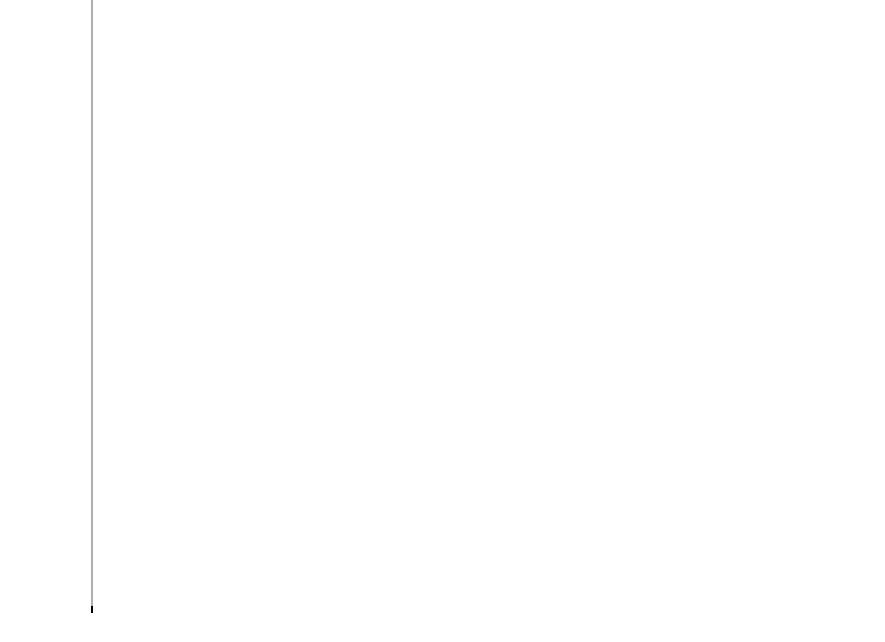}\\

		(b)
		\subfloat{\def\svgwidth{\equiwidth}\tiny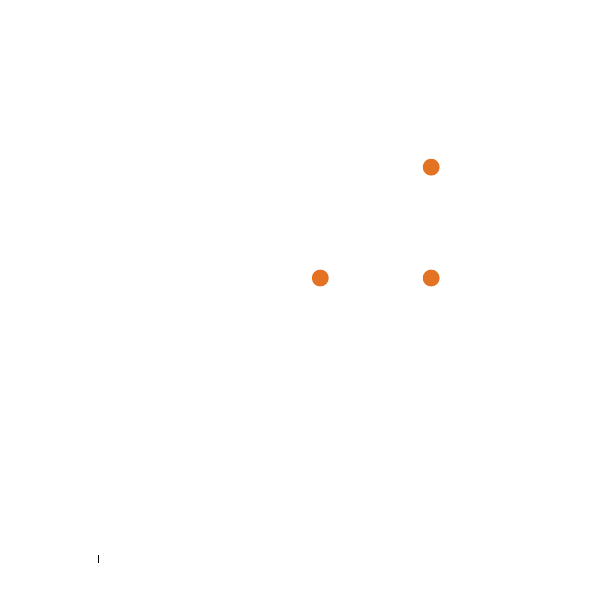}&
		\subfloat{\def\svgwidth{\geowidth}\tiny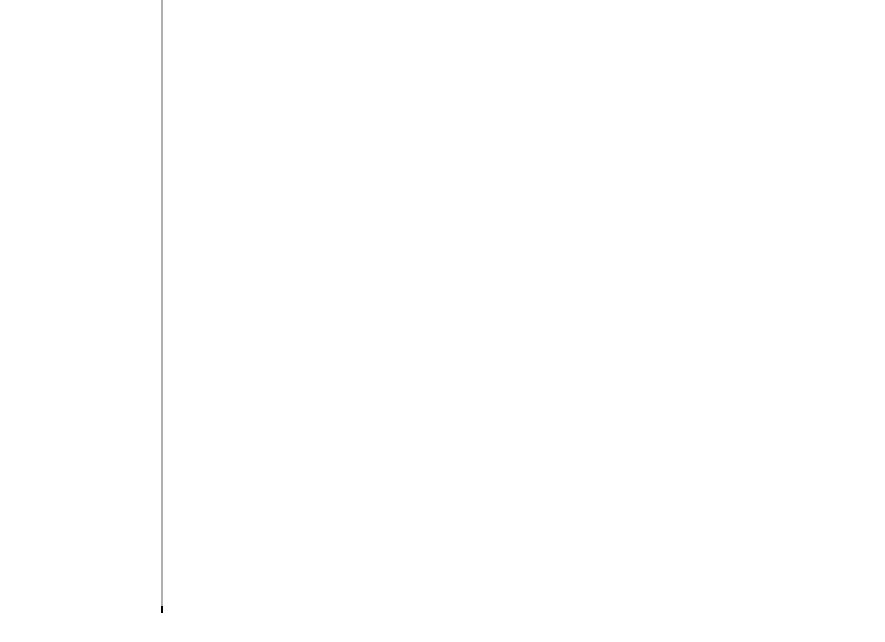}&
		\subfloat{\def\svgwidth{\cartwidth}\tiny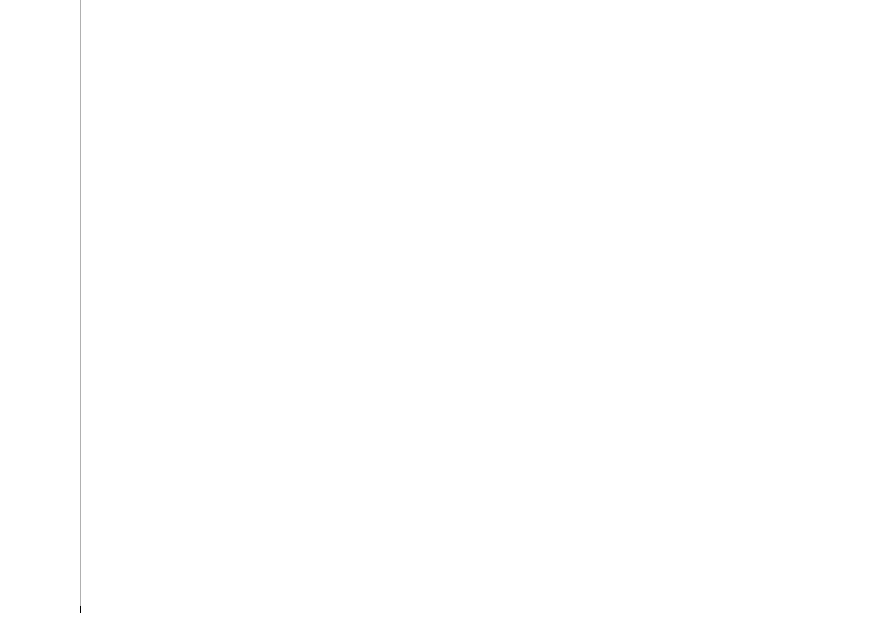}\\

		(c)
		\subfloat{\def\svgwidth{\equiwidth}\tiny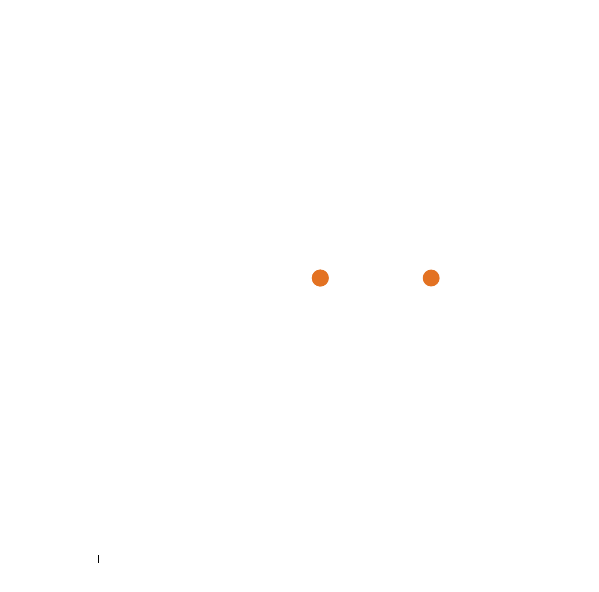}&
		\subfloat{\def\svgwidth{\geowidth}\tiny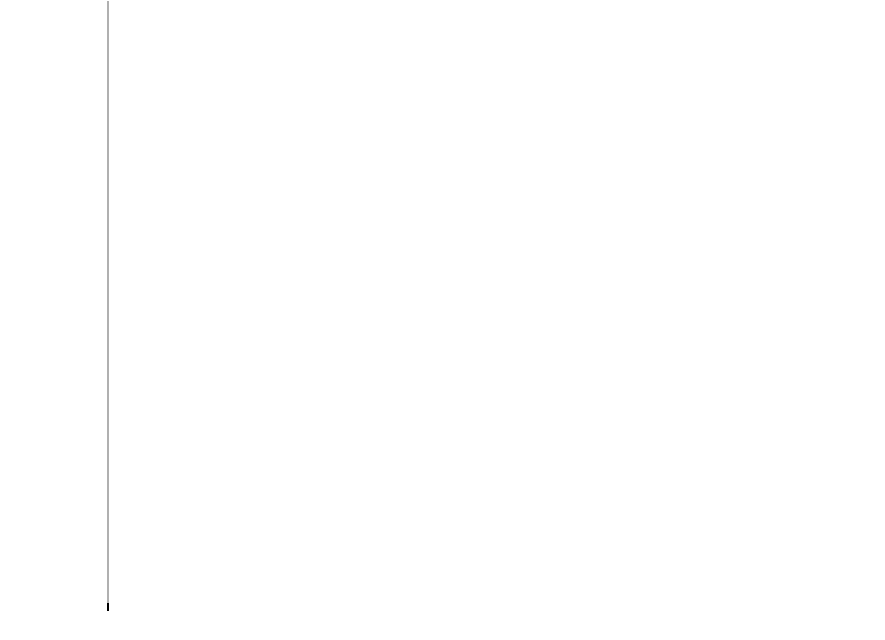}&
		\subfloat{\def\svgwidth{\cartwidth}\tiny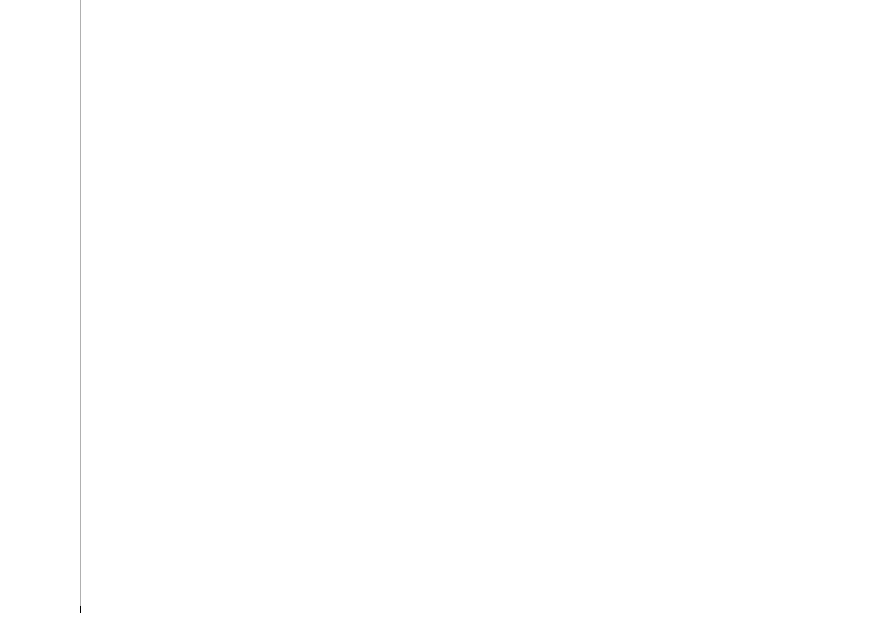}\\
	\end{tabular}
	\vspace{-.5cm}
\end{table}

%% file: pdftex/0_equib.pdf_tex
\begingroup%
  \makeatletter%
  \providecommand\color[2][]{%
    \errmessage{(Inkscape) Color is used for the text in Inkscape, but the package 'color.sty' is not loaded}%
    \renewcommand\color[2][]{}%
  }%
  \providecommand\transparent[1]{%
    \errmessage{(Inkscape) Transparency is used (non-zero) for the text in Inkscape, but the package 'transparent.sty' is not loaded}%
    \renewcommand\transparent[1]{}%
  }%
  \providecommand\rotatebox[2]{#2}%
  \newcommand*\fsize{\dimexpr\f@size pt\relax}%
  \newcommand*\lineheight[1]{\fontsize{\fsize}{#1\fsize}\selectfont}%
  \ifx\svgwidth\undefined%
    \setlength{\unitlength}{287.19363403bp}%
    \ifx\svgscale\undefined%
      \relax%
    \else%
      \setlength{\unitlength}{\unitlength * \real{\svgscale}}%
    \fi%
  \else%
    \setlength{\unitlength}{\svgwidth}%
  \fi%
  \global\let\svgwidth\undefined%
  \global\let\svgscale\undefined%
  \makeatother%
  \begin{picture}(1,0.98801648)%
    \lineheight{1}%
    \setlength\tabcolsep{0pt}%
    \put(0,0){\includegraphics[width=\unitlength,page=1]{0_equib.pdf}}%
    \put(0.16467231,0.00049305){\makebox(0,0)[t]{\lineheight{1.25}\smash{\begin{tabular}[t]{c}−4\end{tabular}}}}%
    \put(0,0){\includegraphics[width=\unitlength,page=2]{0_equib.pdf}}%
    \put(0.34999122,0.00049305){\makebox(0,0)[t]{\lineheight{1.25}\smash{\begin{tabular}[t]{c}−2\end{tabular}}}}%
    \put(0,0){\includegraphics[width=\unitlength,page=3]{0_equib.pdf}}%
    \put(0.53531006,0.00049305){\makebox(0,0)[t]{\lineheight{1.25}\smash{\begin{tabular}[t]{c}0\end{tabular}}}}%
    \put(0,0){\includegraphics[width=\unitlength,page=4]{0_equib.pdf}}%
    \put(0.72062886,0.00049305){\makebox(0,0)[t]{\lineheight{1.25}\smash{\begin{tabular}[t]{c}2\end{tabular}}}}%
    \put(0,0){\includegraphics[width=\unitlength,page=5]{0_equib.pdf}}%
    \put(0.90594778,0.00049305){\makebox(0,0)[t]{\lineheight{1.25}\smash{\begin{tabular}[t]{c}4\end{tabular}}}}%
    \put(0,0){\includegraphics[width=\unitlength,page=6]{0_equib.pdf}}%
    \put(0.04763908,0.13945998){\makebox(0,0)[rt]{\lineheight{1.25}\smash{\begin{tabular}[t]{r}−4\end{tabular}}}}%
    \put(0,0){\includegraphics[width=\unitlength,page=7]{0_equib.pdf}}%
    \put(0.04763908,0.32477889){\makebox(0,0)[rt]{\lineheight{1.25}\smash{\begin{tabular}[t]{r}−2\end{tabular}}}}%
    \put(0,0){\includegraphics[width=\unitlength,page=8]{0_equib.pdf}}%
    \put(0.04763908,0.51009775){\makebox(0,0)[rt]{\lineheight{1.25}\smash{\begin{tabular}[t]{r}0\end{tabular}}}}%
    \put(0,0){\includegraphics[width=\unitlength,page=9]{0_equib.pdf}}%
    \put(0.04763908,0.69541661){\makebox(0,0)[rt]{\lineheight{1.25}\smash{\begin{tabular}[t]{r}2\end{tabular}}}}%
    \put(0,0){\includegraphics[width=\unitlength,page=10]{0_equib.pdf}}%
    \put(0.04763908,0.8807355){\makebox(0,0)[rt]{\lineheight{1.25}\smash{\begin{tabular}[t]{r}4\end{tabular}}}}%
    \put(0,0){\includegraphics[width=\unitlength,page=11]{0_equib.pdf}}%
    \put(0.17855818,0.7484004){\color[rgb]{0,0,0}\makebox(0,0)[lt]{\lineheight{1.25}\smash{\begin{tabular}[t]{l}$\q_{eq} = \begin{bmatrix}0\\0\end{bmatrix}$\end{tabular}}}}%
  \end{picture}%
\endgroup%

%% file: pdftex/0_geodesics.pdf_tex
\begingroup%
  \makeatletter%
  \providecommand\color[2][]{%
    \errmessage{(Inkscape) Color is used for the text in Inkscape, but the package 'color.sty' is not loaded}%
    \renewcommand\color[2][]{}%
  }%
  \providecommand\transparent[1]{%
    \errmessage{(Inkscape) Transparency is used (non-zero) for the text in Inkscape, but the package 'transparent.sty' is not loaded}%
    \renewcommand\transparent[1]{}%
  }%
  \providecommand\rotatebox[2]{#2}%
  \newcommand*\fsize{\dimexpr\f@size pt\relax}%
  \newcommand*\lineheight[1]{\fontsize{\fsize}{#1\fsize}\selectfont}%
  \ifx\svgwidth\undefined%
    \setlength{\unitlength}{424.34039307bp}%
    \ifx\svgscale\undefined%
      \relax%
    \else%
      \setlength{\unitlength}{\unitlength * \real{\svgscale}}%
    \fi%
  \else%
    \setlength{\unitlength}{\svgwidth}%
  \fi%
  \global\let\svgwidth\undefined%
  \global\let\svgscale\undefined%
  \makeatother%
  \begin{picture}(1,0.72554673)%
    \lineheight{1}%
    \setlength\tabcolsep{0pt}%
    \put(0,0){\includegraphics[width=\unitlength,page=1]{0_geodesics.pdf}}%
    \put(0.13841301,0.00033371){\makebox(0,0)[t]{\lineheight{1.25}\smash{\begin{tabular}[t]{c}−1.5\end{tabular}}}}%
    \put(0,0){\includegraphics[width=\unitlength,page=2]{0_geodesics.pdf}}%
    \put(0.26688518,0.00033371){\makebox(0,0)[t]{\lineheight{1.25}\smash{\begin{tabular}[t]{c}−1.0\end{tabular}}}}%
    \put(0,0){\includegraphics[width=\unitlength,page=3]{0_geodesics.pdf}}%
    \put(0.39535734,0.00033371){\makebox(0,0)[t]{\lineheight{1.25}\smash{\begin{tabular}[t]{c}−0.5\end{tabular}}}}%
    \put(0,0){\includegraphics[width=\unitlength,page=4]{0_geodesics.pdf}}%
    \put(0.52382951,0.00033371){\makebox(0,0)[t]{\lineheight{1.25}\smash{\begin{tabular}[t]{c}0.0\end{tabular}}}}%
    \put(0,0){\includegraphics[width=\unitlength,page=5]{0_geodesics.pdf}}%
    \put(0.65230172,0.00033371){\makebox(0,0)[t]{\lineheight{1.25}\smash{\begin{tabular}[t]{c}0.5\end{tabular}}}}%
    \put(0,0){\includegraphics[width=\unitlength,page=6]{0_geodesics.pdf}}%
    \put(0.78077388,0.00033371){\makebox(0,0)[t]{\lineheight{1.25}\smash{\begin{tabular}[t]{c}1.0\end{tabular}}}}%
    \put(0,0){\includegraphics[width=\unitlength,page=7]{0_geodesics.pdf}}%
    \put(0.90924605,0.00033371){\makebox(0,0)[t]{\lineheight{1.25}\smash{\begin{tabular}[t]{c}1.5\end{tabular}}}}%
    \put(0,0){\includegraphics[width=\unitlength,page=8]{0_geodesics.pdf}}%
    \put(0.03224214,0.07744098){\makebox(0,0)[rt]{\lineheight{1.25}\smash{\begin{tabular}[t]{r}−6\end{tabular}}}}%
    \put(0,0){\includegraphics[width=\unitlength,page=9]{0_geodesics.pdf}}%
    \put(0.03224214,0.17560469){\makebox(0,0)[rt]{\lineheight{1.25}\smash{\begin{tabular}[t]{r}−4\end{tabular}}}}%
    \put(0,0){\includegraphics[width=\unitlength,page=10]{0_geodesics.pdf}}%
    \put(0.03224214,0.27376845){\makebox(0,0)[rt]{\lineheight{1.25}\smash{\begin{tabular}[t]{r}−2\end{tabular}}}}%
    \put(0,0){\includegraphics[width=\unitlength,page=11]{0_geodesics.pdf}}%
    \put(0.03224214,0.37193217){\makebox(0,0)[rt]{\lineheight{1.25}\smash{\begin{tabular}[t]{r}0\end{tabular}}}}%
    \put(0,0){\includegraphics[width=\unitlength,page=12]{0_geodesics.pdf}}%
    \put(0.03224214,0.4700959){\makebox(0,0)[rt]{\lineheight{1.25}\smash{\begin{tabular}[t]{r}2\end{tabular}}}}%
    \put(0,0){\includegraphics[width=\unitlength,page=13]{0_geodesics.pdf}}%
    \put(0.03224214,0.56825964){\makebox(0,0)[rt]{\lineheight{1.25}\smash{\begin{tabular}[t]{r}4\end{tabular}}}}%
    \put(0,0){\includegraphics[width=\unitlength,page=14]{0_geodesics.pdf}}%
    \put(0.03224214,0.66642337){\makebox(0,0)[rt]{\lineheight{1.25}\smash{\begin{tabular}[t]{r}6\end{tabular}}}}%
    \put(0,0){\includegraphics[width=\unitlength,page=15]{0_geodesics.pdf}}%
  \end{picture}%
\endgroup%

%% file: pdftex/0_geodesics_cartesianly.pdf_tex
\begingroup%
  \makeatletter%
  \providecommand\color[2][]{%
    \errmessage{(Inkscape) Color is used for the text in Inkscape, but the package 'color.sty' is not loaded}%
    \renewcommand\color[2][]{}%
  }%
  \providecommand\transparent[1]{%
    \errmessage{(Inkscape) Transparency is used (non-zero) for the text in Inkscape, but the package 'transparent.sty' is not loaded}%
    \renewcommand\transparent[1]{}%
  }%
  \providecommand\rotatebox[2]{#2}%
  \newcommand*\fsize{\dimexpr\f@size pt\relax}%
  \newcommand*\lineheight[1]{\fontsize{\fsize}{#1\fsize}\selectfont}%
  \ifx\svgwidth\undefined%
    \setlength{\unitlength}{424.44165039bp}%
    \ifx\svgscale\undefined%
      \relax%
    \else%
      \setlength{\unitlength}{\unitlength * \real{\svgscale}}%
    \fi%
  \else%
    \setlength{\unitlength}{\svgwidth}%
  \fi%
  \global\let\svgwidth\undefined%
  \global\let\svgscale\undefined%
  \makeatother%
  \begin{picture}(1,0.7256122)%
    \lineheight{1}%
    \setlength\tabcolsep{0pt}%
    \put(0,0){\includegraphics[width=\unitlength,page=1]{0_geodesics_cartesianly.pdf}}%
    \put(0.10395739,0.00033363){\makebox(0,0)[t]{\lineheight{1.25}\smash{\begin{tabular}[t]{c}−2\end{tabular}}}}%
    \put(0,0){\includegraphics[width=\unitlength,page=2]{0_geodesics_cartesianly.pdf}}%
    \put(0.24594129,0.00033363){\makebox(0,0)[t]{\lineheight{1.25}\smash{\begin{tabular}[t]{c}−1\end{tabular}}}}%
    \put(0,0){\includegraphics[width=\unitlength,page=3]{0_geodesics_cartesianly.pdf}}%
    \put(0.38792516,0.00033363){\makebox(0,0)[t]{\lineheight{1.25}\smash{\begin{tabular}[t]{c}0\end{tabular}}}}%
    \put(0,0){\includegraphics[width=\unitlength,page=4]{0_geodesics_cartesianly.pdf}}%
    \put(0.52990904,0.00033363){\makebox(0,0)[t]{\lineheight{1.25}\smash{\begin{tabular}[t]{c}1\end{tabular}}}}%
    \put(0,0){\includegraphics[width=\unitlength,page=5]{0_geodesics_cartesianly.pdf}}%
    \put(0.67189291,0.00033363){\makebox(0,0)[t]{\lineheight{1.25}\smash{\begin{tabular}[t]{c}2\end{tabular}}}}%
    \put(0,0){\includegraphics[width=\unitlength,page=6]{0_geodesics_cartesianly.pdf}}%
    \put(0.81387682,0.00033363){\makebox(0,0)[t]{\lineheight{1.25}\smash{\begin{tabular}[t]{c}3\end{tabular}}}}%
    \put(0,0){\includegraphics[width=\unitlength,page=7]{0_geodesics_cartesianly.pdf}}%
    \put(0.95586073,0.00033363){\makebox(0,0)[t]{\lineheight{1.25}\smash{\begin{tabular}[t]{c}4\end{tabular}}}}%
    \put(0,0){\includegraphics[width=\unitlength,page=8]{0_geodesics_cartesianly.pdf}}%
    \put(0.03223444,0.06274171){\makebox(0,0)[rt]{\lineheight{1.25}\smash{\begin{tabular}[t]{r}−4\end{tabular}}}}%
    \put(0,0){\includegraphics[width=\unitlength,page=9]{0_geodesics_cartesianly.pdf}}%
    \put(0.03223444,0.14048041){\makebox(0,0)[rt]{\lineheight{1.25}\smash{\begin{tabular}[t]{r}−3\end{tabular}}}}%
    \put(0,0){\includegraphics[width=\unitlength,page=10]{0_geodesics_cartesianly.pdf}}%
    \put(0.03223444,0.21821919){\makebox(0,0)[rt]{\lineheight{1.25}\smash{\begin{tabular}[t]{r}−2\end{tabular}}}}%
    \put(0,0){\includegraphics[width=\unitlength,page=11]{0_geodesics_cartesianly.pdf}}%
    \put(0.03223444,0.2959579){\makebox(0,0)[rt]{\lineheight{1.25}\smash{\begin{tabular}[t]{r}−1\end{tabular}}}}%
    \put(0,0){\includegraphics[width=\unitlength,page=12]{0_geodesics_cartesianly.pdf}}%
    \put(0.03223444,0.37369664){\makebox(0,0)[rt]{\lineheight{1.25}\smash{\begin{tabular}[t]{r}0\end{tabular}}}}%
    \put(0,0){\includegraphics[width=\unitlength,page=13]{0_geodesics_cartesianly.pdf}}%
    \put(0.03223444,0.45143534){\makebox(0,0)[rt]{\lineheight{1.25}\smash{\begin{tabular}[t]{r}1\end{tabular}}}}%
    \put(0,0){\includegraphics[width=\unitlength,page=14]{0_geodesics_cartesianly.pdf}}%
    \put(0.03223444,0.52917407){\makebox(0,0)[rt]{\lineheight{1.25}\smash{\begin{tabular}[t]{r}2\end{tabular}}}}%
    \put(0,0){\includegraphics[width=\unitlength,page=15]{0_geodesics_cartesianly.pdf}}%
    \put(0.03223444,0.6069128){\makebox(0,0)[rt]{\lineheight{1.25}\smash{\begin{tabular}[t]{r}3\end{tabular}}}}%
    \put(0,0){\includegraphics[width=\unitlength,page=16]{0_geodesics_cartesianly.pdf}}%
    \put(0.03223444,0.68465152){\makebox(0,0)[rt]{\lineheight{1.25}\smash{\begin{tabular}[t]{r}4\end{tabular}}}}%
    \put(0,0){\includegraphics[width=\unitlength,page=17]{0_geodesics_cartesianly.pdf}}%
  \end{picture}%
\endgroup%

%% file: pdftex/1_equib.pdf_tex
\begingroup%
  \makeatletter%
  \providecommand\color[2][]{%
    \errmessage{(Inkscape) Color is used for the text in Inkscape, but the package 'color.sty' is not loaded}%
    \renewcommand\color[2][]{}%
  }%
  \providecommand\transparent[1]{%
    \errmessage{(Inkscape) Transparency is used (non-zero) for the text in Inkscape, but the package 'transparent.sty' is not loaded}%
    \renewcommand\transparent[1]{}%
  }%
  \providecommand\rotatebox[2]{#2}%
  \newcommand*\fsize{\dimexpr\f@size pt\relax}%
  \newcommand*\lineheight[1]{\fontsize{\fsize}{#1\fsize}\selectfont}%
  \ifx\svgwidth\undefined%
    \setlength{\unitlength}{287.19363403bp}%
    \ifx\svgscale\undefined%
      \relax%
    \else%
      \setlength{\unitlength}{\unitlength * \real{\svgscale}}%
    \fi%
  \else%
    \setlength{\unitlength}{\svgwidth}%
  \fi%
  \global\let\svgwidth\undefined%
  \global\let\svgscale\undefined%
  \makeatother%
  \begin{picture}(1,0.98801648)%
    \lineheight{1}%
    \setlength\tabcolsep{0pt}%
    \put(0,0){\includegraphics[width=\unitlength,page=1]{1_equib.pdf}}%
    \put(0.16467231,0.00049308){\makebox(0,0)[t]{\lineheight{1.25}\smash{\begin{tabular}[t]{c}−4\end{tabular}}}}%
    \put(0,0){\includegraphics[width=\unitlength,page=2]{1_equib.pdf}}%
    \put(0.34999122,0.00049308){\makebox(0,0)[t]{\lineheight{1.25}\smash{\begin{tabular}[t]{c}−2\end{tabular}}}}%
    \put(0,0){\includegraphics[width=\unitlength,page=3]{1_equib.pdf}}%
    \put(0.53531008,0.00049308){\makebox(0,0)[t]{\lineheight{1.25}\smash{\begin{tabular}[t]{c}0\end{tabular}}}}%
    \put(0,0){\includegraphics[width=\unitlength,page=4]{1_equib.pdf}}%
    \put(0.72062889,0.00049308){\makebox(0,0)[t]{\lineheight{1.25}\smash{\begin{tabular}[t]{c}2\end{tabular}}}}%
    \put(0,0){\includegraphics[width=\unitlength,page=5]{1_equib.pdf}}%
    \put(0.9059478,0.00049308){\makebox(0,0)[t]{\lineheight{1.25}\smash{\begin{tabular}[t]{c}4\end{tabular}}}}%
    \put(0,0){\includegraphics[width=\unitlength,page=6]{1_equib.pdf}}%
    \put(0.04763908,0.13945995){\makebox(0,0)[rt]{\lineheight{1.25}\smash{\begin{tabular}[t]{r}−4\end{tabular}}}}%
    \put(0,0){\includegraphics[width=\unitlength,page=7]{1_equib.pdf}}%
    \put(0.04763908,0.32477887){\makebox(0,0)[rt]{\lineheight{1.25}\smash{\begin{tabular}[t]{r}−2\end{tabular}}}}%
    \put(0,0){\includegraphics[width=\unitlength,page=8]{1_equib.pdf}}%
    \put(0.04763908,0.51009773){\makebox(0,0)[rt]{\lineheight{1.25}\smash{\begin{tabular}[t]{r}0\end{tabular}}}}%
    \put(0,0){\includegraphics[width=\unitlength,page=9]{1_equib.pdf}}%
    \put(0.04763908,0.69541661){\makebox(0,0)[rt]{\lineheight{1.25}\smash{\begin{tabular}[t]{r}2\end{tabular}}}}%
    \put(0,0){\includegraphics[width=\unitlength,page=10]{1_equib.pdf}}%
    \put(0.04763908,0.8807355){\makebox(0,0)[rt]{\lineheight{1.25}\smash{\begin{tabular}[t]{r}4\end{tabular}}}}%
    \put(0,0){\includegraphics[width=\unitlength,page=11]{1_equib.pdf}}%
    \put(0.26007452,0.21155291){\color[rgb]{0,0,0}\makebox(0,0)[lt]{\lineheight{1.25}\smash{\begin{tabular}[t]{l}$\q_{eq} = \begin{bmatrix}0\\\frac{\pi}{2}\end{bmatrix}$\end{tabular}}}}%
  \end{picture}%
\endgroup%

%% file: pdftex/1_geodesics.pdf_tex
\begingroup%
  \makeatletter%
  \providecommand\color[2][]{%
    \errmessage{(Inkscape) Color is used for the text in Inkscape, but the package 'color.sty' is not loaded}%
    \renewcommand\color[2][]{}%
  }%
  \providecommand\transparent[1]{%
    \errmessage{(Inkscape) Transparency is used (non-zero) for the text in Inkscape, but the package 'transparent.sty' is not loaded}%
    \renewcommand\transparent[1]{}%
  }%
  \providecommand\rotatebox[2]{#2}%
  \newcommand*\fsize{\dimexpr\f@size pt\relax}%
  \newcommand*\lineheight[1]{\fontsize{\fsize}{#1\fsize}\selectfont}%
  \ifx\svgwidth\undefined%
    \setlength{\unitlength}{424.34039307bp}%
    \ifx\svgscale\undefined%
      \relax%
    \else%
      \setlength{\unitlength}{\unitlength * \real{\svgscale}}%
    \fi%
  \else%
    \setlength{\unitlength}{\svgwidth}%
  \fi%
  \global\let\svgwidth\undefined%
  \global\let\svgscale\undefined%
  \makeatother%
  \begin{picture}(1,0.72554673)%
    \lineheight{1}%
    \setlength\tabcolsep{0pt}%
    \put(0,0){\includegraphics[width=\unitlength,page=1]{1_geodesics.pdf}}%
    \put(0.18317896,0.00033371){\makebox(0,0)[t]{\lineheight{1.25}\smash{\begin{tabular}[t]{c}−1\end{tabular}}}}%
    \put(0,0){\includegraphics[width=\unitlength,page=2]{1_geodesics.pdf}}%
    \put(0.3647189,0.00033371){\makebox(0,0)[t]{\lineheight{1.25}\smash{\begin{tabular}[t]{c}0\end{tabular}}}}%
    \put(0,0){\includegraphics[width=\unitlength,page=3]{1_geodesics.pdf}}%
    \put(0.54625888,0.00033371){\makebox(0,0)[t]{\lineheight{1.25}\smash{\begin{tabular}[t]{c}1\end{tabular}}}}%
    \put(0,0){\includegraphics[width=\unitlength,page=4]{1_geodesics.pdf}}%
    \put(0.72779881,0.00033371){\makebox(0,0)[t]{\lineheight{1.25}\smash{\begin{tabular}[t]{c}2\end{tabular}}}}%
    \put(0,0){\includegraphics[width=\unitlength,page=5]{1_geodesics.pdf}}%
    \put(0.90933875,0.00033371){\makebox(0,0)[t]{\lineheight{1.25}\smash{\begin{tabular}[t]{c}3\end{tabular}}}}%
    \put(0,0){\includegraphics[width=\unitlength,page=6]{1_geodesics.pdf}}%
    \put(0.03224214,0.0840871){\makebox(0,0)[rt]{\lineheight{1.25}\smash{\begin{tabular}[t]{r}−6\end{tabular}}}}%
    \put(0,0){\includegraphics[width=\unitlength,page=7]{1_geodesics.pdf}}%
    \put(0.03224214,0.1808795){\makebox(0,0)[rt]{\lineheight{1.25}\smash{\begin{tabular}[t]{r}−4\end{tabular}}}}%
    \put(0,0){\includegraphics[width=\unitlength,page=8]{1_geodesics.pdf}}%
    \put(0.03224214,0.27767185){\makebox(0,0)[rt]{\lineheight{1.25}\smash{\begin{tabular}[t]{r}−2\end{tabular}}}}%
    \put(0,0){\includegraphics[width=\unitlength,page=9]{1_geodesics.pdf}}%
    \put(0.03224214,0.37446421){\makebox(0,0)[rt]{\lineheight{1.25}\smash{\begin{tabular}[t]{r}0\end{tabular}}}}%
    \put(0,0){\includegraphics[width=\unitlength,page=10]{1_geodesics.pdf}}%
    \put(0.03224214,0.47125657){\makebox(0,0)[rt]{\lineheight{1.25}\smash{\begin{tabular}[t]{r}2\end{tabular}}}}%
    \put(0,0){\includegraphics[width=\unitlength,page=11]{1_geodesics.pdf}}%
    \put(0.03224214,0.56804892){\makebox(0,0)[rt]{\lineheight{1.25}\smash{\begin{tabular}[t]{r}4\end{tabular}}}}%
    \put(0,0){\includegraphics[width=\unitlength,page=12]{1_geodesics.pdf}}%
    \put(0.03224214,0.66484128){\makebox(0,0)[rt]{\lineheight{1.25}\smash{\begin{tabular}[t]{r}6\end{tabular}}}}%
    \put(0,0){\includegraphics[width=\unitlength,page=13]{1_geodesics.pdf}}%
  \end{picture}%
\endgroup%

%% file: pdftex/1_geodesics_cartesianly.pdf_tex
\begingroup%
  \makeatletter%
  \providecommand\color[2][]{%
    \errmessage{(Inkscape) Color is used for the text in Inkscape, but the package 'color.sty' is not loaded}%
    \renewcommand\color[2][]{}%
  }%
  \providecommand\transparent[1]{%
    \errmessage{(Inkscape) Transparency is used (non-zero) for the text in Inkscape, but the package 'transparent.sty' is not loaded}%
    \renewcommand\transparent[1]{}%
  }%
  \providecommand\rotatebox[2]{#2}%
  \newcommand*\fsize{\dimexpr\f@size pt\relax}%
  \newcommand*\lineheight[1]{\fontsize{\fsize}{#1\fsize}\selectfont}%
  \ifx\svgwidth\undefined%
    \setlength{\unitlength}{424.44165039bp}%
    \ifx\svgscale\undefined%
      \relax%
    \else%
      \setlength{\unitlength}{\unitlength * \real{\svgscale}}%
    \fi%
  \else%
    \setlength{\unitlength}{\svgwidth}%
  \fi%
  \global\let\svgwidth\undefined%
  \global\let\svgscale\undefined%
  \makeatother%
  \begin{picture}(1,0.7256122)%
    \lineheight{1}%
    \setlength\tabcolsep{0pt}%
    \put(0,0){\includegraphics[width=\unitlength,page=1]{1_geodesics_cartesianly.pdf}}%
    \put(0.09122252,0.00033363){\makebox(0,0)[t]{\lineheight{1.25}\smash{\begin{tabular}[t]{c}−4\end{tabular}}}}%
    \put(0,0){\includegraphics[width=\unitlength,page=2]{1_geodesics_cartesianly.pdf}}%
    \put(0.19931991,0.00033363){\makebox(0,0)[t]{\lineheight{1.25}\smash{\begin{tabular}[t]{c}−3\end{tabular}}}}%
    \put(0,0){\includegraphics[width=\unitlength,page=3]{1_geodesics_cartesianly.pdf}}%
    \put(0.30741728,0.00033363){\makebox(0,0)[t]{\lineheight{1.25}\smash{\begin{tabular}[t]{c}−2\end{tabular}}}}%
    \put(0,0){\includegraphics[width=\unitlength,page=4]{1_geodesics_cartesianly.pdf}}%
    \put(0.41551469,0.00033363){\makebox(0,0)[t]{\lineheight{1.25}\smash{\begin{tabular}[t]{c}−1\end{tabular}}}}%
    \put(0,0){\includegraphics[width=\unitlength,page=5]{1_geodesics_cartesianly.pdf}}%
    \put(0.5236121,0.00033363){\makebox(0,0)[t]{\lineheight{1.25}\smash{\begin{tabular}[t]{c}0\end{tabular}}}}%
    \put(0,0){\includegraphics[width=\unitlength,page=6]{1_geodesics_cartesianly.pdf}}%
    \put(0.6317095,0.00033363){\makebox(0,0)[t]{\lineheight{1.25}\smash{\begin{tabular}[t]{c}1\end{tabular}}}}%
    \put(0,0){\includegraphics[width=\unitlength,page=7]{1_geodesics_cartesianly.pdf}}%
    \put(0.73980684,0.00033363){\makebox(0,0)[t]{\lineheight{1.25}\smash{\begin{tabular}[t]{c}2\end{tabular}}}}%
    \put(0,0){\includegraphics[width=\unitlength,page=8]{1_geodesics_cartesianly.pdf}}%
    \put(0.84790425,0.00033363){\makebox(0,0)[t]{\lineheight{1.25}\smash{\begin{tabular}[t]{c}3\end{tabular}}}}%
    \put(0,0){\includegraphics[width=\unitlength,page=9]{1_geodesics_cartesianly.pdf}}%
    \put(0.95600166,0.00033363){\makebox(0,0)[t]{\lineheight{1.25}\smash{\begin{tabular}[t]{c}4\end{tabular}}}}%
    \put(0,0){\includegraphics[width=\unitlength,page=10]{1_geodesics_cartesianly.pdf}}%
    \put(0.03223444,0.06105593){\makebox(0,0)[rt]{\lineheight{1.25}\smash{\begin{tabular}[t]{r}−4\end{tabular}}}}%
    \put(0,0){\includegraphics[width=\unitlength,page=11]{1_geodesics_cartesianly.pdf}}%
    \put(0.03223444,0.13904995){\makebox(0,0)[rt]{\lineheight{1.25}\smash{\begin{tabular}[t]{r}−3\end{tabular}}}}%
    \put(0,0){\includegraphics[width=\unitlength,page=12]{1_geodesics_cartesianly.pdf}}%
    \put(0.03223444,0.21704394){\makebox(0,0)[rt]{\lineheight{1.25}\smash{\begin{tabular}[t]{r}−2\end{tabular}}}}%
    \put(0,0){\includegraphics[width=\unitlength,page=13]{1_geodesics_cartesianly.pdf}}%
    \put(0.03223444,0.29503793){\makebox(0,0)[rt]{\lineheight{1.25}\smash{\begin{tabular}[t]{r}−1\end{tabular}}}}%
    \put(0,0){\includegraphics[width=\unitlength,page=14]{1_geodesics_cartesianly.pdf}}%
    \put(0.03223444,0.37303195){\makebox(0,0)[rt]{\lineheight{1.25}\smash{\begin{tabular}[t]{r}0\end{tabular}}}}%
    \put(0,0){\includegraphics[width=\unitlength,page=15]{1_geodesics_cartesianly.pdf}}%
    \put(0.03223444,0.45102596){\makebox(0,0)[rt]{\lineheight{1.25}\smash{\begin{tabular}[t]{r}1\end{tabular}}}}%
    \put(0,0){\includegraphics[width=\unitlength,page=16]{1_geodesics_cartesianly.pdf}}%
    \put(0.03223444,0.52901997){\makebox(0,0)[rt]{\lineheight{1.25}\smash{\begin{tabular}[t]{r}2\end{tabular}}}}%
    \put(0,0){\includegraphics[width=\unitlength,page=17]{1_geodesics_cartesianly.pdf}}%
    \put(0.03223444,0.60701397){\makebox(0,0)[rt]{\lineheight{1.25}\smash{\begin{tabular}[t]{r}3\end{tabular}}}}%
    \put(0,0){\includegraphics[width=\unitlength,page=18]{1_geodesics_cartesianly.pdf}}%
    \put(0.03223444,0.68500798){\makebox(0,0)[rt]{\lineheight{1.25}\smash{\begin{tabular}[t]{r}4\end{tabular}}}}%
    \put(0,0){\includegraphics[width=\unitlength,page=19]{1_geodesics_cartesianly.pdf}}%
  \end{picture}%
\endgroup%

%% file: pdftex/2_equib.pdf_tex
\begingroup%
  \makeatletter%
  \providecommand\color[2][]{%
    \errmessage{(Inkscape) Color is used for the text in Inkscape, but the package 'color.sty' is not loaded}%
    \renewcommand\color[2][]{}%
  }%
  \providecommand\transparent[1]{%
    \errmessage{(Inkscape) Transparency is used (non-zero) for the text in Inkscape, but the package 'transparent.sty' is not loaded}%
    \renewcommand\transparent[1]{}%
  }%
  \providecommand\rotatebox[2]{#2}%
  \newcommand*\fsize{\dimexpr\f@size pt\relax}%
  \newcommand*\lineheight[1]{\fontsize{\fsize}{#1\fsize}\selectfont}%
  \ifx\svgwidth\undefined%
    \setlength{\unitlength}{287.19363403bp}%
    \ifx\svgscale\undefined%
      \relax%
    \else%
      \setlength{\unitlength}{\unitlength * \real{\svgscale}}%
    \fi%
  \else%
    \setlength{\unitlength}{\svgwidth}%
  \fi%
  \global\let\svgwidth\undefined%
  \global\let\svgscale\undefined%
  \makeatother%
  \begin{picture}(1,0.98801648)%
    \lineheight{1}%
    \setlength\tabcolsep{0pt}%
    \put(0,0){\includegraphics[width=\unitlength,page=1]{2_equib.pdf}}%
    \put(0.16467231,0.00049305){\makebox(0,0)[t]{\lineheight{1.25}\smash{\begin{tabular}[t]{c}−4\end{tabular}}}}%
    \put(0,0){\includegraphics[width=\unitlength,page=2]{2_equib.pdf}}%
    \put(0.34999122,0.00049305){\makebox(0,0)[t]{\lineheight{1.25}\smash{\begin{tabular}[t]{c}−2\end{tabular}}}}%
    \put(0,0){\includegraphics[width=\unitlength,page=3]{2_equib.pdf}}%
    \put(0.53531006,0.00049305){\makebox(0,0)[t]{\lineheight{1.25}\smash{\begin{tabular}[t]{c}0\end{tabular}}}}%
    \put(0,0){\includegraphics[width=\unitlength,page=4]{2_equib.pdf}}%
    \put(0.72062886,0.00049305){\makebox(0,0)[t]{\lineheight{1.25}\smash{\begin{tabular}[t]{c}2\end{tabular}}}}%
    \put(0,0){\includegraphics[width=\unitlength,page=5]{2_equib.pdf}}%
    \put(0.90594778,0.00049305){\makebox(0,0)[t]{\lineheight{1.25}\smash{\begin{tabular}[t]{c}4\end{tabular}}}}%
    \put(0,0){\includegraphics[width=\unitlength,page=6]{2_equib.pdf}}%
    \put(0.04763908,0.13945998){\makebox(0,0)[rt]{\lineheight{1.25}\smash{\begin{tabular}[t]{r}−4\end{tabular}}}}%
    \put(0,0){\includegraphics[width=\unitlength,page=7]{2_equib.pdf}}%
    \put(0.04763908,0.32477889){\makebox(0,0)[rt]{\lineheight{1.25}\smash{\begin{tabular}[t]{r}−2\end{tabular}}}}%
    \put(0,0){\includegraphics[width=\unitlength,page=8]{2_equib.pdf}}%
    \put(0.04763908,0.51009775){\makebox(0,0)[rt]{\lineheight{1.25}\smash{\begin{tabular}[t]{r}0\end{tabular}}}}%
    \put(0,0){\includegraphics[width=\unitlength,page=9]{2_equib.pdf}}%
    \put(0.04763908,0.69541661){\makebox(0,0)[rt]{\lineheight{1.25}\smash{\begin{tabular}[t]{r}2\end{tabular}}}}%
    \put(0,0){\includegraphics[width=\unitlength,page=10]{2_equib.pdf}}%
    \put(0.04763908,0.8807355){\makebox(0,0)[rt]{\lineheight{1.25}\smash{\begin{tabular}[t]{r}4\end{tabular}}}}%
    \put(0,0){\includegraphics[width=\unitlength,page=11]{2_equib.pdf}}%
    \put(0.17804897,0.76661034){\color[rgb]{0,0,0}\makebox(0,0)[lt]{\lineheight{1.25}\smash{\begin{tabular}[t]{l}$\q_{eq} = \begin{bmatrix}0\\\pi\end{bmatrix}$\end{tabular}}}}%
  \end{picture}%
\endgroup%

%% file: pdftex/2_geodesics.pdf_tex
\begingroup%
  \makeatletter%
  \providecommand\color[2][]{%
    \errmessage{(Inkscape) Color is used for the text in Inkscape, but the package 'color.sty' is not loaded}%
    \renewcommand\color[2][]{}%
  }%
  \providecommand\transparent[1]{%
    \errmessage{(Inkscape) Transparency is used (non-zero) for the text in Inkscape, but the package 'transparent.sty' is not loaded}%
    \renewcommand\transparent[1]{}%
  }%
  \providecommand\rotatebox[2]{#2}%
  \newcommand*\fsize{\dimexpr\f@size pt\relax}%
  \newcommand*\lineheight[1]{\fontsize{\fsize}{#1\fsize}\selectfont}%
  \ifx\svgwidth\undefined%
    \setlength{\unitlength}{424.34039307bp}%
    \ifx\svgscale\undefined%
      \relax%
    \else%
      \setlength{\unitlength}{\unitlength * \real{\svgscale}}%
    \fi%
  \else%
    \setlength{\unitlength}{\svgwidth}%
  \fi%
  \global\let\svgwidth\undefined%
  \global\let\svgscale\undefined%
  \makeatother%
  \begin{picture}(1,0.72322602)%
    \lineheight{1}%
    \setlength\tabcolsep{0pt}%
    \put(0,0){\includegraphics[width=\unitlength,page=1]{2_geodesics.pdf}}%
    \put(0.12187027,0.00033369){\makebox(0,0)[t]{\lineheight{1.25}\smash{\begin{tabular}[t]{c}−3\end{tabular}}}}%
    \put(0,0){\includegraphics[width=\unitlength,page=2]{2_geodesics.pdf}}%
    \put(0.25573758,0.00033369){\makebox(0,0)[t]{\lineheight{1.25}\smash{\begin{tabular}[t]{c}−2\end{tabular}}}}%
    \put(0,0){\includegraphics[width=\unitlength,page=3]{2_geodesics.pdf}}%
    \put(0.38960486,0.00033369){\makebox(0,0)[t]{\lineheight{1.25}\smash{\begin{tabular}[t]{c}−1\end{tabular}}}}%
    \put(0,0){\includegraphics[width=\unitlength,page=4]{2_geodesics.pdf}}%
    \put(0.52347215,0.00033369){\makebox(0,0)[t]{\lineheight{1.25}\smash{\begin{tabular}[t]{c}0\end{tabular}}}}%
    \put(0,0){\includegraphics[width=\unitlength,page=5]{2_geodesics.pdf}}%
    \put(0.65733941,0.00033369){\makebox(0,0)[t]{\lineheight{1.25}\smash{\begin{tabular}[t]{c}1\end{tabular}}}}%
    \put(0,0){\includegraphics[width=\unitlength,page=6]{2_geodesics.pdf}}%
    \put(0.79120677,0.00033369){\makebox(0,0)[t]{\lineheight{1.25}\smash{\begin{tabular}[t]{c}2\end{tabular}}}}%
    \put(0,0){\includegraphics[width=\unitlength,page=7]{2_geodesics.pdf}}%
    \put(0.92507405,0.00033369){\makebox(0,0)[t]{\lineheight{1.25}\smash{\begin{tabular}[t]{c}3\end{tabular}}}}%
    \put(0,0){\includegraphics[width=\unitlength,page=8]{2_geodesics.pdf}}%
    \put(0.03224214,0.12244618){\makebox(0,0)[rt]{\lineheight{1.25}\smash{\begin{tabular}[t]{r}−2\end{tabular}}}}%
    \put(0,0){\includegraphics[width=\unitlength,page=9]{2_geodesics.pdf}}%
    \put(0.03224214,0.21966105){\makebox(0,0)[rt]{\lineheight{1.25}\smash{\begin{tabular}[t]{r}0\end{tabular}}}}%
    \put(0,0){\includegraphics[width=\unitlength,page=10]{2_geodesics.pdf}}%
    \put(0.03224214,0.316876){\makebox(0,0)[rt]{\lineheight{1.25}\smash{\begin{tabular}[t]{r}2\end{tabular}}}}%
    \put(0,0){\includegraphics[width=\unitlength,page=11]{2_geodesics.pdf}}%
    \put(0.03224214,0.41409091){\makebox(0,0)[rt]{\lineheight{1.25}\smash{\begin{tabular}[t]{r}4\end{tabular}}}}%
    \put(0,0){\includegraphics[width=\unitlength,page=12]{2_geodesics.pdf}}%
    \put(0.03224214,0.5113058){\makebox(0,0)[rt]{\lineheight{1.25}\smash{\begin{tabular}[t]{r}6\end{tabular}}}}%
    \put(0,0){\includegraphics[width=\unitlength,page=13]{2_geodesics.pdf}}%
    \put(0.03224214,0.60852072){\makebox(0,0)[rt]{\lineheight{1.25}\smash{\begin{tabular}[t]{r}8\end{tabular}}}}%
    \put(0,0){\includegraphics[width=\unitlength,page=14]{2_geodesics.pdf}}%
    \put(0.03224214,0.70573564){\makebox(0,0)[rt]{\lineheight{1.25}\smash{\begin{tabular}[t]{r}10\end{tabular}}}}%
    \put(0,0){\includegraphics[width=\unitlength,page=15]{2_geodesics.pdf}}%
  \end{picture}%
\endgroup%

%% file: pdftex/2_geodesics_cartesianly.pdf_tex
\begingroup%
  \makeatletter%
  \providecommand\color[2][]{%
    \errmessage{(Inkscape) Color is used for the text in Inkscape, but the package 'color.sty' is not loaded}%
    \renewcommand\color[2][]{}%
  }%
  \providecommand\transparent[1]{%
    \errmessage{(Inkscape) Transparency is used (non-zero) for the text in Inkscape, but the package 'transparent.sty' is not loaded}%
    \renewcommand\transparent[1]{}%
  }%
  \providecommand\rotatebox[2]{#2}%
  \newcommand*\fsize{\dimexpr\f@size pt\relax}%
  \newcommand*\lineheight[1]{\fontsize{\fsize}{#1\fsize}\selectfont}%
  \ifx\svgwidth\undefined%
    \setlength{\unitlength}{424.44165039bp}%
    \ifx\svgscale\undefined%
      \relax%
    \else%
      \setlength{\unitlength}{\unitlength * \real{\svgscale}}%
    \fi%
  \else%
    \setlength{\unitlength}{\svgwidth}%
  \fi%
  \global\let\svgwidth\undefined%
  \global\let\svgscale\undefined%
  \makeatother%
  \begin{picture}(1,0.7256122)%
    \lineheight{1}%
    \setlength\tabcolsep{0pt}%
    \put(0,0){\includegraphics[width=\unitlength,page=1]{2_geodesics_cartesianly.pdf}}%
    \put(0.09119336,0.00033363){\makebox(0,0)[t]{\lineheight{1.25}\smash{\begin{tabular}[t]{c}−4\end{tabular}}}}%
    \put(0,0){\includegraphics[width=\unitlength,page=2]{2_geodesics_cartesianly.pdf}}%
    \put(0.19931755,0.00033363){\makebox(0,0)[t]{\lineheight{1.25}\smash{\begin{tabular}[t]{c}−3\end{tabular}}}}%
    \put(0,0){\includegraphics[width=\unitlength,page=3]{2_geodesics_cartesianly.pdf}}%
    \put(0.30744173,0.00033363){\makebox(0,0)[t]{\lineheight{1.25}\smash{\begin{tabular}[t]{c}−2\end{tabular}}}}%
    \put(0,0){\includegraphics[width=\unitlength,page=4]{2_geodesics_cartesianly.pdf}}%
    \put(0.41556592,0.00033363){\makebox(0,0)[t]{\lineheight{1.25}\smash{\begin{tabular}[t]{c}−1\end{tabular}}}}%
    \put(0,0){\includegraphics[width=\unitlength,page=5]{2_geodesics_cartesianly.pdf}}%
    \put(0.52369011,0.00033363){\makebox(0,0)[t]{\lineheight{1.25}\smash{\begin{tabular}[t]{c}0\end{tabular}}}}%
    \put(0,0){\includegraphics[width=\unitlength,page=6]{2_geodesics_cartesianly.pdf}}%
    \put(0.63181426,0.00033363){\makebox(0,0)[t]{\lineheight{1.25}\smash{\begin{tabular}[t]{c}1\end{tabular}}}}%
    \put(0,0){\includegraphics[width=\unitlength,page=7]{2_geodesics_cartesianly.pdf}}%
    \put(0.73993849,0.00033363){\makebox(0,0)[t]{\lineheight{1.25}\smash{\begin{tabular}[t]{c}2\end{tabular}}}}%
    \put(0,0){\includegraphics[width=\unitlength,page=8]{2_geodesics_cartesianly.pdf}}%
    \put(0.84806265,0.00033363){\makebox(0,0)[t]{\lineheight{1.25}\smash{\begin{tabular}[t]{c}3\end{tabular}}}}%
    \put(0,0){\includegraphics[width=\unitlength,page=9]{2_geodesics_cartesianly.pdf}}%
    \put(0.9561868,0.00033363){\makebox(0,0)[t]{\lineheight{1.25}\smash{\begin{tabular}[t]{c}4\end{tabular}}}}%
    \put(0,0){\includegraphics[width=\unitlength,page=10]{2_geodesics_cartesianly.pdf}}%
    \put(0.03223444,0.06225358){\makebox(0,0)[rt]{\lineheight{1.25}\smash{\begin{tabular}[t]{r}−4\end{tabular}}}}%
    \put(0,0){\includegraphics[width=\unitlength,page=11]{2_geodesics_cartesianly.pdf}}%
    \put(0.03223444,0.14005282){\makebox(0,0)[rt]{\lineheight{1.25}\smash{\begin{tabular}[t]{r}−3\end{tabular}}}}%
    \put(0,0){\includegraphics[width=\unitlength,page=12]{2_geodesics_cartesianly.pdf}}%
    \put(0.03223444,0.2178521){\makebox(0,0)[rt]{\lineheight{1.25}\smash{\begin{tabular}[t]{r}−2\end{tabular}}}}%
    \put(0,0){\includegraphics[width=\unitlength,page=13]{2_geodesics_cartesianly.pdf}}%
    \put(0.03223444,0.29565142){\makebox(0,0)[rt]{\lineheight{1.25}\smash{\begin{tabular}[t]{r}−1\end{tabular}}}}%
    \put(0,0){\includegraphics[width=\unitlength,page=14]{2_geodesics_cartesianly.pdf}}%
    \put(0.03223444,0.3734507){\makebox(0,0)[rt]{\lineheight{1.25}\smash{\begin{tabular}[t]{r}0\end{tabular}}}}%
    \put(0,0){\includegraphics[width=\unitlength,page=15]{2_geodesics_cartesianly.pdf}}%
    \put(0.03223444,0.45124998){\makebox(0,0)[rt]{\lineheight{1.25}\smash{\begin{tabular}[t]{r}1\end{tabular}}}}%
    \put(0,0){\includegraphics[width=\unitlength,page=16]{2_geodesics_cartesianly.pdf}}%
    \put(0.03223444,0.52904927){\makebox(0,0)[rt]{\lineheight{1.25}\smash{\begin{tabular}[t]{r}2\end{tabular}}}}%
    \put(0,0){\includegraphics[width=\unitlength,page=17]{2_geodesics_cartesianly.pdf}}%
    \put(0.03223444,0.60684855){\makebox(0,0)[rt]{\lineheight{1.25}\smash{\begin{tabular}[t]{r}3\end{tabular}}}}%
    \put(0,0){\includegraphics[width=\unitlength,page=18]{2_geodesics_cartesianly.pdf}}%
    \put(0.03223444,0.68464784){\makebox(0,0)[rt]{\lineheight{1.25}\smash{\begin{tabular}[t]{r}4\end{tabular}}}}%
    \put(0,0){\includegraphics[width=\unitlength,page=19]{2_geodesics_cartesianly.pdf}}%
  \end{picture}%
\endgroup%

%% file: pdftex/fig_4_wider.pdf_tex
\begingroup%
  \makeatletter%
  \providecommand\color[2][]{%
    \errmessage{(Inkscape) Color is used for the text in Inkscape, but the package 'color.sty' is not loaded}%
    \renewcommand\color[2][]{}%
  }%
  \providecommand\transparent[1]{%
    \errmessage{(Inkscape) Transparency is used (non-zero) for the text in Inkscape, but the package 'transparent.sty' is not loaded}%
    \renewcommand\transparent[1]{}%
  }%
  \providecommand\rotatebox[2]{#2}%
  \newcommand*\fsize{\dimexpr\f@size pt\relax}%
  \newcommand*\lineheight[1]{\fontsize{\fsize}{#1\fsize}\selectfont}%
  \ifx\svgwidth\undefined%
    \setlength{\unitlength}{438.39059448bp}%
    \ifx\svgscale\undefined%
      \relax%
    \else%
      \setlength{\unitlength}{\unitlength * \real{\svgscale}}%
    \fi%
  \else%
    \setlength{\unitlength}{\svgwidth}%
  \fi%
  \global\let\svgwidth\undefined%
  \global\let\svgscale\undefined%
  \makeatother%
  \begin{picture}(1,0.30948805)%
    \lineheight{1}%
    \setlength\tabcolsep{0pt}%
    \put(0,0){\includegraphics[width=\unitlength,page=1]{fig_4_wider.pdf}}%
    \put(0.18635756,0.03594561){\makebox(0,0)[t]{\lineheight{1.25}\smash{\begin{tabular}[t]{c}−1.5\end{tabular}}}}%
    \put(0,0){\includegraphics[width=\unitlength,page=2]{fig_4_wider.pdf}}%
    \put(0.30257846,0.03594561){\makebox(0,0)[t]{\lineheight{1.25}\smash{\begin{tabular}[t]{c}−1.0\end{tabular}}}}%
    \put(0,0){\includegraphics[width=\unitlength,page=3]{fig_4_wider.pdf}}%
    \put(0.41879933,0.03594561){\makebox(0,0)[t]{\lineheight{1.25}\smash{\begin{tabular}[t]{c}−0.5\end{tabular}}}}%
    \put(0,0){\includegraphics[width=\unitlength,page=4]{fig_4_wider.pdf}}%
    \put(0.53502023,0.03594561){\makebox(0,0)[t]{\lineheight{1.25}\smash{\begin{tabular}[t]{c}0.0\end{tabular}}}}%
    \put(0,0){\includegraphics[width=\unitlength,page=5]{fig_4_wider.pdf}}%
    \put(0.65124109,0.03594561){\makebox(0,0)[t]{\lineheight{1.25}\smash{\begin{tabular}[t]{c}0.5\end{tabular}}}}%
    \put(0,0){\includegraphics[width=\unitlength,page=6]{fig_4_wider.pdf}}%
    \put(0.76746203,0.03594561){\makebox(0,0)[t]{\lineheight{1.25}\smash{\begin{tabular}[t]{c}1.0\end{tabular}}}}%
    \put(0,0){\includegraphics[width=\unitlength,page=7]{fig_4_wider.pdf}}%
    \put(0.88368289,0.03594561){\makebox(0,0)[t]{\lineheight{1.25}\smash{\begin{tabular}[t]{c}1.5\end{tabular}}}}%
    \put(0.53530021,0.00474482){\makebox(0,0)[t]{\lineheight{1.25}\smash{\begin{tabular}[t]{c}Angle in Degrees\end{tabular}}}}%
    \put(0,0){\includegraphics[width=\unitlength,page=8]{fig_4_wider.pdf}}%
    \put(0.05554541,0.06628206){\makebox(0,0)[rt]{\lineheight{1.25}\smash{\begin{tabular}[t]{r}0\end{tabular}}}}%
    \put(0,0){\includegraphics[width=\unitlength,page=9]{fig_4_wider.pdf}}%
    \put(0.05554541,0.11365732){\makebox(0,0)[rt]{\lineheight{1.25}\smash{\begin{tabular}[t]{r}250\end{tabular}}}}%
    \put(0,0){\includegraphics[width=\unitlength,page=10]{fig_4_wider.pdf}}%
    \put(0.05554541,0.16103258){\makebox(0,0)[rt]{\lineheight{1.25}\smash{\begin{tabular}[t]{r}500\end{tabular}}}}%
    \put(0,0){\includegraphics[width=\unitlength,page=11]{fig_4_wider.pdf}}%
    \put(0.05554541,0.20840783){\makebox(0,0)[rt]{\lineheight{1.25}\smash{\begin{tabular}[t]{r}750\end{tabular}}}}%
    \put(0,0){\includegraphics[width=\unitlength,page=12]{fig_4_wider.pdf}}%
    \put(0.05554541,0.25578309){\makebox(0,0)[rt]{\lineheight{1.25}\smash{\begin{tabular}[t]{r}1000\end{tabular}}}}%
    \put(0,0){\includegraphics[width=\unitlength,page=13]{fig_4_wider.pdf}}%
  \end{picture}%
\endgroup%

%% file: pdftex/surface_potential.pdf_tex
\begingroup%
  \makeatletter%
  \providecommand\color[2][]{%
    \errmessage{(Inkscape) Color is used for the text in Inkscape, but the package 'color.sty' is not loaded}%
    \renewcommand\color[2][]{}%
  }%
  \providecommand\transparent[1]{%
    \errmessage{(Inkscape) Transparency is used (non-zero) for the text in Inkscape, but the package 'transparent.sty' is not loaded}%
    \renewcommand\transparent[1]{}%
  }%
  \providecommand\rotatebox[2]{#2}%
  \newcommand*\fsize{\dimexpr\f@size pt\relax}%
  \newcommand*\lineheight[1]{\fontsize{\fsize}{#1\fsize}\selectfont}%
  \ifx\svgwidth\undefined%
    \setlength{\unitlength}{346.18914795bp}%
    \ifx\svgscale\undefined%
      \relax%
    \else%
      \setlength{\unitlength}{\unitlength * \real{\svgscale}}%
    \fi%
  \else%
    \setlength{\unitlength}{\svgwidth}%
  \fi%
  \global\let\svgwidth\undefined%
  \global\let\svgscale\undefined%
  \makeatother%
  \begin{picture}(1,0.75152236)%
    \lineheight{1}%
    \setlength\tabcolsep{0pt}%
    \put(0,0){\includegraphics[width=\unitlength,page=1]{surface_potential.pdf}}%
    \put(0.11054184,0.05664605){\rotatebox{-23.059787}{\makebox(0,0)[lt]{\lineheight{1.25}\smash{\begin{tabular}[t]{l}\textsl{q}1\end{tabular}}}}}%
    \put(0,0){\includegraphics[width=\unitlength,page=2]{surface_potential.pdf}}%
    \put(0.02395927,0.13235637){\makebox(0,0)[t]{\lineheight{1.25}\smash{\begin{tabular}[t]{c}−1\end{tabular}}}}%
    \put(0,0){\includegraphics[width=\unitlength,page=3]{surface_potential.pdf}}%
    \put(0.09515979,0.10100382){\makebox(0,0)[t]{\lineheight{1.25}\smash{\begin{tabular}[t]{c}0\end{tabular}}}}%
    \put(0,0){\includegraphics[width=\unitlength,page=4]{surface_potential.pdf}}%
    \put(0.16875158,0.06859837){\makebox(0,0)[t]{\lineheight{1.25}\smash{\begin{tabular}[t]{c}1\end{tabular}}}}%
    \put(0,0){\includegraphics[width=\unitlength,page=5]{surface_potential.pdf}}%
    \put(0.24485715,0.03508599){\makebox(0,0)[t]{\lineheight{1.25}\smash{\begin{tabular}[t]{c}2\end{tabular}}}}%
    \put(0,0){\includegraphics[width=\unitlength,page=6]{surface_potential.pdf}}%
    \put(0.32360753,0.00040902){\makebox(0,0)[t]{\lineheight{1.25}\smash{\begin{tabular}[t]{c}3\end{tabular}}}}%
    \put(0,0){\includegraphics[width=\unitlength,page=7]{surface_potential.pdf}}%
    \put(0.68527419,0.03231309){\rotatebox{18.362154}{\makebox(0,0)[lt]{\lineheight{1.25}\smash{\begin{tabular}[t]{l}\textsl{q}2\end{tabular}}}}}%
    \put(0,0){\includegraphics[width=\unitlength,page=8]{surface_potential.pdf}}%
    \put(0.51856632,0.00957119){\makebox(0,0)[t]{\lineheight{1.25}\smash{\begin{tabular}[t]{c}−2\end{tabular}}}}%
    \put(0,0){\includegraphics[width=\unitlength,page=9]{surface_potential.pdf}}%
    \put(0.60456044,0.03904114){\makebox(0,0)[t]{\lineheight{1.25}\smash{\begin{tabular}[t]{c}0\end{tabular}}}}%
    \put(0,0){\includegraphics[width=\unitlength,page=10]{surface_potential.pdf}}%
    \put(0.68808294,0.06766404){\makebox(0,0)[t]{\lineheight{1.25}\smash{\begin{tabular}[t]{c}2\end{tabular}}}}%
    \put(0,0){\includegraphics[width=\unitlength,page=11]{surface_potential.pdf}}%
    \put(0.76923888,0.09547602){\makebox(0,0)[t]{\lineheight{1.25}\smash{\begin{tabular}[t]{c}4\end{tabular}}}}%
    \put(0,0){\includegraphics[width=\unitlength,page=12]{surface_potential.pdf}}%
    \put(0.84812744,0.12251084){\makebox(0,0)[t]{\lineheight{1.25}\smash{\begin{tabular}[t]{c}6\end{tabular}}}}%
    \put(0,0){\includegraphics[width=\unitlength,page=13]{surface_potential.pdf}}%
    \put(0.93989193,0.60071421){\rotatebox{88.444117}{\makebox(0,0)[lt]{\lineheight{1.25}\smash{\begin{tabular}[t]{l}Potential U(q)\end{tabular}}}}}%
    \put(0,0){\includegraphics[width=\unitlength,page=14]{surface_potential.pdf}}%
    \put(0.92115172,0.19029129){\makebox(0,0)[t]{\lineheight{1.25}\smash{\begin{tabular}[t]{c}0\end{tabular}}}}%
    \put(0,0){\includegraphics[width=\unitlength,page=15]{surface_potential.pdf}}%
    \put(0.92298953,0.24827409){\makebox(0,0)[t]{\lineheight{1.25}\smash{\begin{tabular}[t]{c}5\end{tabular}}}}%
    \put(0,0){\includegraphics[width=\unitlength,page=16]{surface_potential.pdf}}%
    \put(0.92484118,0.30669174){\makebox(0,0)[t]{\lineheight{1.25}\smash{\begin{tabular}[t]{c}10\end{tabular}}}}%
    \put(0,0){\includegraphics[width=\unitlength,page=17]{surface_potential.pdf}}%
    \put(0.92670676,0.36554922){\makebox(0,0)[t]{\lineheight{1.25}\smash{\begin{tabular}[t]{c}15\end{tabular}}}}%
    \put(0,0){\includegraphics[width=\unitlength,page=18]{surface_potential.pdf}}%
    \put(0.92858644,0.42485144){\makebox(0,0)[t]{\lineheight{1.25}\smash{\begin{tabular}[t]{c}20\end{tabular}}}}%
    \put(0,0){\includegraphics[width=\unitlength,page=19]{surface_potential.pdf}}%
    \put(0.93048041,0.48460351){\makebox(0,0)[t]{\lineheight{1.25}\smash{\begin{tabular}[t]{c}25\end{tabular}}}}%
    \put(0,0){\includegraphics[width=\unitlength,page=20]{surface_potential.pdf}}%
    \put(0.93238874,0.54481058){\makebox(0,0)[t]{\lineheight{1.25}\smash{\begin{tabular}[t]{c}30\end{tabular}}}}%
    \put(0,0){\includegraphics[width=\unitlength,page=21]{surface_potential.pdf}}%
  \end{picture}%
\endgroup%

%% file: pdftex/pp_phase.pdf_tex
\begingroup%
  \makeatletter%
  \providecommand\color[2][]{%
    \errmessage{(Inkscape) Color is used for the text in Inkscape, but the package 'color.sty' is not loaded}%
    \renewcommand\color[2][]{}%
  }%
  \providecommand\transparent[1]{%
    \errmessage{(Inkscape) Transparency is used (non-zero) for the text in Inkscape, but the package 'transparent.sty' is not loaded}%
    \renewcommand\transparent[1]{}%
  }%
  \providecommand\rotatebox[2]{#2}%
  \newcommand*\fsize{\dimexpr\f@size pt\relax}%
  \newcommand*\lineheight[1]{\fontsize{\fsize}{#1\fsize}\selectfont}%
  \ifx\svgwidth\undefined%
    \setlength{\unitlength}{432.48800659bp}%
    \ifx\svgscale\undefined%
      \relax%
    \else%
      \setlength{\unitlength}{\unitlength * \real{\svgscale}}%
    \fi%
  \else%
    \setlength{\unitlength}{\svgwidth}%
  \fi%
  \global\let\svgwidth\undefined%
  \global\let\svgscale\undefined%
  \makeatother%
  \begin{picture}(1,0.65609229)%
    \lineheight{1}%
    \setlength\tabcolsep{0pt}%
    \put(0,0){\includegraphics[width=\unitlength,page=1]{pp_phase.pdf}}%
    \put(0.14642774,0.00032743){\makebox(0,0)[t]{\lineheight{1.25}\smash{\begin{tabular}[t]{c}−1\end{tabular}}}}%
    \put(0,0){\includegraphics[width=\unitlength,page=2]{pp_phase.pdf}}%
    \put(0.32824367,0.00032743){\makebox(0,0)[t]{\lineheight{1.25}\smash{\begin{tabular}[t]{c}0\end{tabular}}}}%
    \put(0,0){\includegraphics[width=\unitlength,page=3]{pp_phase.pdf}}%
    \put(0.5100596,0.00032743){\makebox(0,0)[t]{\lineheight{1.25}\smash{\begin{tabular}[t]{c}1\end{tabular}}}}%
    \put(0,0){\includegraphics[width=\unitlength,page=4]{pp_phase.pdf}}%
    \put(0.69187553,0.00032743){\makebox(0,0)[t]{\lineheight{1.25}\smash{\begin{tabular}[t]{c}2\end{tabular}}}}%
    \put(0,0){\includegraphics[width=\unitlength,page=5]{pp_phase.pdf}}%
    \put(0.87369146,0.00032743){\makebox(0,0)[t]{\lineheight{1.25}\smash{\begin{tabular}[t]{c}3\end{tabular}}}}%
    \put(0,0){\includegraphics[width=\unitlength,page=6]{pp_phase.pdf}}%
    \put(0.05751595,0.10560781){\makebox(0,0)[rt]{\lineheight{1.25}\smash{\begin{tabular}[t]{r}−2\end{tabular}}}}%
    \put(0,0){\includegraphics[width=\unitlength,page=7]{pp_phase.pdf}}%
    \put(0.05751595,0.23617957){\makebox(0,0)[rt]{\lineheight{1.25}\smash{\begin{tabular}[t]{r}0\end{tabular}}}}%
    \put(0,0){\includegraphics[width=\unitlength,page=8]{pp_phase.pdf}}%
    \put(0.05751595,0.36675132){\makebox(0,0)[rt]{\lineheight{1.25}\smash{\begin{tabular}[t]{r}2\end{tabular}}}}%
    \put(0,0){\includegraphics[width=\unitlength,page=9]{pp_phase.pdf}}%
    \put(0.05751595,0.49732308){\makebox(0,0)[rt]{\lineheight{1.25}\smash{\begin{tabular}[t]{r}4\end{tabular}}}}%
    \put(0,0){\includegraphics[width=\unitlength,page=10]{pp_phase.pdf}}%
    \put(0.05751595,0.62789483){\makebox(0,0)[rt]{\lineheight{1.25}\smash{\begin{tabular}[t]{r}6\end{tabular}}}}%
    \put(0,0){\includegraphics[width=\unitlength,page=11]{pp_phase.pdf}}%
    \put(0.61253373,0.61166703){\makebox(0,0)[lt]{\lineheight{1.25}\smash{\begin{tabular}[t]{l}Geodesics\end{tabular}}}}%
    \put(0,0){\includegraphics[width=\unitlength,page=12]{pp_phase.pdf}}%
    \put(0.61253373,0.57772823){\makebox(0,0)[lt]{\lineheight{1.25}\smash{\begin{tabular}[t]{l}Trajectory 1\end{tabular}}}}%
    \put(0,0){\includegraphics[width=\unitlength,page=13]{pp_phase.pdf}}%
    \put(0.61253373,0.54378943){\makebox(0,0)[lt]{\lineheight{1.25}\smash{\begin{tabular}[t]{l}Trajectory 2\end{tabular}}}}%
    \put(0,0){\includegraphics[width=\unitlength,page=14]{pp_phase.pdf}}%
    \put(0.61253373,0.50985063){\makebox(0,0)[lt]{\lineheight{1.25}\smash{\begin{tabular}[t]{l}Trajectory 3\end{tabular}}}}%
    \put(0,0){\includegraphics[width=\unitlength,page=15]{pp_phase.pdf}}%
    \put(0.61253373,0.47591181){\makebox(0,0)[lt]{\lineheight{1.25}\smash{\begin{tabular}[t]{l}Trajectory 4\end{tabular}}}}%
    \put(0,0){\includegraphics[width=\unitlength,page=16]{pp_phase.pdf}}%
    \put(0.61253373,0.44197303){\makebox(0,0)[lt]{\lineheight{1.25}\smash{\begin{tabular}[t]{l}Trajectory 5\end{tabular}}}}%
    \put(0,0){\includegraphics[width=\unitlength,page=17]{pp_phase.pdf}}%
    \put(0.61253373,0.40803421){\makebox(0,0)[lt]{\lineheight{1.25}\smash{\begin{tabular}[t]{l}Trajectory 6\end{tabular}}}}%
    \put(0,0){\includegraphics[width=\unitlength,page=18]{pp_phase.pdf}}%
    \put(0.61253373,0.37409543){\makebox(0,0)[lt]{\lineheight{1.25}\smash{\begin{tabular}[t]{l}$U(\qq)=\text{const}$\end{tabular}}}}%
    \put(0.52846131,0.00549487){\color[rgb]{0,0,0}\makebox(0,0)[lt]{\lineheight{1.25}\smash{\begin{tabular}[t]{l}$q_1$ in rad\end{tabular}}}}%
    \put(-0.00187608,0.62873445){\color[rgb]{0,0,0}\makebox(0,0)[lt]{\lineheight{1.25}\smash{\begin{tabular}[t]{l}$q_2$\\in rad\end{tabular}}}}%
  \end{picture}%
\endgroup%

%% file: pdftex/pp_q_3.pdf_tex
\begingroup%
  \makeatletter%
  \providecommand\color[2][]{%
    \errmessage{(Inkscape) Color is used for the text in Inkscape, but the package 'color.sty' is not loaded}%
    \renewcommand\color[2][]{}%
  }%
  \providecommand\transparent[1]{%
    \errmessage{(Inkscape) Transparency is used (non-zero) for the text in Inkscape, but the package 'transparent.sty' is not loaded}%
    \renewcommand\transparent[1]{}%
  }%
  \providecommand\rotatebox[2]{#2}%
  \newcommand*\fsize{\dimexpr\f@size pt\relax}%
  \newcommand*\lineheight[1]{\fontsize{\fsize}{#1\fsize}\selectfont}%
  \ifx\svgwidth\undefined%
    \setlength{\unitlength}{387.74264526bp}%
    \ifx\svgscale\undefined%
      \relax%
    \else%
      \setlength{\unitlength}{\unitlength * \real{\svgscale}}%
    \fi%
  \else%
    \setlength{\unitlength}{\svgwidth}%
  \fi%
  \global\let\svgwidth\undefined%
  \global\let\svgscale\undefined%
  \makeatother%
  \begin{picture}(1,0.76708141)%
    \lineheight{1}%
    \setlength\tabcolsep{0pt}%
    \put(0,0){\includegraphics[width=\unitlength,page=1]{pp_q_3.pdf}}%
    \put(0.11980984,0.03564153){\makebox(0,0)[t]{\lineheight{1.25}\smash{\begin{tabular}[t]{c}0.0\end{tabular}}}}%
    \put(0,0){\includegraphics[width=\unitlength,page=2]{pp_q_3.pdf}}%
    \put(0.32965794,0.03564153){\makebox(0,0)[t]{\lineheight{1.25}\smash{\begin{tabular}[t]{c}5.0\end{tabular}}}}%
    \put(0,0){\includegraphics[width=\unitlength,page=3]{pp_q_3.pdf}}%
    \put(0.53950602,0.03564153){\makebox(0,0)[t]{\lineheight{1.25}\smash{\begin{tabular}[t]{c}10.0\end{tabular}}}}%
    \put(0,0){\includegraphics[width=\unitlength,page=4]{pp_q_3.pdf}}%
    \put(0.74935414,0.03564153){\makebox(0,0)[t]{\lineheight{1.25}\smash{\begin{tabular}[t]{c}15.0\end{tabular}}}}%
    \put(0,0){\includegraphics[width=\unitlength,page=5]{pp_q_3.pdf}}%
    \put(0.95920218,0.03564153){\makebox(0,0)[t]{\lineheight{1.25}\smash{\begin{tabular}[t]{c}20.0\end{tabular}}}}%
    \put(0.53845681,0.00036519){\makebox(0,0)[t]{\lineheight{1.25}\smash{\begin{tabular}[t]{c}Time in sec\end{tabular}}}}%
    \put(0,0){\includegraphics[width=\unitlength,page=6]{pp_q_3.pdf}}%
    \put(0.05989193,0.11234659){\makebox(0,0)[rt]{\lineheight{1.25}\smash{\begin{tabular}[t]{r}−0.5\end{tabular}}}}%
    \put(0,0){\includegraphics[width=\unitlength,page=7]{pp_q_3.pdf}}%
    \put(0.05989193,0.25511651){\makebox(0,0)[rt]{\lineheight{1.25}\smash{\begin{tabular}[t]{r}0.0\end{tabular}}}}%
    \put(0,0){\includegraphics[width=\unitlength,page=8]{pp_q_3.pdf}}%
    \put(0.05989193,0.39788647){\makebox(0,0)[rt]{\lineheight{1.25}\smash{\begin{tabular}[t]{r}0.5\end{tabular}}}}%
    \put(0,0){\includegraphics[width=\unitlength,page=9]{pp_q_3.pdf}}%
    \put(0.05989193,0.54065639){\makebox(0,0)[rt]{\lineheight{1.25}\smash{\begin{tabular}[t]{r}1.0\end{tabular}}}}%
    \put(0,0){\includegraphics[width=\unitlength,page=10]{pp_q_3.pdf}}%
    \put(0.05989193,0.68342635){\makebox(0,0)[rt]{\lineheight{1.25}\smash{\begin{tabular}[t]{r}1.5\end{tabular}}}}%
    \put(0,0){\includegraphics[width=\unitlength,page=11]{pp_q_3.pdf}}%
  \end{picture}%
\endgroup%

%% file: pdftex/pp_q_5.pdf_tex
\begingroup%
  \makeatletter%
  \providecommand\color[2][]{%
    \errmessage{(Inkscape) Color is used for the text in Inkscape, but the package 'color.sty' is not loaded}%
    \renewcommand\color[2][]{}%
  }%
  \providecommand\transparent[1]{%
    \errmessage{(Inkscape) Transparency is used (non-zero) for the text in Inkscape, but the package 'transparent.sty' is not loaded}%
    \renewcommand\transparent[1]{}%
  }%
  \providecommand\rotatebox[2]{#2}%
  \newcommand*\fsize{\dimexpr\f@size pt\relax}%
  \newcommand*\lineheight[1]{\fontsize{\fsize}{#1\fsize}\selectfont}%
  \ifx\svgwidth\undefined%
    \setlength{\unitlength}{378.20162964bp}%
    \ifx\svgscale\undefined%
      \relax%
    \else%
      \setlength{\unitlength}{\unitlength * \real{\svgscale}}%
    \fi%
  \else%
    \setlength{\unitlength}{\svgwidth}%
  \fi%
  \global\let\svgwidth\undefined%
  \global\let\svgscale\undefined%
  \makeatother%
  \begin{picture}(1,0.78779086)%
    \lineheight{1}%
    \setlength\tabcolsep{0pt}%
    \put(0,0){\includegraphics[width=\unitlength,page=1]{pp_q_5.pdf}}%
    \put(0.09760499,0.03654069){\makebox(0,0)[t]{\lineheight{1.25}\smash{\begin{tabular}[t]{c}0.0\end{tabular}}}}%
    \put(0,0){\includegraphics[width=\unitlength,page=2]{pp_q_5.pdf}}%
    \put(0.312747,0.03654069){\makebox(0,0)[t]{\lineheight{1.25}\smash{\begin{tabular}[t]{c}5.0\end{tabular}}}}%
    \put(0,0){\includegraphics[width=\unitlength,page=3]{pp_q_5.pdf}}%
    \put(0.52788899,0.03654069){\makebox(0,0)[t]{\lineheight{1.25}\smash{\begin{tabular}[t]{c}10.0\end{tabular}}}}%
    \put(0,0){\includegraphics[width=\unitlength,page=4]{pp_q_5.pdf}}%
    \put(0.74303102,0.03654069){\makebox(0,0)[t]{\lineheight{1.25}\smash{\begin{tabular}[t]{c}15.0\end{tabular}}}}%
    \put(0,0){\includegraphics[width=\unitlength,page=5]{pp_q_5.pdf}}%
    \put(0.95817296,0.03654069){\makebox(0,0)[t]{\lineheight{1.25}\smash{\begin{tabular}[t]{c}20.0\end{tabular}}}}%
    \put(0.52681327,0.00037449){\makebox(0,0)[t]{\lineheight{1.25}\smash{\begin{tabular}[t]{c}Time in sec\end{tabular}}}}%
    \put(0,0){\includegraphics[width=\unitlength,page=6]{pp_q_5.pdf}}%
    \put(0.03617552,0.09526894){\makebox(0,0)[rt]{\lineheight{1.25}\smash{\begin{tabular}[t]{r}−2\end{tabular}}}}%
    \put(0,0){\includegraphics[width=\unitlength,page=7]{pp_q_5.pdf}}%
    \put(0.03617552,0.2298485){\makebox(0,0)[rt]{\lineheight{1.25}\smash{\begin{tabular}[t]{r}−1\end{tabular}}}}%
    \put(0,0){\includegraphics[width=\unitlength,page=8]{pp_q_5.pdf}}%
    \put(0.03617552,0.36442806){\makebox(0,0)[rt]{\lineheight{1.25}\smash{\begin{tabular}[t]{r}0\end{tabular}}}}%
    \put(0,0){\includegraphics[width=\unitlength,page=9]{pp_q_5.pdf}}%
    \put(0.03617552,0.49900762){\makebox(0,0)[rt]{\lineheight{1.25}\smash{\begin{tabular}[t]{r}1\end{tabular}}}}%
    \put(0,0){\includegraphics[width=\unitlength,page=10]{pp_q_5.pdf}}%
    \put(0.03617552,0.63358718){\makebox(0,0)[rt]{\lineheight{1.25}\smash{\begin{tabular}[t]{r}2\end{tabular}}}}%
    \put(0,0){\includegraphics[width=\unitlength,page=11]{pp_q_5.pdf}}%
    \put(0.03617552,0.76816674){\makebox(0,0)[rt]{\lineheight{1.25}\smash{\begin{tabular}[t]{r}3\end{tabular}}}}%
    \put(0,0){\includegraphics[width=\unitlength,page=12]{pp_q_5.pdf}}%
  \end{picture}%
\endgroup%

%% file: pdftex/pp_q_6.pdf_tex
\begingroup%
  \makeatletter%
  \providecommand\color[2][]{%
    \errmessage{(Inkscape) Color is used for the text in Inkscape, but the package 'color.sty' is not loaded}%
    \renewcommand\color[2][]{}%
  }%
  \providecommand\transparent[1]{%
    \errmessage{(Inkscape) Transparency is used (non-zero) for the text in Inkscape, but the package 'transparent.sty' is not loaded}%
    \renewcommand\transparent[1]{}%
  }%
  \providecommand\rotatebox[2]{#2}%
  \newcommand*\fsize{\dimexpr\f@size pt\relax}%
  \newcommand*\lineheight[1]{\fontsize{\fsize}{#1\fsize}\selectfont}%
  \ifx\svgwidth\undefined%
    \setlength{\unitlength}{378.20162964bp}%
    \ifx\svgscale\undefined%
      \relax%
    \else%
      \setlength{\unitlength}{\unitlength * \real{\svgscale}}%
    \fi%
  \else%
    \setlength{\unitlength}{\svgwidth}%
  \fi%
  \global\let\svgwidth\undefined%
  \global\let\svgscale\undefined%
  \makeatother%
  \begin{picture}(1,0.78643282)%
    \lineheight{1}%
    \setlength\tabcolsep{0pt}%
    \put(0,0){\includegraphics[width=\unitlength,page=1]{pp_q_6.pdf}}%
    \put(0.09760499,0.03654065){\makebox(0,0)[t]{\lineheight{1.25}\smash{\begin{tabular}[t]{c}0.0\end{tabular}}}}%
    \put(0,0){\includegraphics[width=\unitlength,page=2]{pp_q_6.pdf}}%
    \put(0.31274702,0.03654065){\makebox(0,0)[t]{\lineheight{1.25}\smash{\begin{tabular}[t]{c}5.0\end{tabular}}}}%
    \put(0,0){\includegraphics[width=\unitlength,page=3]{pp_q_6.pdf}}%
    \put(0.52788897,0.03654065){\makebox(0,0)[t]{\lineheight{1.25}\smash{\begin{tabular}[t]{c}10.0\end{tabular}}}}%
    \put(0,0){\includegraphics[width=\unitlength,page=4]{pp_q_6.pdf}}%
    \put(0.743031,0.03654065){\makebox(0,0)[t]{\lineheight{1.25}\smash{\begin{tabular}[t]{c}15.0\end{tabular}}}}%
    \put(0,0){\includegraphics[width=\unitlength,page=5]{pp_q_6.pdf}}%
    \put(0.95817294,0.03654065){\makebox(0,0)[t]{\lineheight{1.25}\smash{\begin{tabular}[t]{c}20.0\end{tabular}}}}%
    \put(0.52681327,0.00037441){\makebox(0,0)[t]{\lineheight{1.25}\smash{\begin{tabular}[t]{c}Time in sec\end{tabular}}}}%
    \put(0,0){\includegraphics[width=\unitlength,page=6]{pp_q_6.pdf}}%
    \put(0.03617552,0.12588636){\makebox(0,0)[rt]{\lineheight{1.25}\smash{\begin{tabular}[t]{r}−2\end{tabular}}}}%
    \put(0,0){\includegraphics[width=\unitlength,page=7]{pp_q_6.pdf}}%
    \put(0.03617552,0.25344546){\makebox(0,0)[rt]{\lineheight{1.25}\smash{\begin{tabular}[t]{r}−1\end{tabular}}}}%
    \put(0,0){\includegraphics[width=\unitlength,page=8]{pp_q_6.pdf}}%
    \put(0.03617552,0.38100455){\makebox(0,0)[rt]{\lineheight{1.25}\smash{\begin{tabular}[t]{r}0\end{tabular}}}}%
    \put(0,0){\includegraphics[width=\unitlength,page=9]{pp_q_6.pdf}}%
    \put(0.03617552,0.50856363){\makebox(0,0)[rt]{\lineheight{1.25}\smash{\begin{tabular}[t]{r}1\end{tabular}}}}%
    \put(0,0){\includegraphics[width=\unitlength,page=10]{pp_q_6.pdf}}%
    \put(0.03617552,0.63612274){\makebox(0,0)[rt]{\lineheight{1.25}\smash{\begin{tabular}[t]{r}2\end{tabular}}}}%
    \put(0,0){\includegraphics[width=\unitlength,page=11]{pp_q_6.pdf}}%
    \put(0.03617552,0.76368183){\makebox(0,0)[rt]{\lineheight{1.25}\smash{\begin{tabular}[t]{r}3\end{tabular}}}}%
    \put(0,0){\includegraphics[width=\unitlength,page=12]{pp_q_6.pdf}}%
  \end{picture}%
\endgroup%

%% file: sec/controllers.tex
\section{Controllers}\label{sec:controllers}
In this section two controllers will be derived and evaluated for the example system.
The first controller will stabilize the system onto one of the infinitely many strict normal modes, while the other will swing up and keep the system at the desired energy level.

\subsection{Mode Selection Controller}
In order to derive a mode selection controller some way of labeling the (infinitely many) strict normal modes is required.
Generally, this can be achieved by a function $\ttheta: \mathcal{Q} \rightarrow \mathcal{X}$, where $\dim\mathcal{X} = n-1$, i.e., the space of available strict modes is of dimension $n-1$.
The value of $\ttheta(\qq)$ will encode to which mode the point $\qq$ belongs.
Because all modes pass through the equilibrium $\qeq$, such a function cannot exist at that point.

Here, we have $n=2$, and thus the labeling function will be a function into one dimension.
The strict normal modes of the system can be identified by the direction of the velocity vector $\qdot$ when it passes through the equilibrium.
Note that $\qdot$ and $-\qdot$ identify the same mode and that the space of directions of a vector is not homeomorphic to the real numbers.
Consequently, we design a function $\theta: \mathcal{Q} \rightarrow \mathbb{S}^1$.

\begin{figure}
	\centering
	\vspace{.3cm}
	\def\svgwidth{.5\linewidth}\tiny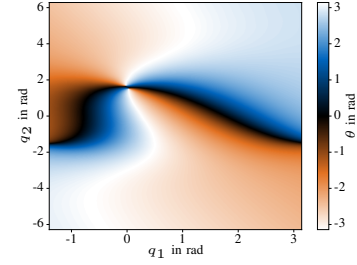
	\caption{Mode labeling function}
	\label{fig:fig_label}
\end{figure}

Detailed derivations on the labeling function $\theta(\qq)$ are attached in the appendix App.~\ref{app:labeling_function}.
We again used machine learning techniques to represent the function $\theta(\qq)$ by using a neural network model.
Fig.~\ref{fig:fig_label} shows a color-coded rendering of the labeling function for our example.
One can recognize from the rendering that, as expected, the level sets of $\theta(\qq)$ correspond to the strict modes of the system (see Fig.~\ref{subfig:pp_phase}).

Based on the labeling function $\theta(\qq)$ we derive the controller \begin{equation}\label{eq:mode_selection_raw}
	\tilde{\ttau}_\theta = \M(\qq) \frac{\partial \theta}{\partial \qq} \left(-k_\theta \Delta\theta(\q, \theta_d) - d_\theta(\qq)\frac{\partial \theta}{\partial \qq\tran}\qdot\right),
\end{equation}
where $k_\theta$ is a proportional gain and the deviation from the desired strict mode $\theta_d$ is computed by the function $\Delta\theta(\q, \theta_d)$.
The equations for $\Delta\theta(\qq, \theta_d)$ (\ref{eq:delta_theta}) and $\frac{\partial \theta}{\partial \qq}$ (\ref{eq:theta_jacobian}) can also be found in App.~\ref{app:labeling_function}.
The damping coefficient $d_\theta(\qq)$ (\ref{eq:damping_design}) is computed based on a desired damping ratio $\zeta$.

As the labeling function cannot exist at the equilibrium and does not perform well very close the equilibrium, we only enable the controller (\ref{eq:mode_selection_raw}) when the system is outside an $\varepsilon$-ball centered at the equilibrium, i.e., we set \begin{equation}\label{eq:mode_selection}
	\ttau_\theta(\q, \qdot) = \begin{cases}
		\nnull & \text{if } \Vert\qq - \qeq\Vert_2 \le \varepsilon,\\
		\tilde{\ttau}_\theta(\q, \qdot) & \text{otherwise}.
	\end{cases}	
\end{equation}
The mode selection controller also allows to switch between various modes during operation.
In (\ref{eq:delta_theta}) the desired mode is specified by the real number $\theta_d$.
Using this value the desired mode can be selected.

\begin{figure}
	\centering
	\subfloat[\label{subfig:mult_phase}]{
		\def\svgwidth{.49\columnwidth}\tiny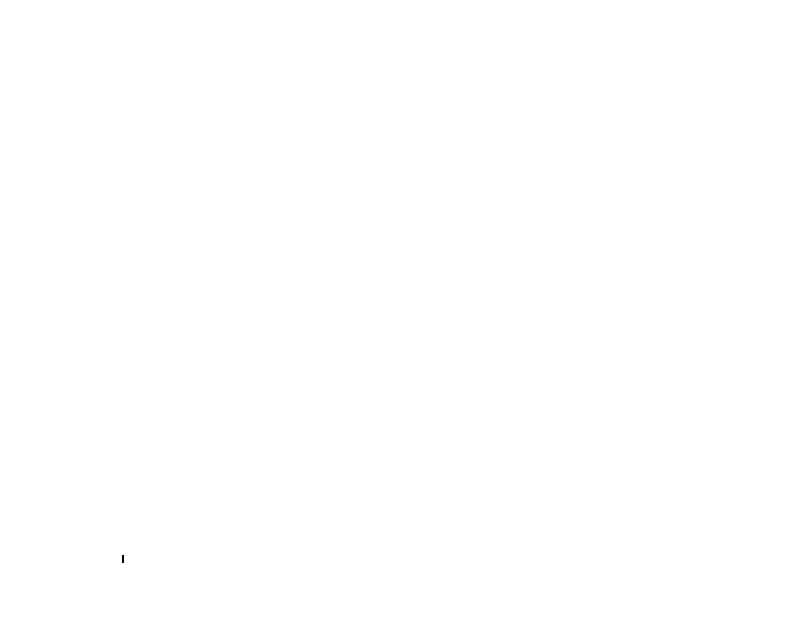
	}
	\subfloat[\label{subfig:mult_q}]{
		\def\svgwidth{.49\columnwidth}\tiny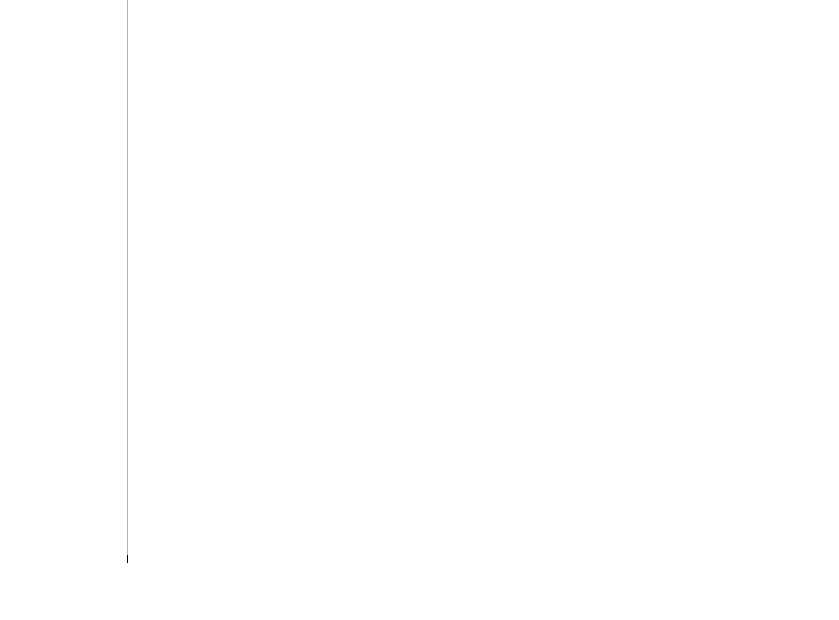
	}\\
	\subfloat[\label{subfig:mult_tau}]{
		\def\svgwidth{.49\columnwidth}\tiny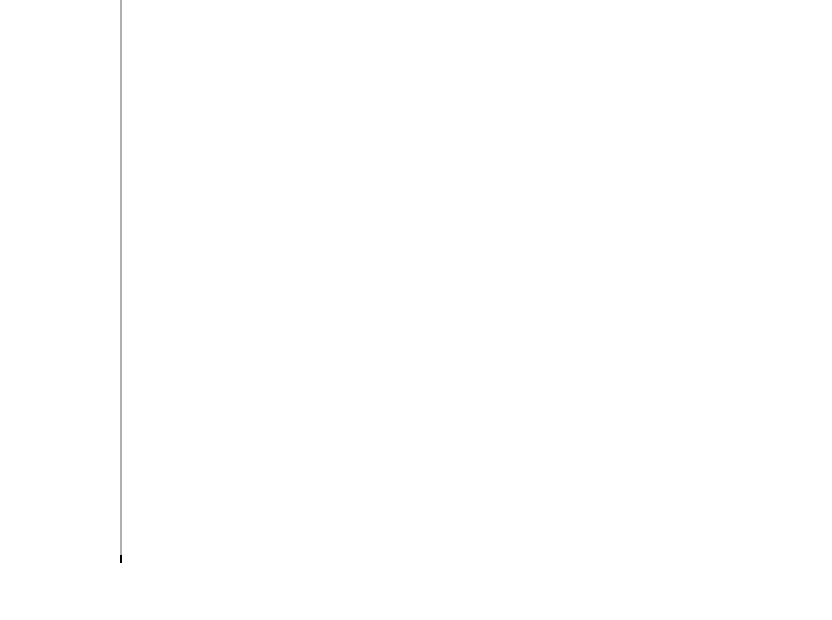
	}
	\subfloat[\label{subfig:mult_label}]{
		\def\svgwidth{.49\columnwidth}\tiny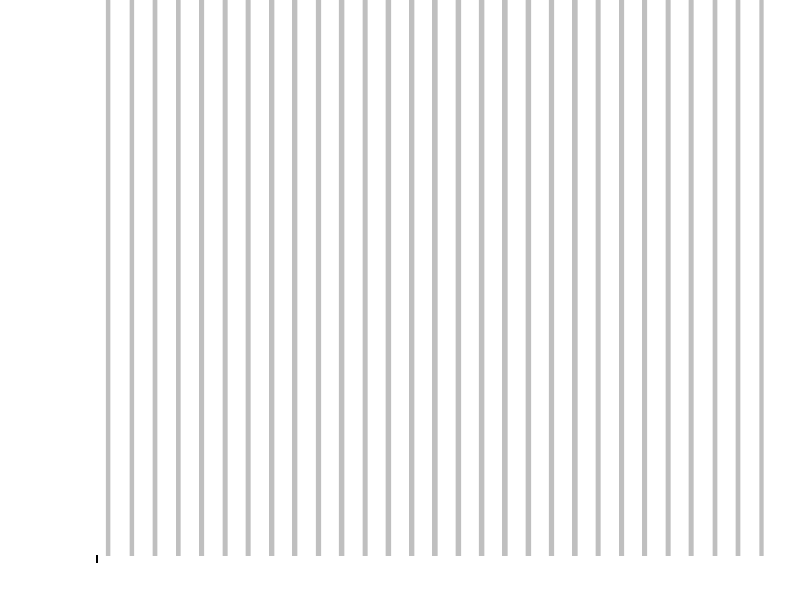
	}
	\caption{Simulation results with activated mode selection controller when switching the desired mode step-wisely.}
	\label{fig:mult_label}
\end{figure}

For the experiments, we set $k_\theta = 1$, $\zeta = 0.7$ and $\varepsilon = 0.1$ and apply the controller (\ref{eq:mode_selection}) together with the optimized potential on the system.
We increase $\theta_d$ step-wisely by an amount of $\frac{\pi}{6}$ every $15s$.
The results are shown in Fig.~\ref{fig:mult_label}.
Compared to the torques due to the potential, the controller torques are quite low also when switching between modes.
Fig.~\ref{subfig:mult_label} shows a time evolution of the function $\theta(\qq(t))$.
Especially in this plot it becomes clear that the function cannot be used at the equilibrium.
Each of the spikes in Fig.~\ref{subfig:mult_label} corresponds to a passage through the equilibrium, where the function value is undefined.
This does not compromise the performance as the controller (\ref{eq:mode_selection}) is disabled when close to the equilibrium (gray areas).

\subsection{Energy Regulation}
The task of the energy regulation controller is to swing up the system to the desired energy level.
Additionally, during operation on the mode the energy regulation controller will compensate for energy losses by reinjecting dissipated energy.

The mechanical system in combination with the mode selection controller $\ttau_\theta$ will already be stabilized onto the desired mode.
Therefore, negative damping can be used to inject energy into the system.
We scale the amount of negative damping by the difference of the current energy estimate to the target energy $E_d$ and obtain
\begin{equation}
	\tau_{\text{nD}}(\qq, \qdot) = \beta_{\text{nD}}\left(E_{d} - E(\qq, \qdot)\right)\qdot,
\end{equation}
where $\beta_{\text{nD}}$ is a scalar constant and 
\begin{equation}\label{eq:energy}
	E(\q, \qdot) = \frac{1}{2}\qdot\tran \MM(\qq) \qdot + U(\qq).
\end{equation}
This controller will in general not be compatible with the modes, i.e., the accelerations due to torques by this controller will not be tangent to the mode.

Let us consider the energy regulation controller by Della Santina et al. \cite[III.D]{Santina2021} as another candidate controller \begin{equation}\label{eq:energy_regulation}
	\ttau_{E}(\qq, \qdot) = \beta_E \MM(\qq) \left(E_d - E(\q, \qdot)\right)\qdot.
\end{equation}
where $\beta_E$ is another scalar parameter.
The controller (\ref{eq:energy_regulation}) accelerates or decelerates the system along the velocity vector.
Therefore, it will be tangent to the mode if we are on a mode.

We simulate the system with the potential $U_e$, the mode selection controller $\ttau_\theta$ and additionally model joint friction by a diagonal damping matrix $\DD = 0.3\II$.
Further, we initialize the system at a configuration $\q_0$ and set the initial velocity to zero.
Our chosen $\q_0$ corresponds to the energy level of $E_0 = 5.0\text{J}$.
The target energy level is set to $E_d = 20.0\text{J}$.
We test the two candidate controllers $\ttau_{\text{nD}}$ and $\ttau_{E}$ on this problem setting.
The constant $\beta_{\text{nD}}$ is set to $\beta_{\text{nD}} = 0.4$ and we have tuned $\beta_E$ such that both controllers let the system converge to the same steady state energy after the same time.
We got a value of $\beta_E = 0.13$.

\begin{figure}[h]
	\centering
	\subfloat[\label{subfig:dampex_phase}]{
		\def\svgwidth{.47\columnwidth}\tiny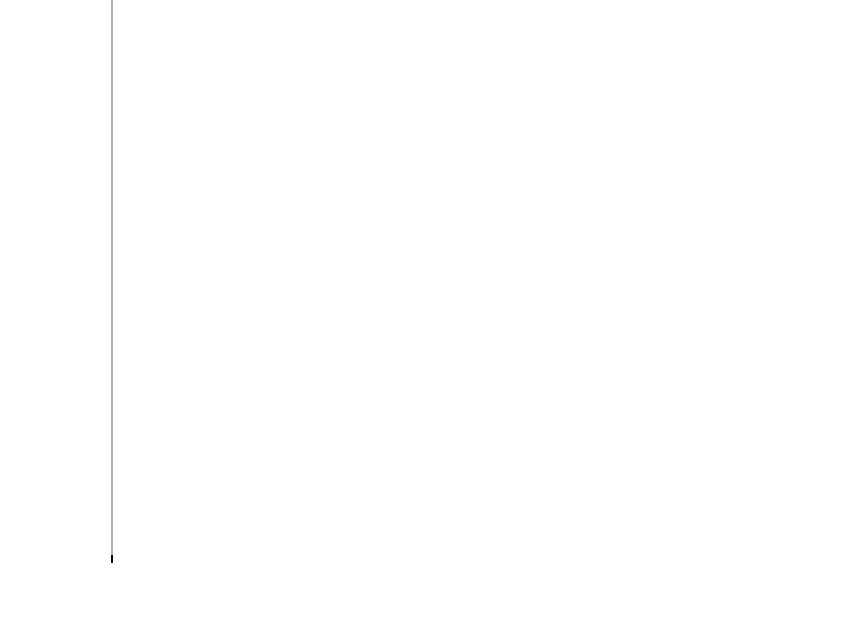
	}
	\subfloat[\label{subfig:dampex_q}]{
		\def\svgwidth{.47\columnwidth}\tiny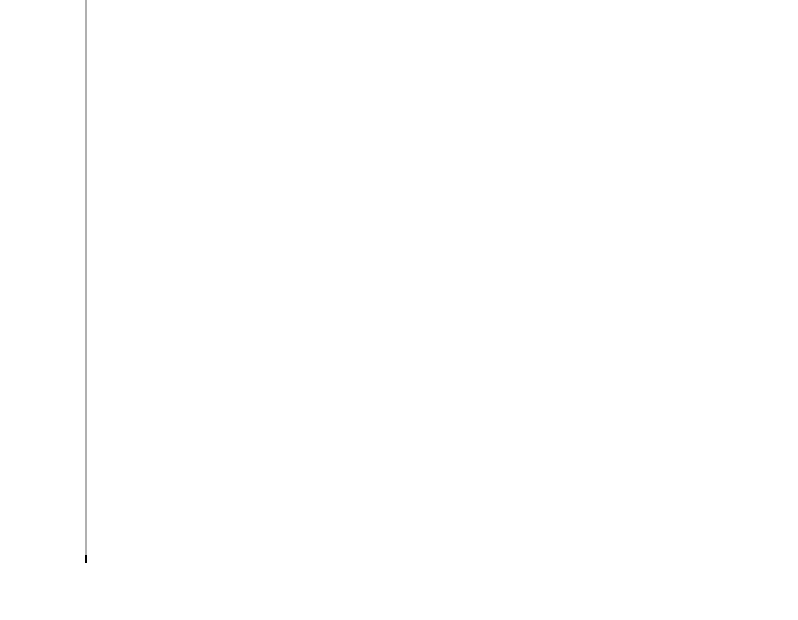
	}\\
	\subfloat[$\tau_{\text{nD}}$\label{subfig:dampex_naive}]{
		\def\svgwidth{.47\columnwidth}\tiny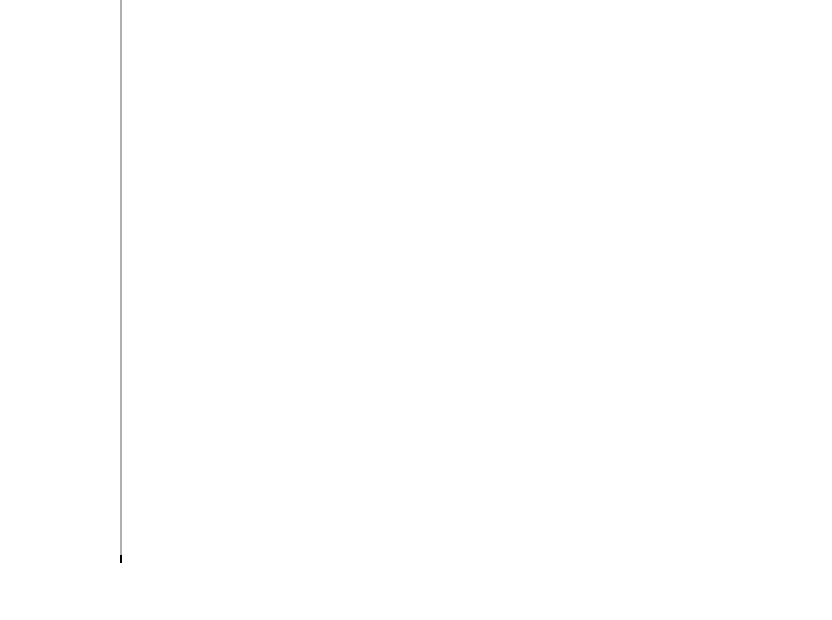
	}
	\subfloat[$\tau_{E}$\label{subfig:dampex_my}]{
		\def\svgwidth{.47\columnwidth}\tiny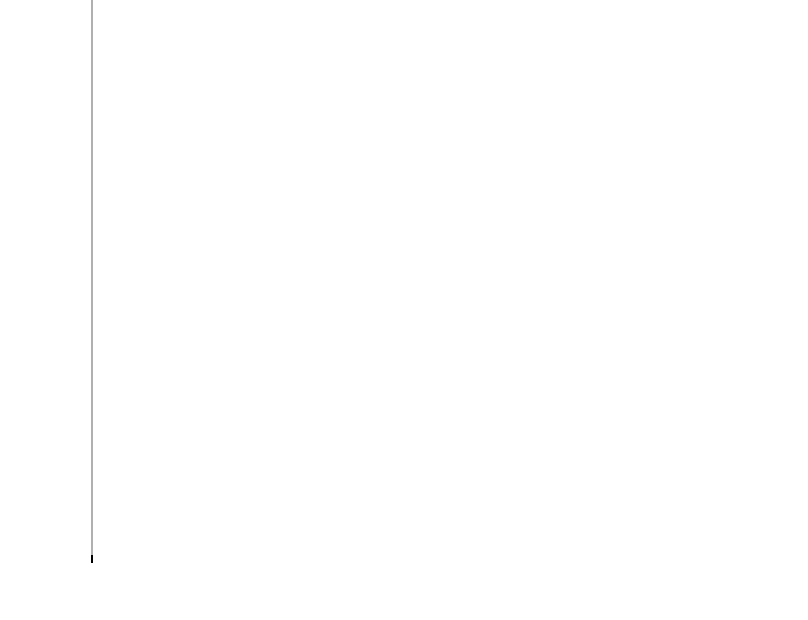
	}
	\caption{Damped system including the optimized potential, the mode selection controller and two types of energy regulation.}
	\label{fig:dampex}
\end{figure}

The simulation results are shown in Fig.~\ref{fig:dampex}.
It turns out that the two candidate controllers have advantages and disadvantages.

On the one hand, the mode-compatible controller $\ttau_E$ performs much better during the swing up phase.
As can be seen in Fig.~\ref{subfig:dampex_phase} (blue line) it accelerates the system along the mode.
The control actions by $\ttau_\theta$ and $\ttau_E$ are dynamically decoupled, i.e., a control action by $\ttau_\theta$ will not create an immediate change in potential energy and a control action by $\ttau_E$ will not pull the system off the mode.
In contrast, the simple negative-damping controller $\ttau_{\text{nD}}$ shoots the system into a different direction (yellow line in Fig.~\ref{subfig:dampex_phase}), which must be compensated by the mode stabilization controller $\ttau_\theta$ (Fig.~\ref{subfig:dampex_naive}).

On the other hand, after the swing-up phase the ranking of performances is swapped.
The negative damping controller $\ttau_{\text{nD}}$ performs much better at compensating the damping as the structure is the same as the isotropic system damping.
The mode-compatible controller $\ttau_E$ cannot compensate the system damping completely, as it has a different structure.
Corrective torques by the mode selection controller $\ttau_\theta$ are required, as the system damping is not compatible with the modes (Fig.~\ref{subfig:dampex_my}).

We want to highlight, that the negative-damping controller performs so well after the swing up phase only because the modeled joint friction matches the structure of the controller.
This is a match, which is quite unrealistic for real world scenarios.
Therefore, for applications on hardware, a classical damping compensation should be used and the mode-compatible controller $\ttau_E$ should be used for energy regulation and swing up.

\subsection{Sensitivity Analysis}
The final potential function $U_e$ is represented by a neural network, which provides a lot of flexibility and easily accessible algorithms for training.
When moving from neural networks to actual mechanical elastic elements it will be hard to \emph{exactly} implement the optimal potential.
A mechanically implemented potential $\tilde{U}_e$ will in general not exactly resemble the optimal potential $U_e$.
Although the mechanical implementation is not addressed in this paper, we perform a sensitivity analysis here.
We add another parasitic potential to the optimal one and obtain the disturbed potential \begin{equation}
	\tilde{U}_e(\qq) = U_e(\qq) + \frac{\epsilon}{2} (\qq - \qq_{\text{eq}})\tran \begin{bmatrix}
		\kappa_1 & 0\\0 & \kappa_2	
	\end{bmatrix} (\qq - \qq_{\text{eq}}),
\end{equation}
i.e., we add two linear parasitic springs with stiffness $\epsilon\kappa_i$ to the system.
The constants $\kappa_i$ are selected such that \begin{equation}
	\kappa_i = \frac{\tau_i^{\text{max}}}{\Delta q_i^{\text{max}}} 
	\end{equation}
where $\tau_i^{\text{max}} = \max\limits_{\bar\Qc} \frac{\partial U_e}{\partial q_i}$ and $\Delta q_i^{\text{max}} = \max\limits_{\bar\Qc} |q_i - q_{\text{eq},i}|$, i.e., when setting $\epsilon = 1$ the torques to the parasitic potential grow as high as the ones due to the optimal potential $U_e$.

\begin{figure}[h]
	\centering
	\subfloat[\label{subfig:sensitivity_phase}]{
		\def\svgwidth{.47\columnwidth}\tiny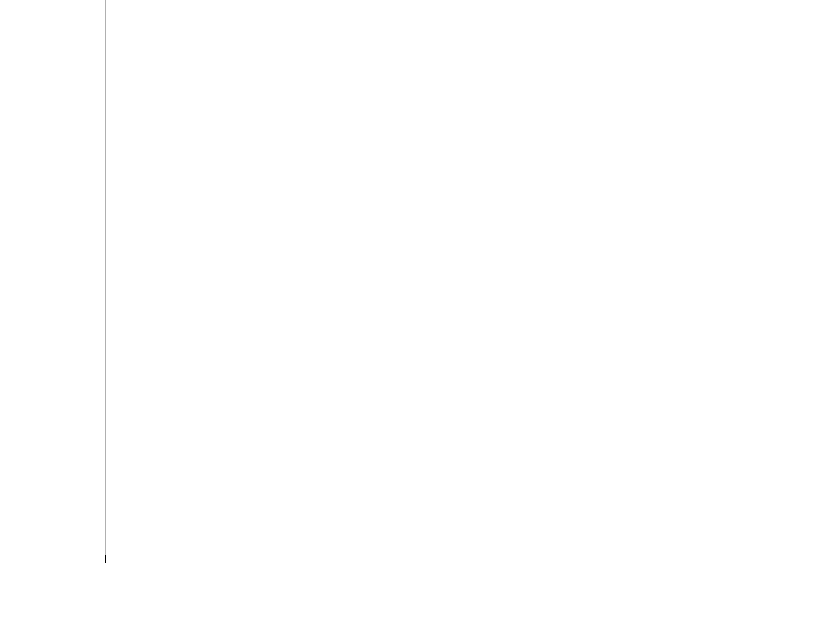
	}
	\subfloat[\label{subfig:sensitivity_q}]{
		\def\svgwidth{.47\columnwidth}\tiny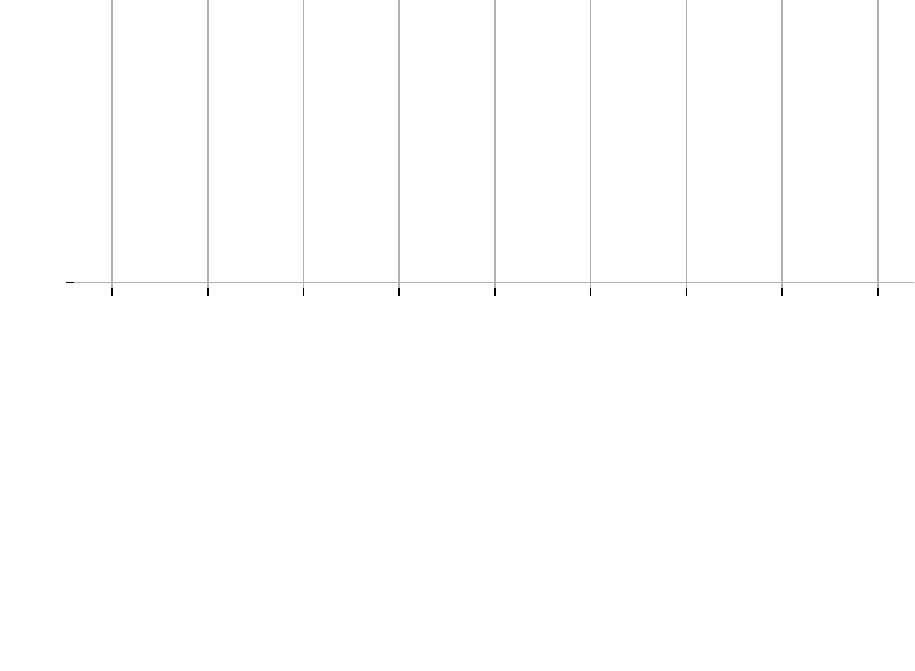
		}\\
	\subfloat[$\epsilon=0.01$]{
		\def\svgwidth{0.31\columnwidth}\tiny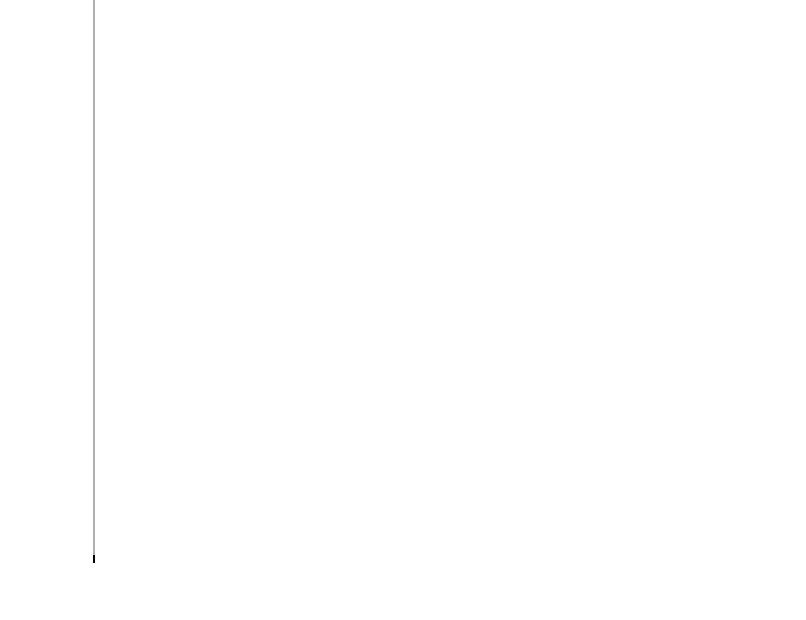
	}
	\subfloat[$\epsilon=0.1$]{
		\def\svgwidth{0.31\columnwidth}\tiny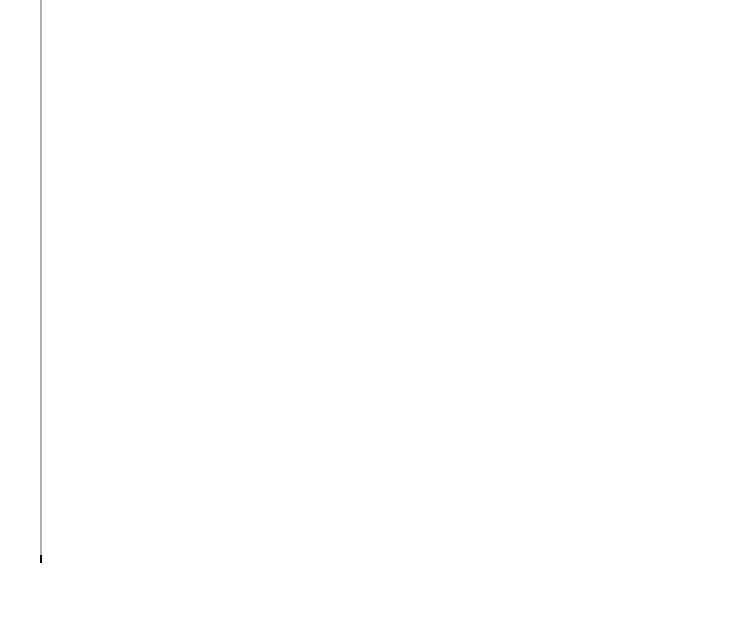
	}
	\subfloat[$\epsilon=1$]{
		\def\svgwidth{0.31\columnwidth}\tiny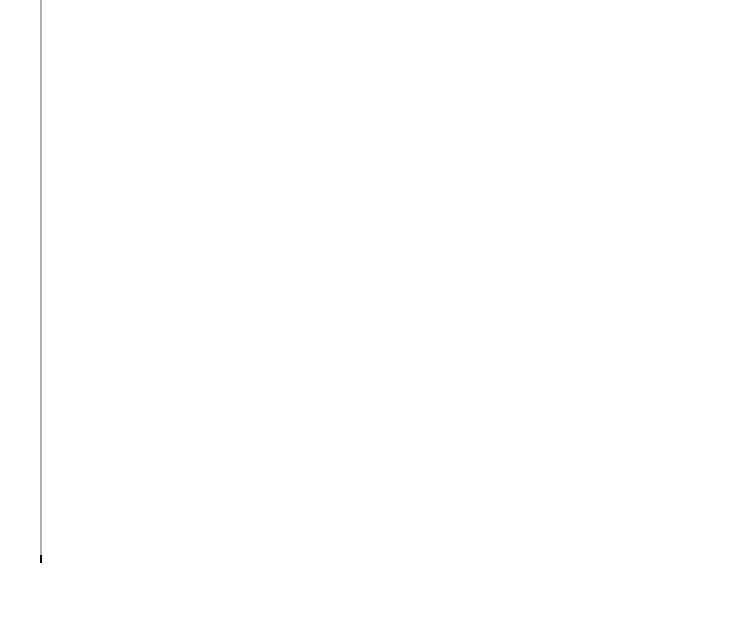
	}
	\caption{Results for the sensitivity analysis. The parasitical potential is increased over the experiments. The top row shows the evolution of the system in terms of the joint angles. The bottom row shows power flow due to the individual torques sources for three different values of $\epsilon$.}
	\label{fig:sensitivity_analysis}
\end{figure}

The system is simulated for different settings of $\epsilon$.
Note that for $\epsilon = 0$ the system is the optimal one, i.e., the optimal potential $U_e$ is present.
Both controllers, $\ttau_\theta$ and $\ttau_E$, are active with the same settings as before.
Results are shown in Fig.~\ref{fig:sensitivity_analysis}.
Up to a value of $\epsilon = 0.1$, the trajectory traced out in the $q_1 q_2$-plane (Fig.~\ref{subfig:sensitivity_phase}) does not change significantly.
However, the frequency of oscillation is slightly altered (Fig.~\ref{subfig:sensitivity_q}).
For $\epsilon=0.5$ and $\epsilon=1.0$, we observe larger disturbances.
We want to highlight that we introduced quite high parasitic forces.
For $\epsilon = 1$ the torques due to the parasitic potential are on average as high as the torques due to the optimal potential $U_e$. 

From the power plots (Fig.~\ref{fig:sensitivity_analysis}c-e) can be concluded that increased control action is required for increasing $\epsilon$.
Especially the energy regulation controller (black lines) is removing and adding energy periodically.
This is no surprise, as the control law (\ref{eq:energy_regulation}) relies on the computation of the energy (\ref{eq:energy}).
Within (\ref{eq:energy}) the optimal potential $U_e$ is used as the truly implemented potential $\tilde{U}_e$ is generally not known.
Consequently, the energy regulation controller uses a false estimate of the energy.

In a next step we compute the total energy flux introduced by the individual torque components.
Therefore, we integrate the absolute value of the powers using numerical integration \begin{equation}
	\int |P_\theta| \approx \sum \left|\ttau_\theta\tran \qdot\right| \delta t.
\end{equation}
Likewise we compute the total energy fluxes $\int|P_{\tilde{U}}|$ for the disturbed potential and $\int|P_E|$ for the energy regulation controller.
Table~\ref{tab:sensitivity_analysis} shows the results.
In a physical system the energy flux in the first row is handled by elastic elements, which store and release energy.
Therefore, for ideal springs, the first row does not effect the energy efficiency.
The last two rows correspond to energy fluxes through the actuators of the system.
As electrical actuators and their drivers usually do not recover energy, this energy is lost.
Hence, for energy efficient system the values in the last two rows should be as small as possible.

\begin{table}[h]
	\caption{Comparison of total energy fluxes}\label{tab:sensitivity_analysis}
	\centering
	\vspace{-.3cm}
	\begin{small}
		\begin{tabular}{c|c|c|c}
			$\epsilon$ & $\int |P_{\tilde{U}}|$ & $\int |P_\theta|$ & $\int |P_E|$ \\\hline\hline
			0    & 354.56 & 2.86  & 0.03   \\\hline
			0.01 & 356.23 & 2.85  & 1.87   \\\hline
			0.05 & 363.25 & 2.94  & 9.44   \\\hline
			0.1  & 372.43 & 3.74  & 18.75  \\\hline
			0.5  & 447.18 & 11.72 & 89.26  \\\hline
			1.0  & 565.60 & 20.92 & 181.83 \\\hline
		\end{tabular}
	\end{small}
\end{table}

We conclude that the final system admits certain tolerances in the potential implemented by mechanical elements.
The controllers are able to compensate inaccuracies in the potential to a certain extent.
On the one hand, for moderate inaccuracies of the potential, the system trajectories change shape only slightly.
However, the timing and frequency of oscillation changes.
The aim of this work was not to do trajectory tracking, but to basically retain as much as possible from the dynamics of the mechanical system.
On the other hand, the energy efficiency decreases when the implemented potential deviates from the optimized one.

%% file: pdftex/label.pdf_tex
\begingroup%
  \makeatletter%
  \providecommand\color[2][]{%
    \errmessage{(Inkscape) Color is used for the text in Inkscape, but the package 'color.sty' is not loaded}%
    \renewcommand\color[2][]{}%
  }%
  \providecommand\transparent[1]{%
    \errmessage{(Inkscape) Transparency is used (non-zero) for the text in Inkscape, but the package 'transparent.sty' is not loaded}%
    \renewcommand\transparent[1]{}%
  }%
  \providecommand\rotatebox[2]{#2}%
  \newcommand*\fsize{\dimexpr\f@size pt\relax}%
  \newcommand*\lineheight[1]{\fontsize{\fsize}{#1\fsize}\selectfont}%
  \ifx\svgwidth\undefined%
    \setlength{\unitlength}{427.78988647bp}%
    \ifx\svgscale\undefined%
      \relax%
    \else%
      \setlength{\unitlength}{\unitlength * \real{\svgscale}}%
    \fi%
  \else%
    \setlength{\unitlength}{\svgwidth}%
  \fi%
  \global\let\svgwidth\undefined%
  \global\let\svgscale\undefined%
  \makeatother%
  \begin{picture}(1,0.75685219)%
    \lineheight{1}%
    \setlength\tabcolsep{0pt}%
    \put(0,0){\includegraphics[width=\unitlength,page=1]{label.pdf}}%
    \put(0.14986612,0.03748694){\makebox(0,0)[t]{\lineheight{1.25}\smash{\begin{tabular}[t]{c}−1\end{tabular}}}}%
    \put(0,0){\includegraphics[width=\unitlength,page=2]{label.pdf}}%
    \put(0.31591469,0.03748694){\makebox(0,0)[t]{\lineheight{1.25}\smash{\begin{tabular}[t]{c}0\end{tabular}}}}%
    \put(0,0){\includegraphics[width=\unitlength,page=3]{label.pdf}}%
    \put(0.48196324,0.03748694){\makebox(0,0)[t]{\lineheight{1.25}\smash{\begin{tabular}[t]{c}1\end{tabular}}}}%
    \put(0,0){\includegraphics[width=\unitlength,page=4]{label.pdf}}%
    \put(0.64801175,0.03748694){\makebox(0,0)[t]{\lineheight{1.25}\smash{\begin{tabular}[t]{c}2\end{tabular}}}}%
    \put(0,0){\includegraphics[width=\unitlength,page=5]{label.pdf}}%
    \put(0.81406034,0.03748694){\makebox(0,0)[t]{\lineheight{1.25}\smash{\begin{tabular}[t]{c}3\end{tabular}}}}%
    \put(0.46050915,0.005513){\makebox(0,0)[t]{\lineheight{1.25}\smash{\begin{tabular}[t]{c}$q_1$ in rad\end{tabular}}}}%
    \put(0,0){\includegraphics[width=\unitlength,page=6]{label.pdf}}%
    \put(0.06708352,0.08386434){\makebox(0,0)[rt]{\lineheight{1.25}\smash{\begin{tabular}[t]{r}−6\end{tabular}}}}%
    \put(0,0){\includegraphics[width=\unitlength,page=7]{label.pdf}}%
    \put(0.06708352,0.19184477){\makebox(0,0)[rt]{\lineheight{1.25}\smash{\begin{tabular}[t]{r}−4\end{tabular}}}}%
    \put(0,0){\includegraphics[width=\unitlength,page=8]{label.pdf}}%
    \put(0.06708352,0.29982519){\makebox(0,0)[rt]{\lineheight{1.25}\smash{\begin{tabular}[t]{r}−2\end{tabular}}}}%
    \put(0,0){\includegraphics[width=\unitlength,page=9]{label.pdf}}%
    \put(0.06708352,0.40780561){\makebox(0,0)[rt]{\lineheight{1.25}\smash{\begin{tabular}[t]{r}0\end{tabular}}}}%
    \put(0,0){\includegraphics[width=\unitlength,page=10]{label.pdf}}%
    \put(0.06708352,0.51578601){\makebox(0,0)[rt]{\lineheight{1.25}\smash{\begin{tabular}[t]{r}2\end{tabular}}}}%
    \put(0,0){\includegraphics[width=\unitlength,page=11]{label.pdf}}%
    \put(0.06708352,0.62376645){\makebox(0,0)[rt]{\lineheight{1.25}\smash{\begin{tabular}[t]{r}4\end{tabular}}}}%
    \put(0,0){\includegraphics[width=\unitlength,page=12]{label.pdf}}%
    \put(0.06708352,0.73174688){\makebox(0,0)[rt]{\lineheight{1.25}\smash{\begin{tabular}[t]{r}6\end{tabular}}}}%
    \put(0.01776025,0.41668667){\rotatebox{90}{\makebox(0,0)[t]{\lineheight{1.25}\smash{\begin{tabular}[t]{c}$q_2$ in rad\end{tabular}}}}}%
    \put(0,0){\includegraphics[width=\unitlength,page=13]{label.pdf}}%
    \put(0.93499063,0.08386313){\makebox(0,0)[lt]{\lineheight{1.25}\smash{\begin{tabular}[t]{l}−3\end{tabular}}}}%
    \put(0,0){\includegraphics[width=\unitlength,page=14]{label.pdf}}%
    \put(0.93499063,0.19184384){\makebox(0,0)[lt]{\lineheight{1.25}\smash{\begin{tabular}[t]{l}−2\end{tabular}}}}%
    \put(0,0){\includegraphics[width=\unitlength,page=15]{label.pdf}}%
    \put(0.93499063,0.29982458){\makebox(0,0)[lt]{\lineheight{1.25}\smash{\begin{tabular}[t]{l}−1\end{tabular}}}}%
    \put(0,0){\includegraphics[width=\unitlength,page=16]{label.pdf}}%
    \put(0.93499063,0.40780529){\makebox(0,0)[lt]{\lineheight{1.25}\smash{\begin{tabular}[t]{l}0\end{tabular}}}}%
    \put(0,0){\includegraphics[width=\unitlength,page=17]{label.pdf}}%
    \put(0.93499063,0.51578601){\makebox(0,0)[lt]{\lineheight{1.25}\smash{\begin{tabular}[t]{l}1\end{tabular}}}}%
    \put(0,0){\includegraphics[width=\unitlength,page=18]{label.pdf}}%
    \put(0.93499063,0.62376675){\makebox(0,0)[lt]{\lineheight{1.25}\smash{\begin{tabular}[t]{l}2\end{tabular}}}}%
    \put(0,0){\includegraphics[width=\unitlength,page=19]{label.pdf}}%
    \put(0.93499063,0.73174749){\makebox(0,0)[lt]{\lineheight{1.25}\smash{\begin{tabular}[t]{l}3\end{tabular}}}}%
    \put(0.99656436,0.41668664){\rotatebox{90}{\makebox(0,0)[t]{\lineheight{1.25}\smash{\begin{tabular}[t]{c}$\theta$ in rad\end{tabular}}}}}%
    \put(0,0){\includegraphics[width=\unitlength,page=20]{label.pdf}}%
  \end{picture}%
\endgroup%

%% file: pdftex/mult_phase.pdf_tex
\begingroup%
  \makeatletter%
  \providecommand\color[2][]{%
    \errmessage{(Inkscape) Color is used for the text in Inkscape, but the package 'color.sty' is not loaded}%
    \renewcommand\color[2][]{}%
  }%
  \providecommand\transparent[1]{%
    \errmessage{(Inkscape) Transparency is used (non-zero) for the text in Inkscape, but the package 'transparent.sty' is not loaded}%
    \renewcommand\transparent[1]{}%
  }%
  \providecommand\rotatebox[2]{#2}%
  \newcommand*\fsize{\dimexpr\f@size pt\relax}%
  \newcommand*\lineheight[1]{\fontsize{\fsize}{#1\fsize}\selectfont}%
  \ifx\svgwidth\undefined%
    \setlength{\unitlength}{385.06097412bp}%
    \ifx\svgscale\undefined%
      \relax%
    \else%
      \setlength{\unitlength}{\unitlength * \real{\svgscale}}%
    \fi%
  \else%
    \setlength{\unitlength}{\svgwidth}%
  \fi%
  \global\let\svgwidth\undefined%
  \global\let\svgscale\undefined%
  \makeatother%
  \begin{picture}(1,0.77629925)%
    \lineheight{1}%
    \setlength\tabcolsep{0pt}%
    \put(0,0){\includegraphics[width=\unitlength,page=1]{mult_phase.pdf}}%
    \put(0.15320755,0.03976542){\makebox(0,0)[t]{\lineheight{1.25}\smash{\begin{tabular}[t]{c}−1\end{tabular}}}}%
    \put(0,0){\includegraphics[width=\unitlength,page=2]{mult_phase.pdf}}%
    \put(0.35741731,0.03976542){\makebox(0,0)[t]{\lineheight{1.25}\smash{\begin{tabular}[t]{c}0\end{tabular}}}}%
    \put(0,0){\includegraphics[width=\unitlength,page=3]{mult_phase.pdf}}%
    \put(0.56162707,0.03976542){\makebox(0,0)[t]{\lineheight{1.25}\smash{\begin{tabular}[t]{c}1\end{tabular}}}}%
    \put(0,0){\includegraphics[width=\unitlength,page=4]{mult_phase.pdf}}%
    \put(0.76583683,0.03976542){\makebox(0,0)[t]{\lineheight{1.25}\smash{\begin{tabular}[t]{c}2\end{tabular}}}}%
    \put(0,0){\includegraphics[width=\unitlength,page=5]{mult_phase.pdf}}%
    \put(0.97004659,0.03976542){\makebox(0,0)[t]{\lineheight{1.25}\smash{\begin{tabular}[t]{c}3\end{tabular}}}}%
    \put(0,0){\includegraphics[width=\unitlength,page=6]{mult_phase.pdf}}%
    \put(0.05334471,0.15801293){\makebox(0,0)[rt]{\lineheight{1.25}\smash{\begin{tabular}[t]{r}−2\end{tabular}}}}%
    \put(0,0){\includegraphics[width=\unitlength,page=7]{mult_phase.pdf}}%
    \put(0.05334471,0.3046669){\makebox(0,0)[rt]{\lineheight{1.25}\smash{\begin{tabular}[t]{r}0\end{tabular}}}}%
    \put(0,0){\includegraphics[width=\unitlength,page=8]{mult_phase.pdf}}%
    \put(0.05334471,0.45132086){\makebox(0,0)[rt]{\lineheight{1.25}\smash{\begin{tabular}[t]{r}2\end{tabular}}}}%
    \put(0,0){\includegraphics[width=\unitlength,page=9]{mult_phase.pdf}}%
    \put(0.05334471,0.59797483){\makebox(0,0)[rt]{\lineheight{1.25}\smash{\begin{tabular}[t]{r}4\end{tabular}}}}%
    \put(0,0){\includegraphics[width=\unitlength,page=10]{mult_phase.pdf}}%
    \put(0.05334471,0.74462879){\makebox(0,0)[rt]{\lineheight{1.25}\smash{\begin{tabular}[t]{r}6\end{tabular}}}}%
    \put(0,0){\includegraphics[width=\unitlength,page=11]{mult_phase.pdf}}%
    \put(0.54984013,0.70739352){\makebox(0,0)[lt]{\lineheight{1.25}\smash{\begin{tabular}[t]{l}Geodesics\end{tabular}}}}%
    \put(0,0){\includegraphics[width=\unitlength,page=12]{mult_phase.pdf}}%
    \put(0.54984013,0.66927454){\makebox(0,0)[lt]{\lineheight{1.25}\smash{\begin{tabular}[t]{l}Simulated Trajectory\end{tabular}}}}%
    \put(0,0){\includegraphics[width=\unitlength,page=13]{mult_phase.pdf}}%
    \put(0.54984013,0.63115556){\makebox(0,0)[lt]{\lineheight{1.25}\smash{\begin{tabular}[t]{l}Isolines of $U(\q)$\end{tabular}}}}%
    \put(0,0){\includegraphics[width=\unitlength,page=14]{mult_phase.pdf}}%
    \put(0.54984013,0.59303666){\makebox(0,0)[lt]{\lineheight{1.25}\smash{\begin{tabular}[t]{l}Initial Condition\end{tabular}}}}%
    \put(0.51233269,0.00431092){\makebox(0,0)[lt]{\lineheight{1.25}\smash{\begin{tabular}[t]{l}$q_1$ in rad\end{tabular}}}}%
    \put(0.00923015,0.41413248){\rotatebox{90}{\makebox(0,0)[lt]{\lineheight{1.25}\smash{\begin{tabular}[t]{l}$q_2$ in rad\end{tabular}}}}}%
  \end{picture}%
\endgroup%

%% file: pdftex/mult_q.pdf_tex
\begingroup%
  \makeatletter%
  \providecommand\color[2][]{%
    \errmessage{(Inkscape) Color is used for the text in Inkscape, but the package 'color.sty' is not loaded}%
    \renewcommand\color[2][]{}%
  }%
  \providecommand\transparent[1]{%
    \errmessage{(Inkscape) Transparency is used (non-zero) for the text in Inkscape, but the package 'transparent.sty' is not loaded}%
    \renewcommand\transparent[1]{}%
  }%
  \providecommand\rotatebox[2]{#2}%
  \newcommand*\fsize{\dimexpr\f@size pt\relax}%
  \newcommand*\lineheight[1]{\fontsize{\fsize}{#1\fsize}\selectfont}%
  \ifx\svgwidth\undefined%
    \setlength{\unitlength}{402.48016357bp}%
    \ifx\svgscale\undefined%
      \relax%
    \else%
      \setlength{\unitlength}{\unitlength * \real{\svgscale}}%
    \fi%
  \else%
    \setlength{\unitlength}{\svgwidth}%
  \fi%
  \global\let\svgwidth\undefined%
  \global\let\svgscale\undefined%
  \makeatother%
  \begin{picture}(1,0.73899338)%
    \lineheight{1}%
    \setlength\tabcolsep{0pt}%
    \put(0,0){\includegraphics[width=\unitlength,page=1]{mult_q.pdf}}%
    \put(0.1520395,0.03433645){\makebox(0,0)[t]{\lineheight{1.25}\smash{\begin{tabular}[t]{c}0\end{tabular}}}}%
    \put(0,0){\includegraphics[width=\unitlength,page=2]{mult_q.pdf}}%
    \put(0.31336649,0.03433645){\makebox(0,0)[t]{\lineheight{1.25}\smash{\begin{tabular}[t]{c}20\end{tabular}}}}%
    \put(0,0){\includegraphics[width=\unitlength,page=3]{mult_q.pdf}}%
    \put(0.47469347,0.03433645){\makebox(0,0)[t]{\lineheight{1.25}\smash{\begin{tabular}[t]{c}40\end{tabular}}}}%
    \put(0,0){\includegraphics[width=\unitlength,page=4]{mult_q.pdf}}%
    \put(0.63602045,0.03433645){\makebox(0,0)[t]{\lineheight{1.25}\smash{\begin{tabular}[t]{c}60\end{tabular}}}}%
    \put(0,0){\includegraphics[width=\unitlength,page=5]{mult_q.pdf}}%
    \put(0.79734742,0.03433645){\makebox(0,0)[t]{\lineheight{1.25}\smash{\begin{tabular}[t]{c}80\end{tabular}}}}%
    \put(0,0){\includegraphics[width=\unitlength,page=6]{mult_q.pdf}}%
    \put(0.9586744,0.03433645){\makebox(0,0)[t]{\lineheight{1.25}\smash{\begin{tabular}[t]{c}100\end{tabular}}}}%
    \put(0.55535696,0.00035184){\makebox(0,0)[t]{\lineheight{1.25}\smash{\begin{tabular}[t]{c}Time in sec\end{tabular}}}}%
    \put(0,0){\includegraphics[width=\unitlength,page=7]{mult_q.pdf}}%
    \put(0.09431559,0.12808114){\makebox(0,0)[rt]{\lineheight{1.25}\smash{\begin{tabular}[t]{r}−1.5\end{tabular}}}}%
    \put(0,0){\includegraphics[width=\unitlength,page=8]{mult_q.pdf}}%
    \put(0.09431559,0.20325749){\makebox(0,0)[rt]{\lineheight{1.25}\smash{\begin{tabular}[t]{r}−1.0\end{tabular}}}}%
    \put(0,0){\includegraphics[width=\unitlength,page=9]{mult_q.pdf}}%
    \put(0.09431559,0.27843387){\makebox(0,0)[rt]{\lineheight{1.25}\smash{\begin{tabular}[t]{r}−0.5\end{tabular}}}}%
    \put(0,0){\includegraphics[width=\unitlength,page=10]{mult_q.pdf}}%
    \put(0.09431559,0.35361026){\makebox(0,0)[rt]{\lineheight{1.25}\smash{\begin{tabular}[t]{r}0.0\end{tabular}}}}%
    \put(0,0){\includegraphics[width=\unitlength,page=11]{mult_q.pdf}}%
    \put(0.09431559,0.42878661){\makebox(0,0)[rt]{\lineheight{1.25}\smash{\begin{tabular}[t]{r}0.5\end{tabular}}}}%
    \put(0,0){\includegraphics[width=\unitlength,page=12]{mult_q.pdf}}%
    \put(0.09431559,0.50396299){\makebox(0,0)[rt]{\lineheight{1.25}\smash{\begin{tabular}[t]{r}1.0\end{tabular}}}}%
    \put(0,0){\includegraphics[width=\unitlength,page=13]{mult_q.pdf}}%
    \put(0.09431559,0.5791394){\makebox(0,0)[rt]{\lineheight{1.25}\smash{\begin{tabular}[t]{r}1.5\end{tabular}}}}%
    \put(0,0){\includegraphics[width=\unitlength,page=14]{mult_q.pdf}}%
    \put(0.09431559,0.65431574){\makebox(0,0)[rt]{\lineheight{1.25}\smash{\begin{tabular}[t]{r}2.0\end{tabular}}}}%
    \put(0.01887709,0.40740933){\rotatebox{90}{\makebox(0,0)[t]{\lineheight{1.25}\smash{\begin{tabular}[t]{c}Angle in rad\end{tabular}}}}}%
    \put(0,0){\includegraphics[width=\unitlength,page=15]{mult_q.pdf}}%
    \put(0.43446433,0.15464734){\makebox(0,0)[lt]{\lineheight{1.25}\smash{\begin{tabular}[t]{l}$q_1$\end{tabular}}}}%
    \put(0,0){\includegraphics[width=\unitlength,page=16]{mult_q.pdf}}%
    \put(0.43446433,0.11812769){\makebox(0,0)[lt]{\lineheight{1.25}\smash{\begin{tabular}[t]{l}$q_2$\end{tabular}}}}%
  \end{picture}%
\endgroup%

%% file: pdftex/mult_tau.pdf_tex
\begingroup%
  \makeatletter%
  \providecommand\color[2][]{%
    \errmessage{(Inkscape) Color is used for the text in Inkscape, but the package 'color.sty' is not loaded}%
    \renewcommand\color[2][]{}%
  }%
  \providecommand\transparent[1]{%
    \errmessage{(Inkscape) Transparency is used (non-zero) for the text in Inkscape, but the package 'transparent.sty' is not loaded}%
    \renewcommand\transparent[1]{}%
  }%
  \providecommand\rotatebox[2]{#2}%
  \newcommand*\fsize{\dimexpr\f@size pt\relax}%
  \newcommand*\lineheight[1]{\fontsize{\fsize}{#1\fsize}\selectfont}%
  \ifx\svgwidth\undefined%
    \setlength{\unitlength}{399.30203247bp}%
    \ifx\svgscale\undefined%
      \relax%
    \else%
      \setlength{\unitlength}{\unitlength * \real{\svgscale}}%
    \fi%
  \else%
    \setlength{\unitlength}{\svgwidth}%
  \fi%
  \global\let\svgwidth\undefined%
  \global\let\svgscale\undefined%
  \makeatother%
  \begin{picture}(1,0.74487519)%
    \lineheight{1}%
    \setlength\tabcolsep{0pt}%
    \put(0,0){\includegraphics[width=\unitlength,page=1]{mult_tau.pdf}}%
    \put(0.14529041,0.03460975){\makebox(0,0)[t]{\lineheight{1.25}\smash{\begin{tabular}[t]{c}0\end{tabular}}}}%
    \put(0,0){\includegraphics[width=\unitlength,page=2]{mult_tau.pdf}}%
    \put(0.30790144,0.03460975){\makebox(0,0)[t]{\lineheight{1.25}\smash{\begin{tabular}[t]{c}20\end{tabular}}}}%
    \put(0,0){\includegraphics[width=\unitlength,page=3]{mult_tau.pdf}}%
    \put(0.47051246,0.03460975){\makebox(0,0)[t]{\lineheight{1.25}\smash{\begin{tabular}[t]{c}40\end{tabular}}}}%
    \put(0,0){\includegraphics[width=\unitlength,page=4]{mult_tau.pdf}}%
    \put(0.63312347,0.03460975){\makebox(0,0)[t]{\lineheight{1.25}\smash{\begin{tabular}[t]{c}60\end{tabular}}}}%
    \put(0,0){\includegraphics[width=\unitlength,page=5]{mult_tau.pdf}}%
    \put(0.79573448,0.03460975){\makebox(0,0)[t]{\lineheight{1.25}\smash{\begin{tabular}[t]{c}80\end{tabular}}}}%
    \put(0,0){\includegraphics[width=\unitlength,page=6]{mult_tau.pdf}}%
    \put(0.9583455,0.03460975){\makebox(0,0)[t]{\lineheight{1.25}\smash{\begin{tabular}[t]{c}100\end{tabular}}}}%
    \put(0.55181796,0.00035465){\makebox(0,0)[t]{\lineheight{1.25}\smash{\begin{tabular}[t]{c}Time in sec\end{tabular}}}}%
    \put(0,0){\includegraphics[width=\unitlength,page=7]{mult_tau.pdf}}%
    \put(0.08710707,0.15702443){\makebox(0,0)[rt]{\lineheight{1.25}\smash{\begin{tabular}[t]{r}−10\end{tabular}}}}%
    \put(0,0){\includegraphics[width=\unitlength,page=8]{mult_tau.pdf}}%
    \put(0.08710707,0.27923723){\makebox(0,0)[rt]{\lineheight{1.25}\smash{\begin{tabular}[t]{r}−5\end{tabular}}}}%
    \put(0,0){\includegraphics[width=\unitlength,page=9]{mult_tau.pdf}}%
    \put(0.08710707,0.40145003){\makebox(0,0)[rt]{\lineheight{1.25}\smash{\begin{tabular}[t]{r}0\end{tabular}}}}%
    \put(0,0){\includegraphics[width=\unitlength,page=10]{mult_tau.pdf}}%
    \put(0.08710707,0.52366287){\makebox(0,0)[rt]{\lineheight{1.25}\smash{\begin{tabular}[t]{r}5\end{tabular}}}}%
    \put(0,0){\includegraphics[width=\unitlength,page=11]{mult_tau.pdf}}%
    \put(0.08710707,0.64587567){\makebox(0,0)[rt]{\lineheight{1.25}\smash{\begin{tabular}[t]{r}10\end{tabular}}}}%
    \put(0.01902734,0.41065199){\rotatebox{90}{\makebox(0,0)[t]{\lineheight{1.25}\smash{\begin{tabular}[t]{c}Torque in Nm\end{tabular}}}}}%
    \put(0,0){\includegraphics[width=\unitlength,page=12]{mult_tau.pdf}}%
    \put(0.82965962,0.28789786){\makebox(0,0)[lt]{\lineheight{1.25}\smash{\begin{tabular}[t]{l}\textsl{$\tau_{U_1}$}\end{tabular}}}}%
    \put(0,0){\includegraphics[width=\unitlength,page=13]{mult_tau.pdf}}%
    \put(0.82965962,0.2338804){\makebox(0,0)[lt]{\lineheight{1.25}\smash{\begin{tabular}[t]{l}$\tau_{U_2}$\end{tabular}}}}%
    \put(0,0){\includegraphics[width=\unitlength,page=14]{mult_tau.pdf}}%
    \put(0.82965962,0.18363648){\makebox(0,0)[lt]{\lineheight{1.25}\smash{\begin{tabular}[t]{l}\textsl{$\tau_{\theta_1}$}\end{tabular}}}}%
    \put(0,0){\includegraphics[width=\unitlength,page=15]{mult_tau.pdf}}%
    \put(0.82965962,0.13150581){\makebox(0,0)[lt]{\lineheight{1.25}\smash{\begin{tabular}[t]{l}\textsl{$\tau_{\theta_2}$}\end{tabular}}}}%
  \end{picture}%
\endgroup%

%% file: pdftex/mult_label.pdf_tex
\begingroup%
  \makeatletter%
  \providecommand\color[2][]{%
    \errmessage{(Inkscape) Color is used for the text in Inkscape, but the package 'color.sty' is not loaded}%
    \renewcommand\color[2][]{}%
  }%
  \providecommand\transparent[1]{%
    \errmessage{(Inkscape) Transparency is used (non-zero) for the text in Inkscape, but the package 'transparent.sty' is not loaded}%
    \renewcommand\transparent[1]{}%
  }%
  \providecommand\rotatebox[2]{#2}%
  \newcommand*\fsize{\dimexpr\f@size pt\relax}%
  \newcommand*\lineheight[1]{\fontsize{\fsize}{#1\fsize}\selectfont}%
  \ifx\svgwidth\undefined%
    \setlength{\unitlength}{387.74264526bp}%
    \ifx\svgscale\undefined%
      \relax%
    \else%
      \setlength{\unitlength}{\unitlength * \real{\svgscale}}%
    \fi%
  \else%
    \setlength{\unitlength}{\svgwidth}%
  \fi%
  \global\let\svgwidth\undefined%
  \global\let\svgscale\undefined%
  \makeatother%
  \begin{picture}(1,0.75777178)%
    \lineheight{1}%
    \setlength\tabcolsep{0pt}%
    \put(0,0){\includegraphics[width=\unitlength,page=1]{mult_label.pdf}}%
    \put(0.11980984,0.0263319){\makebox(0,0)[t]{\lineheight{1.25}\smash{\begin{tabular}[t]{c}0\end{tabular}}}}%
    \put(0,0){\includegraphics[width=\unitlength,page=2]{mult_label.pdf}}%
    \put(0.28726863,0.0263319){\makebox(0,0)[t]{\lineheight{1.25}\smash{\begin{tabular}[t]{c}20\end{tabular}}}}%
    \put(0,0){\includegraphics[width=\unitlength,page=3]{mult_label.pdf}}%
    \put(0.45472741,0.0263319){\makebox(0,0)[t]{\lineheight{1.25}\smash{\begin{tabular}[t]{c}40\end{tabular}}}}%
    \put(0,0){\includegraphics[width=\unitlength,page=4]{mult_label.pdf}}%
    \put(0.62218618,0.0263319){\makebox(0,0)[t]{\lineheight{1.25}\smash{\begin{tabular}[t]{c}60\end{tabular}}}}%
    \put(0,0){\includegraphics[width=\unitlength,page=5]{mult_label.pdf}}%
    \put(0.78964496,0.0263319){\makebox(0,0)[t]{\lineheight{1.25}\smash{\begin{tabular}[t]{c}80\end{tabular}}}}%
    \put(0,0){\includegraphics[width=\unitlength,page=6]{mult_label.pdf}}%
    \put(0.95710373,0.0263319){\makebox(0,0)[t]{\lineheight{1.25}\smash{\begin{tabular}[t]{c}100\end{tabular}}}}%
    \put(0,0){\includegraphics[width=\unitlength,page=7]{mult_label.pdf}}%
    \put(0.05989193,0.12466168){\makebox(0,0)[rt]{\lineheight{1.25}\smash{\begin{tabular}[t]{r}−1.0\end{tabular}}}}%
    \put(0,0){\includegraphics[width=\unitlength,page=8]{mult_label.pdf}}%
    \put(0.05989193,0.20522808){\makebox(0,0)[rt]{\lineheight{1.25}\smash{\begin{tabular}[t]{r}−0.5\end{tabular}}}}%
    \put(0,0){\includegraphics[width=\unitlength,page=9]{mult_label.pdf}}%
    \put(0.05989193,0.28579448){\makebox(0,0)[rt]{\lineheight{1.25}\smash{\begin{tabular}[t]{r}0.0\end{tabular}}}}%
    \put(0,0){\includegraphics[width=\unitlength,page=10]{mult_label.pdf}}%
    \put(0.05989193,0.36636084){\makebox(0,0)[rt]{\lineheight{1.25}\smash{\begin{tabular}[t]{r}0.5\end{tabular}}}}%
    \put(0,0){\includegraphics[width=\unitlength,page=11]{mult_label.pdf}}%
    \put(0.05989193,0.44692724){\makebox(0,0)[rt]{\lineheight{1.25}\smash{\begin{tabular}[t]{r}1.0\end{tabular}}}}%
    \put(0,0){\includegraphics[width=\unitlength,page=12]{mult_label.pdf}}%
    \put(0.05989193,0.52749364){\makebox(0,0)[rt]{\lineheight{1.25}\smash{\begin{tabular}[t]{r}1.5\end{tabular}}}}%
    \put(0,0){\includegraphics[width=\unitlength,page=13]{mult_label.pdf}}%
    \put(0.05989193,0.60806006){\makebox(0,0)[rt]{\lineheight{1.25}\smash{\begin{tabular}[t]{r}2.0\end{tabular}}}}%
    \put(0,0){\includegraphics[width=\unitlength,page=14]{mult_label.pdf}}%
    \put(0.05989193,0.68862646){\makebox(0,0)[rt]{\lineheight{1.25}\smash{\begin{tabular}[t]{r}2.5\end{tabular}}}}%
    \put(0,0){\includegraphics[width=\unitlength,page=15]{mult_label.pdf}}%
    \put(0.53815797,-0.00273605){\makebox(0,0)[t]{\lineheight{1.25}\smash{\begin{tabular}[t]{c}Time in sec\end{tabular}}}}%
    \put(0,0){\includegraphics[width=\unitlength,page=16]{mult_label.pdf}}%
    \put(0.25926461,0.69583706){\makebox(0,0)[lt]{\lineheight{1.25}\smash{\begin{tabular}[t]{l}\textsl{$\theta$}\end{tabular}}}}%
    \put(0,0){\includegraphics[width=\unitlength,page=17]{mult_label.pdf}}%
    \put(0.25926461,0.64165933){\makebox(0,0)[lt]{\lineheight{1.25}\smash{\begin{tabular}[t]{l}$\theta_d$\end{tabular}}}}%
    \put(0,0){\includegraphics[width=\unitlength,page=18]{mult_label.pdf}}%
    \put(0.25937722,0.58750687){\color[rgb]{0,0,0}\makebox(0,0)[lt]{\lineheight{1.25}\smash{\begin{tabular}[t]{l}in $\varepsilon$-ball\end{tabular}}}}%
  \end{picture}%
\endgroup%

%% file: pdftex/dampex_phase.pdf_tex
\begingroup%
  \makeatletter%
  \providecommand\color[2][]{%
    \errmessage{(Inkscape) Color is used for the text in Inkscape, but the package 'color.sty' is not loaded}%
    \renewcommand\color[2][]{}%
  }%
  \providecommand\transparent[1]{%
    \errmessage{(Inkscape) Transparency is used (non-zero) for the text in Inkscape, but the package 'transparent.sty' is not loaded}%
    \renewcommand\transparent[1]{}%
  }%
  \providecommand\rotatebox[2]{#2}%
  \newcommand*\fsize{\dimexpr\f@size pt\relax}%
  \newcommand*\lineheight[1]{\fontsize{\fsize}{#1\fsize}\selectfont}%
  \ifx\svgwidth\undefined%
    \setlength{\unitlength}{404.01998901bp}%
    \ifx\svgscale\undefined%
      \relax%
    \else%
      \setlength{\unitlength}{\unitlength * \real{\svgscale}}%
    \fi%
  \else%
    \setlength{\unitlength}{\svgwidth}%
  \fi%
  \global\let\svgwidth\undefined%
  \global\let\svgscale\undefined%
  \makeatother%
  \begin{picture}(1,0.74278089)%
    \lineheight{1}%
    \setlength\tabcolsep{0pt}%
    \put(0,0){\includegraphics[width=\unitlength,page=1]{dampex_phase.pdf}}%
    \put(0.13318085,0.0408096){\makebox(0,0)[t]{\lineheight{1.25}\smash{\begin{tabular}[t]{c}−1.0\end{tabular}}}}%
    \put(0,0){\includegraphics[width=\unitlength,page=2]{dampex_phase.pdf}}%
    \put(0.24736869,0.0408096){\makebox(0,0)[t]{\lineheight{1.25}\smash{\begin{tabular}[t]{c}−0.5\end{tabular}}}}%
    \put(0,0){\includegraphics[width=\unitlength,page=3]{dampex_phase.pdf}}%
    \put(0.36155648,0.0408096){\makebox(0,0)[t]{\lineheight{1.25}\smash{\begin{tabular}[t]{c}0.0\end{tabular}}}}%
    \put(0,0){\includegraphics[width=\unitlength,page=4]{dampex_phase.pdf}}%
    \put(0.47574432,0.0408096){\makebox(0,0)[t]{\lineheight{1.25}\smash{\begin{tabular}[t]{c}0.5\end{tabular}}}}%
    \put(0,0){\includegraphics[width=\unitlength,page=5]{dampex_phase.pdf}}%
    \put(0.58993212,0.0408096){\makebox(0,0)[t]{\lineheight{1.25}\smash{\begin{tabular}[t]{c}1.0\end{tabular}}}}%
    \put(0,0){\includegraphics[width=\unitlength,page=6]{dampex_phase.pdf}}%
    \put(0.70411993,0.0408096){\makebox(0,0)[t]{\lineheight{1.25}\smash{\begin{tabular}[t]{c}1.5\end{tabular}}}}%
    \put(0,0){\includegraphics[width=\unitlength,page=7]{dampex_phase.pdf}}%
    \put(0.81830781,0.0408096){\makebox(0,0)[t]{\lineheight{1.25}\smash{\begin{tabular}[t]{c}2.0\end{tabular}}}}%
    \put(0,0){\includegraphics[width=\unitlength,page=8]{dampex_phase.pdf}}%
    \put(0.93249561,0.0408096){\makebox(0,0)[t]{\lineheight{1.25}\smash{\begin{tabular}[t]{c}2.5\end{tabular}}}}%
    \put(0.54331471,0.0070048){\makebox(0,0)[lt]{\lineheight{1.25}\smash{\begin{tabular}[t]{l}$q_1$ in rad\end{tabular}}}}%
    \put(0,0){\includegraphics[width=\unitlength,page=9]{dampex_phase.pdf}}%
    \put(0.09776745,0.12702958){\makebox(0,0)[rt]{\lineheight{1.25}\smash{\begin{tabular}[t]{r}−1.5\end{tabular}}}}%
    \put(0,0){\includegraphics[width=\unitlength,page=10]{dampex_phase.pdf}}%
    \put(0.09776745,0.21065969){\makebox(0,0)[rt]{\lineheight{1.25}\smash{\begin{tabular}[t]{r}−1.0\end{tabular}}}}%
    \put(0,0){\includegraphics[width=\unitlength,page=11]{dampex_phase.pdf}}%
    \put(0.09776745,0.29428988){\makebox(0,0)[rt]{\lineheight{1.25}\smash{\begin{tabular}[t]{r}−0.5\end{tabular}}}}%
    \put(0,0){\includegraphics[width=\unitlength,page=12]{dampex_phase.pdf}}%
    \put(0.09776745,0.37792004){\makebox(0,0)[rt]{\lineheight{1.25}\smash{\begin{tabular}[t]{r}0.0\end{tabular}}}}%
    \put(0,0){\includegraphics[width=\unitlength,page=13]{dampex_phase.pdf}}%
    \put(0.09776745,0.46155015){\makebox(0,0)[rt]{\lineheight{1.25}\smash{\begin{tabular}[t]{r}0.5\end{tabular}}}}%
    \put(0,0){\includegraphics[width=\unitlength,page=14]{dampex_phase.pdf}}%
    \put(0.09776745,0.54518029){\makebox(0,0)[rt]{\lineheight{1.25}\smash{\begin{tabular}[t]{r}1.0\end{tabular}}}}%
    \put(0,0){\includegraphics[width=\unitlength,page=15]{dampex_phase.pdf}}%
    \put(0.09776745,0.62881046){\makebox(0,0)[rt]{\lineheight{1.25}\smash{\begin{tabular}[t]{r}1.5\end{tabular}}}}%
    \put(0,0){\includegraphics[width=\unitlength,page=16]{dampex_phase.pdf}}%
    \put(0.09776745,0.7124406){\makebox(0,0)[rt]{\lineheight{1.25}\smash{\begin{tabular}[t]{r}2.0\end{tabular}}}}%
    \put(0.02256618,0.39872367){\rotatebox{90}{\makebox(0,0)[lt]{\lineheight{1.25}\smash{\begin{tabular}[t]{l}$q_2$ in rad\end{tabular}}}}}%
    \put(0,0){\includegraphics[width=\unitlength,page=17]{dampex_phase.pdf}}%
    \put(0.66944055,0.21988256){\makebox(0,0)[lt]{\lineheight{1.25}\smash{\begin{tabular}[t]{l}using $\ttau_E$\end{tabular}}}}%
    \put(0,0){\includegraphics[width=\unitlength,page=18]{dampex_phase.pdf}}%
    \put(0.66944055,0.16431096){\makebox(0,0)[lt]{\lineheight{1.25}\smash{\begin{tabular}[t]{l}using $\ttau_{\text{nD}}$\end{tabular}}}}%
  \end{picture}%
\endgroup%

%% file: pdftex/dampex_q.pdf_tex
\begingroup%
  \makeatletter%
  \providecommand\color[2][]{%
    \errmessage{(Inkscape) Color is used for the text in Inkscape, but the package 'color.sty' is not loaded}%
    \renewcommand\color[2][]{}%
  }%
  \providecommand\transparent[1]{%
    \errmessage{(Inkscape) Transparency is used (non-zero) for the text in Inkscape, but the package 'transparent.sty' is not loaded}%
    \renewcommand\transparent[1]{}%
  }%
  \providecommand\rotatebox[2]{#2}%
  \newcommand*\fsize{\dimexpr\f@size pt\relax}%
  \newcommand*\lineheight[1]{\fontsize{\fsize}{#1\fsize}\selectfont}%
  \ifx\svgwidth\undefined%
    \setlength{\unitlength}{382.5696106bp}%
    \ifx\svgscale\undefined%
      \relax%
    \else%
      \setlength{\unitlength}{\unitlength * \real{\svgscale}}%
    \fi%
  \else%
    \setlength{\unitlength}{\svgwidth}%
  \fi%
  \global\let\svgwidth\undefined%
  \global\let\svgscale\undefined%
  \makeatother%
  \begin{picture}(1,0.77745374)%
    \lineheight{1}%
    \setlength\tabcolsep{0pt}%
    \put(0,0){\includegraphics[width=\unitlength,page=1]{dampex_q.pdf}}%
    \put(0.107908,0.03612347){\makebox(0,0)[t]{\lineheight{1.25}\smash{\begin{tabular}[t]{c}0.0\end{tabular}}}}%
    \put(0,0){\includegraphics[width=\unitlength,page=2]{dampex_q.pdf}}%
    \put(0.2142508,0.03612347){\makebox(0,0)[t]{\lineheight{1.25}\smash{\begin{tabular}[t]{c}2.5\end{tabular}}}}%
    \put(0,0){\includegraphics[width=\unitlength,page=3]{dampex_q.pdf}}%
    \put(0.32059363,0.03612347){\makebox(0,0)[t]{\lineheight{1.25}\smash{\begin{tabular}[t]{c}5.0\end{tabular}}}}%
    \put(0,0){\includegraphics[width=\unitlength,page=4]{dampex_q.pdf}}%
    \put(0.42693641,0.03612347){\makebox(0,0)[t]{\lineheight{1.25}\smash{\begin{tabular}[t]{c}7.5\end{tabular}}}}%
    \put(0,0){\includegraphics[width=\unitlength,page=5]{dampex_q.pdf}}%
    \put(0.53327923,0.03612347){\makebox(0,0)[t]{\lineheight{1.25}\smash{\begin{tabular}[t]{c}10.0\end{tabular}}}}%
    \put(0,0){\includegraphics[width=\unitlength,page=6]{dampex_q.pdf}}%
    \put(0.63962206,0.03612347){\makebox(0,0)[t]{\lineheight{1.25}\smash{\begin{tabular}[t]{c}12.5\end{tabular}}}}%
    \put(0,0){\includegraphics[width=\unitlength,page=7]{dampex_q.pdf}}%
    \put(0.74596488,0.03612347){\makebox(0,0)[t]{\lineheight{1.25}\smash{\begin{tabular}[t]{c}15.0\end{tabular}}}}%
    \put(0,0){\includegraphics[width=\unitlength,page=8]{dampex_q.pdf}}%
    \put(0.85230762,0.03612347){\makebox(0,0)[t]{\lineheight{1.25}\smash{\begin{tabular}[t]{c}17.5\end{tabular}}}}%
    \put(0,0){\includegraphics[width=\unitlength,page=9]{dampex_q.pdf}}%
    \put(0.95865045,0.03612347){\makebox(0,0)[t]{\lineheight{1.25}\smash{\begin{tabular}[t]{c}20.0\end{tabular}}}}%
    \put(0.53221582,0.00037016){\makebox(0,0)[t]{\lineheight{1.25}\smash{\begin{tabular}[t]{c}Time in s\end{tabular}}}}%
    \put(0,0){\includegraphics[width=\unitlength,page=10]{dampex_q.pdf}}%
    \put(0.04717989,0.19757053){\makebox(0,0)[rt]{\lineheight{1.25}\smash{\begin{tabular}[t]{r}−1\end{tabular}}}}%
    \put(0,0){\includegraphics[width=\unitlength,page=11]{dampex_q.pdf}}%
    \put(0.04717989,0.34618504){\makebox(0,0)[rt]{\lineheight{1.25}\smash{\begin{tabular}[t]{r}0\end{tabular}}}}%
    \put(0,0){\includegraphics[width=\unitlength,page=12]{dampex_q.pdf}}%
    \put(0.04717989,0.49479954){\makebox(0,0)[rt]{\lineheight{1.25}\smash{\begin{tabular}[t]{r}1\end{tabular}}}}%
    \put(0,0){\includegraphics[width=\unitlength,page=13]{dampex_q.pdf}}%
    \put(0.04717989,0.64341408){\makebox(0,0)[rt]{\lineheight{1.25}\smash{\begin{tabular}[t]{r}2\end{tabular}}}}%
    \put(0.01985954,0.43001977){\rotatebox{90}{\makebox(0,0)[t]{\lineheight{1.25}\smash{\begin{tabular}[t]{c}Angle in rad\end{tabular}}}}}%
    \put(0,0){\includegraphics[width=\unitlength,page=14]{dampex_q.pdf}}%
    \put(0.17327248,0.72908503){\makebox(0,0)[lt]{\lineheight{1.25}\smash{\begin{tabular}[t]{l}$q_1$\end{tabular}}}}%
    \put(0,0){\includegraphics[width=\unitlength,page=15]{dampex_q.pdf}}%
    \put(0.17327248,0.68999084){\makebox(0,0)[lt]{\lineheight{1.25}\smash{\begin{tabular}[t]{l}$q_2$\end{tabular}}}}%
  \end{picture}%
\endgroup%

%% file: pdftex/dampex_naive_tau.pdf_tex
\begingroup%
  \makeatletter%
  \providecommand\color[2][]{%
    \errmessage{(Inkscape) Color is used for the text in Inkscape, but the package 'color.sty' is not loaded}%
    \renewcommand\color[2][]{}%
  }%
  \providecommand\transparent[1]{%
    \errmessage{(Inkscape) Transparency is used (non-zero) for the text in Inkscape, but the package 'transparent.sty' is not loaded}%
    \renewcommand\transparent[1]{}%
  }%
  \providecommand\rotatebox[2]{#2}%
  \newcommand*\fsize{\dimexpr\f@size pt\relax}%
  \newcommand*\lineheight[1]{\fontsize{\fsize}{#1\fsize}\selectfont}%
  \ifx\svgwidth\undefined%
    \setlength{\unitlength}{399.30203247bp}%
    \ifx\svgscale\undefined%
      \relax%
    \else%
      \setlength{\unitlength}{\unitlength * \real{\svgscale}}%
    \fi%
  \else%
    \setlength{\unitlength}{\svgwidth}%
  \fi%
  \global\let\svgwidth\undefined%
  \global\let\svgscale\undefined%
  \makeatother%
  \begin{picture}(1,0.74487519)%
    \lineheight{1}%
    \setlength\tabcolsep{0pt}%
    \put(0,0){\includegraphics[width=\unitlength,page=1]{dampex_naive_tau.pdf}}%
    \put(0.14529041,0.03460975){\makebox(0,0)[t]{\lineheight{1.25}\smash{\begin{tabular}[t]{c}0.0\end{tabular}}}}%
    \put(0,0){\includegraphics[width=\unitlength,page=2]{dampex_naive_tau.pdf}}%
    \put(0.24717701,0.03460975){\makebox(0,0)[t]{\lineheight{1.25}\smash{\begin{tabular}[t]{c}2.5\end{tabular}}}}%
    \put(0,0){\includegraphics[width=\unitlength,page=3]{dampex_naive_tau.pdf}}%
    \put(0.34906362,0.03460975){\makebox(0,0)[t]{\lineheight{1.25}\smash{\begin{tabular}[t]{c}5.0\end{tabular}}}}%
    \put(0,0){\includegraphics[width=\unitlength,page=4]{dampex_naive_tau.pdf}}%
    \put(0.4509502,0.03460975){\makebox(0,0)[t]{\lineheight{1.25}\smash{\begin{tabular}[t]{c}7.5\end{tabular}}}}%
    \put(0,0){\includegraphics[width=\unitlength,page=5]{dampex_naive_tau.pdf}}%
    \put(0.55283682,0.03460975){\makebox(0,0)[t]{\lineheight{1.25}\smash{\begin{tabular}[t]{c}10.0\end{tabular}}}}%
    \put(0,0){\includegraphics[width=\unitlength,page=6]{dampex_naive_tau.pdf}}%
    \put(0.65472343,0.03460975){\makebox(0,0)[t]{\lineheight{1.25}\smash{\begin{tabular}[t]{c}12.5\end{tabular}}}}%
    \put(0,0){\includegraphics[width=\unitlength,page=7]{dampex_naive_tau.pdf}}%
    \put(0.75661005,0.03460975){\makebox(0,0)[t]{\lineheight{1.25}\smash{\begin{tabular}[t]{c}15.0\end{tabular}}}}%
    \put(0,0){\includegraphics[width=\unitlength,page=8]{dampex_naive_tau.pdf}}%
    \put(0.85849659,0.03460975){\makebox(0,0)[t]{\lineheight{1.25}\smash{\begin{tabular}[t]{c}17.5\end{tabular}}}}%
    \put(0,0){\includegraphics[width=\unitlength,page=9]{dampex_naive_tau.pdf}}%
    \put(0.9603832,0.03460975){\makebox(0,0)[t]{\lineheight{1.25}\smash{\begin{tabular}[t]{c}20.0\end{tabular}}}}%
    \put(0.55181796,0.00035465){\makebox(0,0)[t]{\lineheight{1.25}\smash{\begin{tabular}[t]{c}Time in sec\end{tabular}}}}%
    \put(0,0){\includegraphics[width=\unitlength,page=10]{dampex_naive_tau.pdf}}%
    \put(0.08710707,0.1599331){\makebox(0,0)[rt]{\lineheight{1.25}\smash{\begin{tabular}[t]{r}−20\end{tabular}}}}%
    \put(0,0){\includegraphics[width=\unitlength,page=11]{dampex_naive_tau.pdf}}%
    \put(0.08710707,0.26372352){\makebox(0,0)[rt]{\lineheight{1.25}\smash{\begin{tabular}[t]{r}−10\end{tabular}}}}%
    \put(0,0){\includegraphics[width=\unitlength,page=12]{dampex_naive_tau.pdf}}%
    \put(0.08710707,0.36751394){\makebox(0,0)[rt]{\lineheight{1.25}\smash{\begin{tabular}[t]{r}0\end{tabular}}}}%
    \put(0,0){\includegraphics[width=\unitlength,page=13]{dampex_naive_tau.pdf}}%
    \put(0.08710707,0.47130436){\makebox(0,0)[rt]{\lineheight{1.25}\smash{\begin{tabular}[t]{r}10\end{tabular}}}}%
    \put(0,0){\includegraphics[width=\unitlength,page=14]{dampex_naive_tau.pdf}}%
    \put(0.08710707,0.57509482){\makebox(0,0)[rt]{\lineheight{1.25}\smash{\begin{tabular}[t]{r}20\end{tabular}}}}%
    \put(0,0){\includegraphics[width=\unitlength,page=15]{dampex_naive_tau.pdf}}%
    \put(0.08710707,0.67888522){\makebox(0,0)[rt]{\lineheight{1.25}\smash{\begin{tabular}[t]{r}30\end{tabular}}}}%
    \put(0.01902734,0.41065199){\rotatebox{90}{\makebox(0,0)[t]{\lineheight{1.25}\smash{\begin{tabular}[t]{c}Torque in Nm\end{tabular}}}}}%
    \put(0,0){\includegraphics[width=\unitlength,page=16]{dampex_naive_tau.pdf}}%
    \put(0.64048672,0.68186892){\makebox(0,0)[lt]{\lineheight{1.25}\smash{\begin{tabular}[t]{l}$\tau_{U,1}$\end{tabular}}}}%
    \put(0,0){\includegraphics[width=\unitlength,page=17]{dampex_naive_tau.pdf}}%
    \put(0.81387279,0.68419557){\makebox(0,0)[lt]{\lineheight{1.25}\smash{\begin{tabular}[t]{l}$\tau_{U,2}$\end{tabular}}}}%
    \put(0,0){\includegraphics[width=\unitlength,page=18]{dampex_naive_tau.pdf}}%
    \put(0.64048672,0.62472732){\makebox(0,0)[lt]{\lineheight{1.25}\smash{\begin{tabular}[t]{l}$\tau_{\theta,1}$\end{tabular}}}}%
    \put(0,0){\includegraphics[width=\unitlength,page=19]{dampex_naive_tau.pdf}}%
    \put(0.81419279,0.6273317){\makebox(0,0)[lt]{\lineheight{1.25}\smash{\begin{tabular}[t]{l}$\tau_{\theta,2}$\end{tabular}}}}%
    \put(0,0){\includegraphics[width=\unitlength,page=20]{dampex_naive_tau.pdf}}%
    \put(0.64154453,0.56794125){\makebox(0,0)[lt]{\lineheight{1.25}\smash{\begin{tabular}[t]{l}$\tau_{\text{nD},1}$\end{tabular}}}}%
    \put(0,0){\includegraphics[width=\unitlength,page=21]{dampex_naive_tau.pdf}}%
    \put(0.81439058,0.57015871){\makebox(0,0)[lt]{\lineheight{1.25}\smash{\begin{tabular}[t]{l}$\tau_{\text{nD},2}$\end{tabular}}}}%
  \end{picture}%
\endgroup%

%% file: pdftex/dampex_my_tau.pdf_tex
\begingroup%
  \makeatletter%
  \providecommand\color[2][]{%
    \errmessage{(Inkscape) Color is used for the text in Inkscape, but the package 'color.sty' is not loaded}%
    \renewcommand\color[2][]{}%
  }%
  \providecommand\transparent[1]{%
    \errmessage{(Inkscape) Transparency is used (non-zero) for the text in Inkscape, but the package 'transparent.sty' is not loaded}%
    \renewcommand\transparent[1]{}%
  }%
  \providecommand\rotatebox[2]{#2}%
  \newcommand*\fsize{\dimexpr\f@size pt\relax}%
  \newcommand*\lineheight[1]{\fontsize{\fsize}{#1\fsize}\selectfont}%
  \ifx\svgwidth\undefined%
    \setlength{\unitlength}{387.75540161bp}%
    \ifx\svgscale\undefined%
      \relax%
    \else%
      \setlength{\unitlength}{\unitlength * \real{\svgscale}}%
    \fi%
  \else%
    \setlength{\unitlength}{\svgwidth}%
  \fi%
  \global\let\svgwidth\undefined%
  \global\let\svgscale\undefined%
  \makeatother%
  \begin{picture}(1,0.76705618)%
    \lineheight{1}%
    \setlength\tabcolsep{0pt}%
    \put(0,0){\includegraphics[width=\unitlength,page=1]{dampex_my_tau.pdf}}%
    \put(0.11370805,0.03564036){\makebox(0,0)[t]{\lineheight{1.25}\smash{\begin{tabular}[t]{c}0.0\end{tabular}}}}%
    \put(0,0){\includegraphics[width=\unitlength,page=2]{dampex_my_tau.pdf}}%
    \put(0.21862864,0.03564036){\makebox(0,0)[t]{\lineheight{1.25}\smash{\begin{tabular}[t]{c}2.5\end{tabular}}}}%
    \put(0,0){\includegraphics[width=\unitlength,page=3]{dampex_my_tau.pdf}}%
    \put(0.32354924,0.03564036){\makebox(0,0)[t]{\lineheight{1.25}\smash{\begin{tabular}[t]{c}5.0\end{tabular}}}}%
    \put(0,0){\includegraphics[width=\unitlength,page=4]{dampex_my_tau.pdf}}%
    \put(0.42846981,0.03564036){\makebox(0,0)[t]{\lineheight{1.25}\smash{\begin{tabular}[t]{c}7.5\end{tabular}}}}%
    \put(0,0){\includegraphics[width=\unitlength,page=5]{dampex_my_tau.pdf}}%
    \put(0.53339042,0.03564036){\makebox(0,0)[t]{\lineheight{1.25}\smash{\begin{tabular}[t]{c}10.0\end{tabular}}}}%
    \put(0,0){\includegraphics[width=\unitlength,page=6]{dampex_my_tau.pdf}}%
    \put(0.63831103,0.03564036){\makebox(0,0)[t]{\lineheight{1.25}\smash{\begin{tabular}[t]{c}12.5\end{tabular}}}}%
    \put(0,0){\includegraphics[width=\unitlength,page=7]{dampex_my_tau.pdf}}%
    \put(0.74323164,0.03564036){\makebox(0,0)[t]{\lineheight{1.25}\smash{\begin{tabular}[t]{c}15.0\end{tabular}}}}%
    \put(0,0){\includegraphics[width=\unitlength,page=8]{dampex_my_tau.pdf}}%
    \put(0.84815217,0.03564036){\makebox(0,0)[t]{\lineheight{1.25}\smash{\begin{tabular}[t]{c}17.5\end{tabular}}}}%
    \put(0,0){\includegraphics[width=\unitlength,page=9]{dampex_my_tau.pdf}}%
    \put(0.95307278,0.03564036){\makebox(0,0)[t]{\lineheight{1.25}\smash{\begin{tabular}[t]{c}20.0\end{tabular}}}}%
    \put(0.53234123,0.00036521){\makebox(0,0)[t]{\lineheight{1.25}\smash{\begin{tabular}[t]{c}Time in sec\end{tabular}}}}%
    \put(0,0){\includegraphics[width=\unitlength,page=10]{dampex_my_tau.pdf}}%
    \put(0.05379211,0.15807093){\makebox(0,0)[rt]{\lineheight{1.25}\smash{\begin{tabular}[t]{r}−10\end{tabular}}}}%
    \put(0,0){\includegraphics[width=\unitlength,page=11]{dampex_my_tau.pdf}}%
    \put(0.05379211,0.28551603){\makebox(0,0)[rt]{\lineheight{1.25}\smash{\begin{tabular}[t]{r}−5\end{tabular}}}}%
    \put(0,0){\includegraphics[width=\unitlength,page=12]{dampex_my_tau.pdf}}%
    \put(0.05379211,0.41296113){\makebox(0,0)[rt]{\lineheight{1.25}\smash{\begin{tabular}[t]{r}0\end{tabular}}}}%
    \put(0,0){\includegraphics[width=\unitlength,page=13]{dampex_my_tau.pdf}}%
    \put(0.05379211,0.54040619){\makebox(0,0)[rt]{\lineheight{1.25}\smash{\begin{tabular}[t]{r}5\end{tabular}}}}%
    \put(0,0){\includegraphics[width=\unitlength,page=14]{dampex_my_tau.pdf}}%
    \put(0.05379211,0.66785131){\makebox(0,0)[rt]{\lineheight{1.25}\smash{\begin{tabular}[t]{r}10\end{tabular}}}}%
    \put(0.01959394,0.42288044){\rotatebox{90}{\makebox(0,0)[t]{\lineheight{1.25}\smash{\begin{tabular}[t]{c}Torque in Nm\end{tabular}}}}}%
    \put(0,0){\includegraphics[width=\unitlength,page=15]{dampex_my_tau.pdf}}%
    \put(0.69133447,0.23777754){\makebox(0,0)[lt]{\lineheight{1.25}\smash{\begin{tabular}[t]{l}$\tau_{U,1}$\end{tabular}}}}%
    \put(0,0){\includegraphics[width=\unitlength,page=16]{dampex_my_tau.pdf}}%
    \put(0.86988366,0.24017347){\makebox(0,0)[lt]{\lineheight{1.25}\smash{\begin{tabular}[t]{l}$\tau_{U,2}$\end{tabular}}}}%
    \put(0,0){\includegraphics[width=\unitlength,page=17]{dampex_my_tau.pdf}}%
    \put(0.69133447,0.17893437){\makebox(0,0)[lt]{\lineheight{1.25}\smash{\begin{tabular}[t]{l}$\tau_{\theta,1}$\end{tabular}}}}%
    \put(0,0){\includegraphics[width=\unitlength,page=18]{dampex_my_tau.pdf}}%
    \put(0.87021319,0.18161631){\makebox(0,0)[lt]{\lineheight{1.25}\smash{\begin{tabular}[t]{l}$\tau_{\theta,2}$\end{tabular}}}}%
    \put(0,0){\includegraphics[width=\unitlength,page=19]{dampex_my_tau.pdf}}%
    \put(0.69242378,0.12045732){\makebox(0,0)[lt]{\lineheight{1.25}\smash{\begin{tabular}[t]{l}$\tau_{E,1}$\end{tabular}}}}%
    \put(0,0){\includegraphics[width=\unitlength,page=20]{dampex_my_tau.pdf}}%
    \put(0.87041686,0.1227408){\makebox(0,0)[lt]{\lineheight{1.25}\smash{\begin{tabular}[t]{l}$\tau_{E,2}$\end{tabular}}}}%
  \end{picture}%
\endgroup%

%% file: pdftex/sensitivity_phase.pdf_tex
\begingroup%
  \makeatletter%
  \providecommand\color[2][]{%
    \errmessage{(Inkscape) Color is used for the text in Inkscape, but the package 'color.sty' is not loaded}%
    \renewcommand\color[2][]{}%
  }%
  \providecommand\transparent[1]{%
    \errmessage{(Inkscape) Transparency is used (non-zero) for the text in Inkscape, but the package 'transparent.sty' is not loaded}%
    \renewcommand\transparent[1]{}%
  }%
  \providecommand\rotatebox[2]{#2}%
  \newcommand*\fsize{\dimexpr\f@size pt\relax}%
  \newcommand*\lineheight[1]{\fontsize{\fsize}{#1\fsize}\selectfont}%
  \ifx\svgwidth\undefined%
    \setlength{\unitlength}{400.90963745bp}%
    \ifx\svgscale\undefined%
      \relax%
    \else%
      \setlength{\unitlength}{\unitlength * \real{\svgscale}}%
    \fi%
  \else%
    \setlength{\unitlength}{\svgwidth}%
  \fi%
  \global\let\svgwidth\undefined%
  \global\let\svgscale\undefined%
  \makeatother%
  \begin{picture}(1,0.74561078)%
    \lineheight{1}%
    \setlength\tabcolsep{0pt}%
    \put(0,0){\includegraphics[width=\unitlength,page=1]{sensitivity_phase.pdf}}%
    \put(0.12646338,0.03819343){\makebox(0,0)[t]{\lineheight{1.25}\smash{\begin{tabular}[t]{c}−1.0\end{tabular}}}}%
    \put(0,0){\includegraphics[width=\unitlength,page=2]{sensitivity_phase.pdf}}%
    \put(0.24142969,0.03819343){\makebox(0,0)[t]{\lineheight{1.25}\smash{\begin{tabular}[t]{c}−0.5\end{tabular}}}}%
    \put(0,0){\includegraphics[width=\unitlength,page=3]{sensitivity_phase.pdf}}%
    \put(0.35639601,0.03819343){\makebox(0,0)[t]{\lineheight{1.25}\smash{\begin{tabular}[t]{c}0.0\end{tabular}}}}%
    \put(0,0){\includegraphics[width=\unitlength,page=4]{sensitivity_phase.pdf}}%
    \put(0.4713623,0.03819343){\makebox(0,0)[t]{\lineheight{1.25}\smash{\begin{tabular}[t]{c}0.5\end{tabular}}}}%
    \put(0,0){\includegraphics[width=\unitlength,page=5]{sensitivity_phase.pdf}}%
    \put(0.58632863,0.03819343){\makebox(0,0)[t]{\lineheight{1.25}\smash{\begin{tabular}[t]{c}1.0\end{tabular}}}}%
    \put(0,0){\includegraphics[width=\unitlength,page=6]{sensitivity_phase.pdf}}%
    \put(0.70129492,0.03819343){\makebox(0,0)[t]{\lineheight{1.25}\smash{\begin{tabular}[t]{c}1.5\end{tabular}}}}%
    \put(0,0){\includegraphics[width=\unitlength,page=7]{sensitivity_phase.pdf}}%
    \put(0.81626122,0.03819343){\makebox(0,0)[t]{\lineheight{1.25}\smash{\begin{tabular}[t]{c}2.0\end{tabular}}}}%
    \put(0,0){\includegraphics[width=\unitlength,page=8]{sensitivity_phase.pdf}}%
    \put(0.93122751,0.03819343){\makebox(0,0)[t]{\lineheight{1.25}\smash{\begin{tabular}[t]{c}2.5\end{tabular}}}}%
    \put(0.53977164,0.00412636){\makebox(0,0)[lt]{\lineheight{1.25}\smash{\begin{tabular}[t]{l}$q_1$ in rad\end{tabular}}}}%
    \put(0,0){\includegraphics[width=\unitlength,page=9]{sensitivity_phase.pdf}}%
    \put(0.09076771,0.09080701){\makebox(0,0)[rt]{\lineheight{1.25}\smash{\begin{tabular}[t]{r}−2.0\end{tabular}}}}%
    \put(0,0){\includegraphics[width=\unitlength,page=10]{sensitivity_phase.pdf}}%
    \put(0.09076771,0.16836311){\makebox(0,0)[rt]{\lineheight{1.25}\smash{\begin{tabular}[t]{r}−1.5\end{tabular}}}}%
    \put(0,0){\includegraphics[width=\unitlength,page=11]{sensitivity_phase.pdf}}%
    \put(0.09076771,0.24591913){\makebox(0,0)[rt]{\lineheight{1.25}\smash{\begin{tabular}[t]{r}−1.0\end{tabular}}}}%
    \put(0,0){\includegraphics[width=\unitlength,page=12]{sensitivity_phase.pdf}}%
    \put(0.09076771,0.32347519){\makebox(0,0)[rt]{\lineheight{1.25}\smash{\begin{tabular}[t]{r}−0.5\end{tabular}}}}%
    \put(0,0){\includegraphics[width=\unitlength,page=13]{sensitivity_phase.pdf}}%
    \put(0.09076771,0.40103125){\makebox(0,0)[rt]{\lineheight{1.25}\smash{\begin{tabular}[t]{r}0.0\end{tabular}}}}%
    \put(0,0){\includegraphics[width=\unitlength,page=14]{sensitivity_phase.pdf}}%
    \put(0.09076771,0.47858731){\makebox(0,0)[rt]{\lineheight{1.25}\smash{\begin{tabular}[t]{r}0.5\end{tabular}}}}%
    \put(0,0){\includegraphics[width=\unitlength,page=15]{sensitivity_phase.pdf}}%
    \put(0.09076771,0.55614335){\makebox(0,0)[rt]{\lineheight{1.25}\smash{\begin{tabular}[t]{r}1.0\end{tabular}}}}%
    \put(0,0){\includegraphics[width=\unitlength,page=16]{sensitivity_phase.pdf}}%
    \put(0.09076771,0.63369941){\makebox(0,0)[rt]{\lineheight{1.25}\smash{\begin{tabular}[t]{r}1.5\end{tabular}}}}%
    \put(0,0){\includegraphics[width=\unitlength,page=17]{sensitivity_phase.pdf}}%
    \put(0.09076771,0.71125546){\makebox(0,0)[rt]{\lineheight{1.25}\smash{\begin{tabular}[t]{r}2.0\end{tabular}}}}%
    \put(0.036182,0.39755834){\rotatebox{90}{\makebox(0,0)[lt]{\lineheight{1.25}\smash{\begin{tabular}[t]{l}$q_2$ in rad\end{tabular}}}}}%
    \put(0,0){\includegraphics[width=\unitlength,page=18]{sensitivity_phase.pdf}}%
    \put(0.50586566,0.40436796){\makebox(0,0)[lt]{\lineheight{1.25}\smash{\begin{tabular}[t]{l}$\epsilon=0$\end{tabular}}}}%
    \put(0,0){\includegraphics[width=\unitlength,page=19]{sensitivity_phase.pdf}}%
    \put(0.50586566,0.35278048){\makebox(0,0)[lt]{\lineheight{1.25}\smash{\begin{tabular}[t]{l}$\epsilon=0.01$\end{tabular}}}}%
    \put(0,0){\includegraphics[width=\unitlength,page=20]{sensitivity_phase.pdf}}%
    \put(0.50586566,0.301193){\makebox(0,0)[lt]{\lineheight{1.25}\smash{\begin{tabular}[t]{l}$\epsilon=0.05$\end{tabular}}}}%
    \put(0,0){\includegraphics[width=\unitlength,page=21]{sensitivity_phase.pdf}}%
    \put(0.50586566,0.24960551){\makebox(0,0)[lt]{\lineheight{1.25}\smash{\begin{tabular}[t]{l}$\epsilon=0.1$\end{tabular}}}}%
    \put(0,0){\includegraphics[width=\unitlength,page=22]{sensitivity_phase.pdf}}%
    \put(0.50586566,0.19801808){\makebox(0,0)[lt]{\lineheight{1.25}\smash{\begin{tabular}[t]{l}$\epsilon=0.5$\end{tabular}}}}%
    \put(0,0){\includegraphics[width=\unitlength,page=23]{sensitivity_phase.pdf}}%
    \put(0.50586566,0.14643054){\makebox(0,0)[lt]{\lineheight{1.25}\smash{\begin{tabular}[t]{l}$\epsilon=1.0$\end{tabular}}}}%
  \end{picture}%
\endgroup%

%% file: pdftex/sensitivity_q.pdf_tex
\begingroup%
  \makeatletter%
  \providecommand\color[2][]{%
    \errmessage{(Inkscape) Color is used for the text in Inkscape, but the package 'color.sty' is not loaded}%
    \renewcommand\color[2][]{}%
  }%
  \providecommand\transparent[1]{%
    \errmessage{(Inkscape) Transparency is used (non-zero) for the text in Inkscape, but the package 'transparent.sty' is not loaded}%
    \renewcommand\transparent[1]{}%
  }%
  \providecommand\rotatebox[2]{#2}%
  \newcommand*\fsize{\dimexpr\f@size pt\relax}%
  \newcommand*\lineheight[1]{\fontsize{\fsize}{#1\fsize}\selectfont}%
  \ifx\svgwidth\undefined%
    \setlength{\unitlength}{439.17953491bp}%
    \ifx\svgscale\undefined%
      \relax%
    \else%
      \setlength{\unitlength}{\unitlength * \real{\svgscale}}%
    \fi%
  \else%
    \setlength{\unitlength}{\svgwidth}%
  \fi%
  \global\let\svgwidth\undefined%
  \global\let\svgscale\undefined%
  \makeatother%
  \begin{picture}(1,0.73240702)%
    \lineheight{1}%
    \setlength\tabcolsep{0pt}%
    \put(0,0){\includegraphics[width=\unitlength,page=1]{sensitivity_q.pdf}}%
    \put(0.06471053,0.41517151){\makebox(0,0)[rt]{\lineheight{1.25}\smash{\begin{tabular}[t]{r}−1\end{tabular}}}}%
    \put(0,0){\includegraphics[width=\unitlength,page=2]{sensitivity_q.pdf}}%
    \put(0.06471053,0.49625018){\makebox(0,0)[rt]{\lineheight{1.25}\smash{\begin{tabular}[t]{r}0\end{tabular}}}}%
    \put(0,0){\includegraphics[width=\unitlength,page=3]{sensitivity_q.pdf}}%
    \put(0.06471053,0.57732883){\makebox(0,0)[rt]{\lineheight{1.25}\smash{\begin{tabular}[t]{r}1\end{tabular}}}}%
    \put(0,0){\includegraphics[width=\unitlength,page=4]{sensitivity_q.pdf}}%
    \put(0.06471053,0.65840749){\makebox(0,0)[rt]{\lineheight{1.25}\smash{\begin{tabular}[t]{r}2\end{tabular}}}}%
    \put(0.01729969,0.57444416){\rotatebox{90}{\makebox(0,0)[t]{\lineheight{1.25}\smash{\begin{tabular}[t]{c}$q_1$ in rad\end{tabular}}}}}%
    \put(0,0){\includegraphics[width=\unitlength,page=5]{sensitivity_q.pdf}}%
    \put(0.12239662,0.03146717){\makebox(0,0)[t]{\lineheight{1.25}\smash{\begin{tabular}[t]{c}0.0\end{tabular}}}}%
    \put(0,0){\includegraphics[width=\unitlength,page=6]{sensitivity_q.pdf}}%
    \put(0.22702636,0.03146717){\makebox(0,0)[t]{\lineheight{1.25}\smash{\begin{tabular}[t]{c}2.5\end{tabular}}}}%
    \put(0,0){\includegraphics[width=\unitlength,page=7]{sensitivity_q.pdf}}%
    \put(0.3316561,0.03146717){\makebox(0,0)[t]{\lineheight{1.25}\smash{\begin{tabular}[t]{c}5.0\end{tabular}}}}%
    \put(0,0){\includegraphics[width=\unitlength,page=8]{sensitivity_q.pdf}}%
    \put(0.43628586,0.03146717){\makebox(0,0)[t]{\lineheight{1.25}\smash{\begin{tabular}[t]{c}7.5\end{tabular}}}}%
    \put(0,0){\includegraphics[width=\unitlength,page=9]{sensitivity_q.pdf}}%
    \put(0.54091558,0.03146717){\makebox(0,0)[t]{\lineheight{1.25}\smash{\begin{tabular}[t]{c}10.0\end{tabular}}}}%
    \put(0,0){\includegraphics[width=\unitlength,page=10]{sensitivity_q.pdf}}%
    \put(0.6455453,0.03146717){\makebox(0,0)[t]{\lineheight{1.25}\smash{\begin{tabular}[t]{c}12.5\end{tabular}}}}%
    \put(0,0){\includegraphics[width=\unitlength,page=11]{sensitivity_q.pdf}}%
    \put(0.75017509,0.03146717){\makebox(0,0)[t]{\lineheight{1.25}\smash{\begin{tabular}[t]{c}15.0\end{tabular}}}}%
    \put(0,0){\includegraphics[width=\unitlength,page=12]{sensitivity_q.pdf}}%
    \put(0.85480482,0.03146717){\makebox(0,0)[t]{\lineheight{1.25}\smash{\begin{tabular}[t]{c}17.5\end{tabular}}}}%
    \put(0,0){\includegraphics[width=\unitlength,page=13]{sensitivity_q.pdf}}%
    \put(0.95943454,0.03146717){\makebox(0,0)[t]{\lineheight{1.25}\smash{\begin{tabular}[t]{c}20.0\end{tabular}}}}%
    \put(0.5398693,0.00032244){\makebox(0,0)[t]{\lineheight{1.25}\smash{\begin{tabular}[t]{c}Time in s\end{tabular}}}}%
    \put(0,0){\includegraphics[width=\unitlength,page=14]{sensitivity_q.pdf}}%
    \put(0.06471053,0.07094884){\makebox(0,0)[rt]{\lineheight{1.25}\smash{\begin{tabular}[t]{r}−2\end{tabular}}}}%
    \put(0,0){\includegraphics[width=\unitlength,page=15]{sensitivity_q.pdf}}%
    \put(0.06471053,0.14434974){\makebox(0,0)[rt]{\lineheight{1.25}\smash{\begin{tabular}[t]{r}−1\end{tabular}}}}%
    \put(0,0){\includegraphics[width=\unitlength,page=16]{sensitivity_q.pdf}}%
    \put(0.06471053,0.21775065){\makebox(0,0)[rt]{\lineheight{1.25}\smash{\begin{tabular}[t]{r}0\end{tabular}}}}%
    \put(0,0){\includegraphics[width=\unitlength,page=17]{sensitivity_q.pdf}}%
    \put(0.06471053,0.29115155){\makebox(0,0)[rt]{\lineheight{1.25}\smash{\begin{tabular}[t]{r}1\end{tabular}}}}%
    \put(0,0){\includegraphics[width=\unitlength,page=18]{sensitivity_q.pdf}}%
    \put(0.06471053,0.36455246){\makebox(0,0)[rt]{\lineheight{1.25}\smash{\begin{tabular}[t]{r}2\end{tabular}}}}%
    \put(0.01729966,0.22745191){\rotatebox{90}{\makebox(0,0)[t]{\lineheight{1.25}\smash{\begin{tabular}[t]{c}$q_2$ in rad\end{tabular}}}}}%
    \put(0,0){\includegraphics[width=\unitlength,page=19]{sensitivity_q.pdf}}%
  \end{picture}%
\endgroup%

%% file: pdftex/sensitivity_0.pdf_tex
\begingroup%
  \makeatletter%
  \providecommand\color[2][]{%
    \errmessage{(Inkscape) Color is used for the text in Inkscape, but the package 'color.sty' is not loaded}%
    \renewcommand\color[2][]{}%
  }%
  \providecommand\transparent[1]{%
    \errmessage{(Inkscape) Transparency is used (non-zero) for the text in Inkscape, but the package 'transparent.sty' is not loaded}%
    \renewcommand\transparent[1]{}%
  }%
  \providecommand\rotatebox[2]{#2}%
  \newcommand*\fsize{\dimexpr\f@size pt\relax}%
  \newcommand*\lineheight[1]{\fontsize{\fsize}{#1\fsize}\selectfont}%
  \ifx\svgwidth\undefined%
    \setlength{\unitlength}{386.48535156bp}%
    \ifx\svgscale\undefined%
      \relax%
    \else%
      \setlength{\unitlength}{\unitlength * \real{\svgscale}}%
    \fi%
  \else%
    \setlength{\unitlength}{\svgwidth}%
  \fi%
  \global\let\svgwidth\undefined%
  \global\let\svgscale\undefined%
  \makeatother%
  \begin{picture}(1,0.76957684)%
    \lineheight{1}%
    \setlength\tabcolsep{0pt}%
    \put(0,0){\includegraphics[width=\unitlength,page=1]{sensitivity_0.pdf}}%
    \put(0.11694644,0.03575746){\makebox(0,0)[t]{\lineheight{1.25}\smash{\begin{tabular}[t]{c}0.0\end{tabular}}}}%
    \put(0,0){\includegraphics[width=\unitlength,page=2]{sensitivity_0.pdf}}%
    \put(0.2222118,0.03575746){\makebox(0,0)[t]{\lineheight{1.25}\smash{\begin{tabular}[t]{c}2.5\end{tabular}}}}%
    \put(0,0){\includegraphics[width=\unitlength,page=3]{sensitivity_0.pdf}}%
    \put(0.3274772,0.03575746){\makebox(0,0)[t]{\lineheight{1.25}\smash{\begin{tabular}[t]{c}5.0\end{tabular}}}}%
    \put(0,0){\includegraphics[width=\unitlength,page=4]{sensitivity_0.pdf}}%
    \put(0.43274255,0.03575746){\makebox(0,0)[t]{\lineheight{1.25}\smash{\begin{tabular}[t]{c}7.5\end{tabular}}}}%
    \put(0,0){\includegraphics[width=\unitlength,page=5]{sensitivity_0.pdf}}%
    \put(0.53800795,0.03575746){\makebox(0,0)[t]{\lineheight{1.25}\smash{\begin{tabular}[t]{c}10.0\end{tabular}}}}%
    \put(0,0){\includegraphics[width=\unitlength,page=6]{sensitivity_0.pdf}}%
    \put(0.64327334,0.03575746){\makebox(0,0)[t]{\lineheight{1.25}\smash{\begin{tabular}[t]{c}12.5\end{tabular}}}}%
    \put(0,0){\includegraphics[width=\unitlength,page=7]{sensitivity_0.pdf}}%
    \put(0.74853873,0.03575746){\makebox(0,0)[t]{\lineheight{1.25}\smash{\begin{tabular}[t]{c}15.0\end{tabular}}}}%
    \put(0,0){\includegraphics[width=\unitlength,page=8]{sensitivity_0.pdf}}%
    \put(0.85380405,0.03575746){\makebox(0,0)[t]{\lineheight{1.25}\smash{\begin{tabular}[t]{c}17.5\end{tabular}}}}%
    \put(0,0){\includegraphics[width=\unitlength,page=9]{sensitivity_0.pdf}}%
    \put(0.95906944,0.03575746){\makebox(0,0)[t]{\lineheight{1.25}\smash{\begin{tabular}[t]{c}20.0\end{tabular}}}}%
    \put(0.5319926,-0.00936621){\makebox(0,0)[t]{\lineheight{1.25}\smash{\begin{tabular}[t]{c}Time in sec\end{tabular}}}}%
    \put(0,0){\includegraphics[width=\unitlength,page=10]{sensitivity_0.pdf}}%
    \put(0.05683361,0.1556564){\makebox(0,0)[rt]{\lineheight{1.25}\smash{\begin{tabular}[t]{r}−60\end{tabular}}}}%
    \put(0,0){\includegraphics[width=\unitlength,page=11]{sensitivity_0.pdf}}%
    \put(0.05683361,0.24230798){\makebox(0,0)[rt]{\lineheight{1.25}\smash{\begin{tabular}[t]{r}−40\end{tabular}}}}%
    \put(0,0){\includegraphics[width=\unitlength,page=12]{sensitivity_0.pdf}}%
    \put(0.05683361,0.32895955){\makebox(0,0)[rt]{\lineheight{1.25}\smash{\begin{tabular}[t]{r}−20\end{tabular}}}}%
    \put(0,0){\includegraphics[width=\unitlength,page=13]{sensitivity_0.pdf}}%
    \put(0.05683361,0.41561116){\makebox(0,0)[rt]{\lineheight{1.25}\smash{\begin{tabular}[t]{r}0\end{tabular}}}}%
    \put(0,0){\includegraphics[width=\unitlength,page=14]{sensitivity_0.pdf}}%
    \put(0.05683361,0.50226269){\makebox(0,0)[rt]{\lineheight{1.25}\smash{\begin{tabular}[t]{r}20\end{tabular}}}}%
    \put(0,0){\includegraphics[width=\unitlength,page=15]{sensitivity_0.pdf}}%
    \put(0.05683361,0.58891432){\makebox(0,0)[rt]{\lineheight{1.25}\smash{\begin{tabular}[t]{r}40\end{tabular}}}}%
    \put(0,0){\includegraphics[width=\unitlength,page=16]{sensitivity_0.pdf}}%
    \put(0.05683361,0.67556589){\makebox(0,0)[rt]{\lineheight{1.25}\smash{\begin{tabular}[t]{r}60\end{tabular}}}}%
    \put(-0.00525444,0.42009153){\rotatebox{90}{\makebox(0,0)[t]{\lineheight{1.25}\smash{\begin{tabular}[t]{c}Power\end{tabular}}}}}%
    \put(0,0){\includegraphics[width=\unitlength,page=17]{sensitivity_0.pdf}}%
    \put(0.70351608,0.69738814){\color[rgb]{0,0,0}\makebox(0,0)[lt]{\lineheight{1.25}\smash{\begin{tabular}[t]{l}$P_{\tilde{U}}$\end{tabular}}}}%
    \put(0,0){\includegraphics[width=\unitlength,page=18]{sensitivity_0.pdf}}%
    \put(0.70351608,0.62231612){\color[rgb]{0,0,0}\makebox(0,0)[lt]{\lineheight{1.25}\smash{\begin{tabular}[t]{l}$P_\theta$\end{tabular}}}}%
    \put(0,0){\includegraphics[width=\unitlength,page=19]{sensitivity_0.pdf}}%
    \put(0.70351608,0.547244){\color[rgb]{0,0,0}\makebox(0,0)[lt]{\lineheight{1.25}\smash{\begin{tabular}[t]{l}$P_E$\end{tabular}}}}%
  \end{picture}%
\endgroup%

%% file: pdftex/sensitivity_0_1.pdf_tex
\begingroup%
  \makeatletter%
  \providecommand\color[2][]{%
    \errmessage{(Inkscape) Color is used for the text in Inkscape, but the package 'color.sty' is not loaded}%
    \renewcommand\color[2][]{}%
  }%
  \providecommand\transparent[1]{%
    \errmessage{(Inkscape) Transparency is used (non-zero) for the text in Inkscape, but the package 'transparent.sty' is not loaded}%
    \renewcommand\transparent[1]{}%
  }%
  \providecommand\rotatebox[2]{#2}%
  \newcommand*\fsize{\dimexpr\f@size pt\relax}%
  \newcommand*\lineheight[1]{\fontsize{\fsize}{#1\fsize}\selectfont}%
  \ifx\svgwidth\undefined%
    \setlength{\unitlength}{361.02001953bp}%
    \ifx\svgscale\undefined%
      \relax%
    \else%
      \setlength{\unitlength}{\unitlength * \real{\svgscale}}%
    \fi%
  \else%
    \setlength{\unitlength}{\svgwidth}%
  \fi%
  \global\let\svgwidth\undefined%
  \global\let\svgscale\undefined%
  \makeatother%
  \begin{picture}(1,0.82386062)%
    \lineheight{1}%
    \setlength\tabcolsep{0pt}%
    \put(0,0){\includegraphics[width=\unitlength,page=1]{sensitivity_0_1.pdf}}%
    \put(0.05465826,0.03827971){\makebox(0,0)[t]{\lineheight{1.25}\smash{\begin{tabular}[t]{c}0.0\end{tabular}}}}%
    \put(0,0){\includegraphics[width=\unitlength,page=2]{sensitivity_0_1.pdf}}%
    \put(0.16734876,0.03827971){\makebox(0,0)[t]{\lineheight{1.25}\smash{\begin{tabular}[t]{c}2.5\end{tabular}}}}%
    \put(0,0){\includegraphics[width=\unitlength,page=3]{sensitivity_0_1.pdf}}%
    \put(0.28003928,0.03827971){\makebox(0,0)[t]{\lineheight{1.25}\smash{\begin{tabular}[t]{c}5.0\end{tabular}}}}%
    \put(0,0){\includegraphics[width=\unitlength,page=4]{sensitivity_0_1.pdf}}%
    \put(0.39272975,0.03827971){\makebox(0,0)[t]{\lineheight{1.25}\smash{\begin{tabular}[t]{c}7.5\end{tabular}}}}%
    \put(0,0){\includegraphics[width=\unitlength,page=5]{sensitivity_0_1.pdf}}%
    \put(0.50542027,0.03827971){\makebox(0,0)[t]{\lineheight{1.25}\smash{\begin{tabular}[t]{c}10.0\end{tabular}}}}%
    \put(0,0){\includegraphics[width=\unitlength,page=6]{sensitivity_0_1.pdf}}%
    \put(0.61811079,0.03827971){\makebox(0,0)[t]{\lineheight{1.25}\smash{\begin{tabular}[t]{c}12.5\end{tabular}}}}%
    \put(0,0){\includegraphics[width=\unitlength,page=7]{sensitivity_0_1.pdf}}%
    \put(0.7308013,0.03827971){\makebox(0,0)[t]{\lineheight{1.25}\smash{\begin{tabular}[t]{c}15.0\end{tabular}}}}%
    \put(0,0){\includegraphics[width=\unitlength,page=8]{sensitivity_0_1.pdf}}%
    \put(0.84349174,0.03827971){\makebox(0,0)[t]{\lineheight{1.25}\smash{\begin{tabular}[t]{c}17.5\end{tabular}}}}%
    \put(0,0){\includegraphics[width=\unitlength,page=9]{sensitivity_0_1.pdf}}%
    \put(0.95618225,0.03827971){\makebox(0,0)[t]{\lineheight{1.25}\smash{\begin{tabular}[t]{c}20.0\end{tabular}}}}%
    \put(0.50429338,0.00039225){\makebox(0,0)[t]{\lineheight{1.25}\smash{\begin{tabular}[t]{c}Time in sec\end{tabular}}}}%
    \put(0,0){\includegraphics[width=\unitlength,page=10]{sensitivity_0_1.pdf}}%
  \end{picture}%
\endgroup%

%% file: pdftex/sensitivity_1.pdf_tex
\begingroup%
  \makeatletter%
  \providecommand\color[2][]{%
    \errmessage{(Inkscape) Color is used for the text in Inkscape, but the package 'color.sty' is not loaded}%
    \renewcommand\color[2][]{}%
  }%
  \providecommand\transparent[1]{%
    \errmessage{(Inkscape) Transparency is used (non-zero) for the text in Inkscape, but the package 'transparent.sty' is not loaded}%
    \renewcommand\transparent[1]{}%
  }%
  \providecommand\rotatebox[2]{#2}%
  \newcommand*\fsize{\dimexpr\f@size pt\relax}%
  \newcommand*\lineheight[1]{\fontsize{\fsize}{#1\fsize}\selectfont}%
  \ifx\svgwidth\undefined%
    \setlength{\unitlength}{361.01998901bp}%
    \ifx\svgscale\undefined%
      \relax%
    \else%
      \setlength{\unitlength}{\unitlength * \real{\svgscale}}%
    \fi%
  \else%
    \setlength{\unitlength}{\svgwidth}%
  \fi%
  \global\let\svgwidth\undefined%
  \global\let\svgscale\undefined%
  \makeatother%
  \begin{picture}(1,0.82386069)%
    \lineheight{1}%
    \setlength\tabcolsep{0pt}%
    \put(0,0){\includegraphics[width=\unitlength,page=1]{sensitivity_1.pdf}}%
    \put(0.05465827,0.03827969){\makebox(0,0)[t]{\lineheight{1.25}\smash{\begin{tabular}[t]{c}0.0\end{tabular}}}}%
    \put(0,0){\includegraphics[width=\unitlength,page=2]{sensitivity_1.pdf}}%
    \put(0.16734877,0.03827969){\makebox(0,0)[t]{\lineheight{1.25}\smash{\begin{tabular}[t]{c}2.5\end{tabular}}}}%
    \put(0,0){\includegraphics[width=\unitlength,page=3]{sensitivity_1.pdf}}%
    \put(0.28003928,0.03827969){\makebox(0,0)[t]{\lineheight{1.25}\smash{\begin{tabular}[t]{c}5.0\end{tabular}}}}%
    \put(0,0){\includegraphics[width=\unitlength,page=4]{sensitivity_1.pdf}}%
    \put(0.39272981,0.03827969){\makebox(0,0)[t]{\lineheight{1.25}\smash{\begin{tabular}[t]{c}7.5\end{tabular}}}}%
    \put(0,0){\includegraphics[width=\unitlength,page=5]{sensitivity_1.pdf}}%
    \put(0.50542033,0.03827969){\makebox(0,0)[t]{\lineheight{1.25}\smash{\begin{tabular}[t]{c}10.0\end{tabular}}}}%
    \put(0,0){\includegraphics[width=\unitlength,page=6]{sensitivity_1.pdf}}%
    \put(0.61811086,0.03827969){\makebox(0,0)[t]{\lineheight{1.25}\smash{\begin{tabular}[t]{c}12.5\end{tabular}}}}%
    \put(0,0){\includegraphics[width=\unitlength,page=7]{sensitivity_1.pdf}}%
    \put(0.73080134,0.03827969){\makebox(0,0)[t]{\lineheight{1.25}\smash{\begin{tabular}[t]{c}15.0\end{tabular}}}}%
    \put(0,0){\includegraphics[width=\unitlength,page=8]{sensitivity_1.pdf}}%
    \put(0.84349179,0.03827969){\makebox(0,0)[t]{\lineheight{1.25}\smash{\begin{tabular}[t]{c}17.5\end{tabular}}}}%
    \put(0,0){\includegraphics[width=\unitlength,page=9]{sensitivity_1.pdf}}%
    \put(0.95618231,0.03827969){\makebox(0,0)[t]{\lineheight{1.25}\smash{\begin{tabular}[t]{c}20.0\end{tabular}}}}%
    \put(0.50429344,0.00039223){\makebox(0,0)[t]{\lineheight{1.25}\smash{\begin{tabular}[t]{c}Time in sec\end{tabular}}}}%
    \put(0,0){\includegraphics[width=\unitlength,page=10]{sensitivity_1.pdf}}%
  \end{picture}%
\endgroup%

%% file: sec/concl.tex
\section{Conclusion}\label{sec:concl}
In this paper we have developed a method to find a potential that densely endows the configuration space with strict normal modes, which describe natural periodic oscillations of the system including the potential.
The main take-off message is that, by designing the potential field to harmonize with the inertial properties, i.e., with the natural Riemannian metric of the system, the highly nonlinear system can display a large variety of very regular motions.

The implementation of the potential directly in the mechanics by combining nonlinear elastic elements that create the optimized potential was not within the scope of this paper and will be addressed in future work.
However, considering the sensitivity analysis, it can be concluded that certain discrepancies between the implemented and the optimal potential can be tolerated.

The strict modes allow to efficiently execute a variety of periodic oscillations.
Stabilization onto a desired strict mode can be realized by the proposed mode stabilization controller.
Further the system can be swung up to the desired energy level using the energy regulation controller.
The latter also reinjects energy that is lost due to damping or interactions.
We have shown in simulations experiments that the control torques are very low compared to the torques due to the elastic elements.

%% file: sec/appendix.tex
\section{}\label{app:pendulum_parameters}

\begin{table}[h]
	\vspace{-.4cm}
	\caption{Parameters of the Double Pendulum}\label{tab:pendulum_parameters}
	\centering
	\begin{tabular}{c|c|c}
		\textbf{} & Link 1 & Link 2 \\\hline
		Mass & $m_1=\SI{1}{\kilogram}$ & $m_2=\SI{1}{\kilogram}$ \\
		Length & $l_1=\SI{2}{\meter}$ & $l_2=\SI{2}{\meter}$ \\
		Center of Mass from Origin & $l_{c1} = \SI{1}{\meter}$ & $l_{c2} = \SI{1}{\meter}$ \\
		Moments of Inertia & $I_1 = \SI{1}{\kilogram\square\meter}$ & $I_2 = \SI{1}{\kilogram\square\meter}$ \\
	\end{tabular}	
\end{table}

\vspace{-.3cm}
\section{Mode Identification Coordinate}\label{app:labeling_function}
The direction of the velocity when passing through the equilibrium is used to identify a strict mode, where velocity vectors $\qdot$ and $-\qdot$ identify the same mode.
Therefore, we set the label of the strict mode to \begin{equation}\label{eq:theta_def}
	\theta = 2\atantwo(\dot q_2, \dot q_1),
\end{equation}
where $\dot q_1$ and $\dot q_2$ are the two components of the velocity vector at the equilibrium.
The factor of $2$ ensures that $\qdot$ and $-\qdot$ correspond to the same label $\theta$.

In order to obtain a function $\theta(\qq)$ encoding the label of the mode passing through $\qq$, we generate system geodesics $\ggamma_i$ and assign a label $\theta_i$ to them according to (\ref{eq:theta_def}).
Each geodesic is represented by $T_i$ samples $\ggamma_{it} \in \bar{\mathcal{Q}}$.
This creates an amount of $\sum_i T_i$ training tuples $(\ggamma_{it}, \theta_i)$.
However, the raw function $\theta(\qq)$ cannot be approximated by a scalar function as it parametrizes $\mathbb{S}^1$ and it is not defined at the equilibrium.
We therefore define another function \begin{equation}
	\Lambda(\qq) = ||\qq - \qeq||^2 \begin{bmatrix}
		\cos \theta \\ \sin \theta
	\end{bmatrix},
\end{equation}
which can be used to reconstruct $\theta(\qq)$ by \begin{equation}
	\theta(\qq) = \atantwo(\Lambda_2(\qq), \Lambda_1(\qq)).
\end{equation}
In contrast to $\theta(\qq)$ the function $\LLambda(\qq)$ is smooth.
We take a neural network with $\tanh$ activation functions, two hidden layers and two outputs and use it to approximate $\LLambda(\qq)$ based on the training data $(\ggamma_{it}, \theta_i)$.

As $\theta \in \mathbb{S}^1$ some care must be taken when computing the difference to the desired mode $\theta_d$. 
Let \begin{equation}
	R_d = \begin{bmatrix}
		\cos \theta_d & -\sin \theta_d \\
		\sin \theta_d & \cos \theta_d
	\end{bmatrix} \text{ and }
	A(\qq) = \begin{bmatrix}
		\Lambda_1(\qq) & -\Lambda_2(\qq) \\
		\Lambda_2(\qq) & \Lambda_1(\qq)
	\end{bmatrix}.\nonumber
\end{equation}
The basic idea is to express $\theta(\qq)$ as rotation matrix.
We need to orthonormalize $A(\qq)$ by using the singular value decomposition to obtain $U\Sigma V\tran = A(\qq)$ and set $Q(\qq) = UV\tran$, then 
\begin{subequations}
\begin{align}
	R_\Delta(\qq, \theta_d) &= R_d\tran Q(\qq)\\
	\Delta\theta(\qq, \theta_d) &= \atantwo(r_{21}(\qq, \theta_d), r_{11}(\qq, \theta_d)),\label{eq:delta_theta}
\end{align}
\end{subequations}
where $r_{ij}(\qq, \theta_d)$ are the respective elements in $R_\Delta(\qq, \theta_d)$.
The derivative of $\Delta\theta(\qq)$ with respect to $\qq$ can be computed using
\begin{equation}\label{eq:theta_jacobian}
	\frac{\partial \theta}{\partial \qq\tran} = \begin{bmatrix}
		-\sin \Delta\theta & \cos\Delta\theta
	\end{bmatrix}\frac{\partial \LLambda}{\partial \qq}.
\end{equation}

For the damping design we first transform the mass-matrix $\MM(\qq)$ of the manipulator to $\theta(\qq)$ coordinates and then compute the damping coefficient
\begin{subequations}
	\begin{align}
	m_{\theta}(\qq) &= \frac{1}{\left(\frac{\partial \theta}{\partial \qq}\right)\tran\Minv(\qq)\frac{\partial \theta}{\partial \qq}}\\
	d_{\theta}(\qq) &= 2\zeta\sqrt{k_\theta m_\theta(\qq)}\,\label{eq:damping_design}.
	\end{align}
\end{subequations}